\documentclass[10pt,twocolumn,letterpaper]{article}

\usepackage[pagenumbers]{wacv}            % arXiv: camera-ready style with page numbers
\usepackage[T1]{fontenc}
\usepackage{url}
\usepackage{booktabs}
\usepackage{amsfonts}
\usepackage{amsmath}
\usepackage{nicefrac}
\usepackage{microtype}
\usepackage{graphicx}
\usepackage{float}
\usepackage{enumitem}
\usepackage{colortbl}
\usepackage{pifont}
\usepackage{multirow}
\usepackage{array}
\usepackage{algorithm}
\usepackage{algpseudocode}
\usepackage{caption}
\usepackage{rotating}                     % needed by the appended supplementary material

\newcommand{\cmark}{\textcolor{green!60!black}{\ding{51}}}
\newcommand{\xmark}{\textcolor{red}{\ding{55}}}
\definecolor{rldfblue}{RGB}{219,234,254}
\definecolor{wacvblue}{rgb}{0.21,0.49,0.74}
\usepackage[pagebackref,breaklinks,colorlinks,allcolors=wacvblue]{hyperref}

\def\wacvPaperID{446} % *** Enter the WACV Paper ID here
\def\confName{WACV}
\def\confYear{2027}

\title{RLDF: A Region-Level Deepfake Dataset and Benchmark for\\ Partial-Face Manipulation Detection}

\author{Shukesh Reddy \quad Abhijit Das\\
Machine Intelligence Group, Department of CS\&IS\\
Birla Institute of Technology and Science, Pilani, Hyderabad Campus, India\\
}

\begin{document}
\maketitle
\begin{abstract}
Existing deepfake detection models are built upon manipulated images that are tampered with either exclusively fully swapped or fully synthesised faces, which leads us to an open question: Can off-the-shelf deepfake models detect manipulation effectively while tampering is confined to a local facial sub-region? To fill this gap, we introduce \textbf{RLDF}
(\textbf{R}egion-\textbf{L}evel \textbf{D}eep\textbf{f}ake), the first dataset
for region-level deepfake evaluation. Employing the proposed \textbf{region-aware partial
swap synthesis} model, we generate over \textbf{2M} partial face-swap images by using eight face-swapping methods. Altogether, we considered seven face regions, among which there are four discrete regions (eye, nose, mouth and eyebrow) and four composite facial regions (eye and nose; eye, nose and mouth; eye, nose, mouth and eyebrow; and full face). In addition, we define three evaluation protocols and benchmark \textbf{12} state-of-the-art detectors across the five popular standard cross-dataset evaluations, DF40, and the proposed RLDF, providing a comprehensive audit of where and why existing detectors underperformed. Our evaluation reveals that existing detectors fail to generalize consistently across partial facial regions, resulting in an overall drop of \textbf{10.5\%} AUC and \textbf{10\%} accuracy under region-level evaluation. Further, we propose a \textbf{mask-guided cross-attention} model as a baseline approach that can effectively detect region-level manipulation.
%by jointly predicts a spatial forgery map and a binary label, directing a learnable forgery query toward manipulated patches via a differentiable log-space soft mask. 
The proposed dataset, code, and checkpoints are available at \url{https://github.com/RLDF-Dataset/RLDF}.
\end{abstract}
\section{Introduction}
\label{sec: introduction}
Rapid advancements in the use of deep generative~\cite{goodfellow2014gan,karras2019stylegan}, diffusion~\cite{ho2020ddpm,rombach2022ldm}, and Large Language Models (LLMs)~\cite{brown2020gpt3,achiam2023gpt4}, have dramatically lowered the barrier towords generating content creation across image, video, and audio domains. This has enabled the generation of highly realistic synthetic media, which is often indistinguishable from real content. Face manipulation, in particular, has seen rapid growth through various approaches, including face swapping~\cite{faceswap,li2019faceshifter,SimSwap,facedancer,fsgan, inswapper,uniface}, face reenactment~\cite{lia,sadtalker,pirenderer,wav2lip}, facial attribute editing~\cite{stargan,styleclip,starganv2,e4e}, and fully synthetic face generation~\cite{rombach2022ldm,peebles2023dit,
chen2024pixart,karras2020stylegan2}. The misuse of such techniques for identity fraud, misinformation, and non-consensual content has made deepfake detection an active research problem. 

To mitigate these, numerous deepfake detection methods have been proposed~\cite{ff++,Facex-ray,qian2020thinking,yan2023ucf,RECCE} which are broadly categorized into CNN-based classifiers~\cite{ff++,afchar2018mesonet,nguyen2019capsule}, spatial-domain detectors~\cite{Facex-ray,shiohara2022detecting,yan2023ucf,wangCVPR21rfm}, 
and frequency-domain methods~\cite{qian2020thinking,liu2021spatial,luo2021generalizing}. More recently, Vision Transformer and CLIP-based detectors~\cite{fatformer,effort,prodet,dfd_fcg} have leveraged large-scale pretrained representations to improve generalization. Although these methods exhibit strong in-domain performance, they demonstrate a significant drop when evaluated in a cross domain/cross-dataset 
settings~\cite{DeepfakeBench,df40}, as they tend to overfit to dataset-specific artifacts rather than learning the underlying manipulation patterns.
% Required in preamble:
% \usepackage{booktabs, colortbl, xcolor, pifont, array}
% \newcommand{\cmark}{\textcolor{green!60!black}{\ding{51}}}
% \newcommand{\xmark}{\textcolor{red!70!black}{\ding{55}}}
% \definecolor{rldfblue}{RGB}{219,234,254}

\begin{table*}[tb]
\centering
\caption{Comparison of existing deepfake datasets. All prior datasets annotate manipulation with a
binary image-level label and contain only full-face swaps. \textbf{RLDF} is the first benchmark
to provide \textit{region-level} annotations for partially manipulated faces.
\textbf{Region Labels}: per-region GT labels across 8 sub-facial areas.
\textbf{Localization GT}: spatial ground-truth for forgery localization.
\textbf{Partial-Face}: manipulation on sub-regions only.}
\label{tab:cmp_other_datasets}
\setlength{\tabcolsep}{5pt}
\renewcommand{\arraystretch}{1.0}
\scriptsize
\begin{tabular}{@{}l l c r c c c@{}}
\toprule
\textbf{Dataset} & \textbf{Pub.} & \textbf{\#Methods} & \textbf{Fake Images}
  & \textbf{Region Labels}
  & \textbf{Localization GT}
  & \textbf{Partial-Face} \\
\midrule
DF-TIMIT~\cite{korshunov2018deepfakes}              & ArXiv'18   & 2  & --      & \xmark & \xmark & \xmark \\
UADFV~\cite{uadfv}                       & ICASSP'19  & 1  & 252     & \xmark & \xmark & \xmark \\
FaceForensics++~\cite{ff++}   & ICCV'19    & 4  & --      & \xmark & \xmark & \xmark \\
DeepFakeDetection~\cite{dfd}                        & --         & 5  & --      & \xmark & \xmark & \xmark \\
CDF~\cite{li2019celeb}                              & CVPR'20    & 1  & --      & \xmark & \xmark & \xmark \\
DFFD~\cite{dang2020detection}                       & CVPR'20    & 7  & 0.2M+   & \xmark & \xmark & \xmark \\
DeeperForensics~\cite{jiang2020deeperforensics}     & CVPR'20    & 1  & --      & \xmark & \xmark & \xmark \\
DFDC~\cite{dfdc}                   & ArXiv'20   & 8  & --      & \xmark & \xmark & \xmark \\
ForgeryNet~\cite{he2021forgerynet}                  & CVPR'21    & 15 & 1M+     & \xmark & \xmark & \xmark \\
FakeAVCeleb~\cite{khalid2021fakeavceleb}            & NeurIPS'21 & 4  & --      & \xmark & \xmark & \xmark \\
KoDF~\cite{kwon2021kodf}                            & ICCV'21    & 6  & --      & \xmark & \xmark & \xmark \\
FFIW~\cite{zhou2021face}                            & CVPR'21    & 3  & --      & \xmark & \xmark & \xmark \\
DF3~\cite{ju2023glff}                               & TMM'22     & 6  & --      & \xmark & \xmark & \xmark \\
DeepFakeFace~\cite{song2023robustness}              & ArXiv'23   & 3  & 90K     & \xmark & \xmark & \xmark \\
DF-Platter~\cite{narayan2023df}                     & CVPR'23    & 3  & --      & \xmark & \xmark & \xmark \\
DiffusionDeepfake~\cite{bhattacharyya2024diffusion} & ArXiv'24   & 2  & 0.1M+   & \xmark & \xmark & \xmark \\
DF40~\cite{df40}                                    & NeurIPS'24 & 40 & 1M+     & \xmark & \xmark & \xmark \\
% INDIFACE~\cite{Indiface}                            & FG'24 & 2 & --     & \xmark & \xmark & \xmark \\
\midrule
\rowcolor{rldfblue}
\textbf{RLDF (Ours)} & \textbf{--} & \textbf{8} & \textbf{2M+}
  & \cmark & \cmark & \cmark \\
\bottomrule
\end{tabular}
\renewcommand{\arraystretch}{1.0}
\end{table*}

In this direction, several benchmark datasets have been introduced, including FaceForensics++~\cite{ff++}, CelebDF~\cite{li2019celeb}, DFDC~\cite{dfdc}, DFDCP~\cite{dolhansky2019deepfake}, and most recently DF40~\cite{df40}.  For example, DF40 aggregates 40 distinct generation methods across face swapping, reenactment, full synthesis, and editing. However, all existing datasets share a common limitation that the tampering is committed at the global image level with a binary label of annotation, i.e., indicating whether the image is manipulated or not. %Detectors attempt to localize forgery cues by attending to specific entangled features~\cite{Facex-ray,zhao2021multi,effort,fatformer,prodet}, yet they are still trained and evaluated on fully swapped or fully synthetic faces. 
Hence, existing deepfake detectors~\cite{Facex-ray,zhao2021multi,effort,fatformer,prodet} trained based on the above-mentioned datasets attempt to learn a global representation to distinguish whether an image is fake or real. Therefore, it can be assumed that the representation of these detectors remains unaware of explicit quantified spatial manipulated cues. Consequently, when manipulation is confined to a specific iota of the region, the existing models struggle to detect the manipulation. %In other words, the literature has recognized that deepfake manipulations are largely confined to global, however, facial features and localized artifacts for detection and explicit modeling of fine-grained, region-level manipulation patterns remain underexplored. In addition, existing methods often lack a structured framework for capturing spatial dependencies and interactions across distinct facial subregions. 
To the best of our knowledge, no existing work or datasets have studied deepfake detection in which manipulation is performed at the distinct face region. This leaves a critical and unexplored gap in the literature.
To address this gap, we introduce \textbf{RLDF}, the first dataset designed for partial or region-level deepfake evaluation. By leveraging existing full face-swapping methods~\cite{SimSwap,facedancer,fsgan,inswapper,uniface,blendface,ghost,e4s}. We proposed a region-aware partial swap model that can manipulate specific region such as the eye, nose, mouth, and eyebrow and four composite regions that progressively expand in spatial coverage from paired regions up to full-face replacement (Section~\ref{sec:methodology}). \textbf{RLDF comprises over \textbf{2M}, \textbf{8} face-swapping methods, and \textbf{8} facial regions.}
Our specific contributions are as follows: \textbf{(1)} The first region-level deepfake dataset with partial face-swap annotations. \textbf{(2)} Benchmarking across three evaluation protocols including \textbf{21 manipulation methods} from existing datasets, and RLDF region-level evaluation, rebelling that performance of existing detectors drop significantly under partial-face conditions.
\textbf{(3)} A \textbf{mask-guided cross-attention} baseline that sets a starting point for future work, confirming that patch-level forgery grounding is a promising direction for region-level manipulation detection.
\vspace{-3mm}
\section{Related Work}
\label{sec:related}
\noindent\textbf{Deepfake Detection:}
Seminal deepfake detectors explores physiological signals~\cite{li2018ictu,yang2018exposing}
and generation artefacts in spatial and frequency
domains~\cite{li2019exposing,qian2020thinking}. Following works in the literature paved a complementary line of research by designing attention mechanisms, contrastive
objectives, and feature-disentanglement strategies to learn more discriminative
representations~\cite{dang2020detection,yan2023ucf}.
More recently, Vision Transformer and CLIP-based
features~\cite{fatformer,effort,prodet,dfd_fcg} are employed to improve cross-dataset generalization ~\cite{shiohara2022detecting,Facex-ray,zhou2024freqblender}. Further, pseudo-fake synthesis methods are proposed which simulate blending artefacts at training time to reduce overfitting to seen forgeries~\cite{DeepfakeBench,df40}.
Despite these advances, all existing detectors are designed and evaluated on fully swapped or fully synthesized faces. When subjected to region-level evaluation, where manipulation is confined to a specific facial sub-region, performance drops sharply across all methods, a failure mode that standard benchmarks cannot reveal.

\noindent\textbf{Deepfake Datasets:}
Several Deepfake datasets have been proposed in the literature, such as UADFV~\cite{uadfv}, DFTIMIT~\cite{korshunov2018deepfakes}, FaceForensics++~\cite{ff++}, Celeb-DF~\cite{li2019celeb},
DFDC~\cite{dfdc}, DeeperForensics~\cite{jiang2020deeperforensics}, and
DF40~\cite{df40}. Yet across all of these datasets, manipulation is annotated at the image-level or video label. Moreover, the manipulation is performed considering the full face. There exists no dataset that provides region-level annotations, partial-face samples, or spatial
ground-truth (Table~\ref{tab:cmp_other_datasets}).
We proposed \textit{\textbf{RLDF}} to mitigate the above gaps. Evaluating existing detectors on RLDF reveals a consistent drop in performance. AUC of FatFormer~\cite{fatformer} and DFD-FCG~\cite{dfd_fcg} drop to 0.745 and 0.806
respectively on RLDF, a degradation of up to 10\% AUC from their DF40 scores. (Refer to Section~\ref{sec:experiments} for detailed evaluation protocol).
\vspace{-3mm}
\section{Methodology}
\label{sec:methodology}
We present a unified contribution addressing region-level deepfake detection through two tightly coupled components: a large-scale benchmark that systematically 
exposes how existing detectors fail under partial-face manipulation, and a mask-guided detection framework. 
%We describe the dataset construction in Section~\ref{sec:construction} and the detection framework in Section~\ref{sec:detector}.
\vspace{-3mm}
\subsection{RLDF Dataset}
\label{sec:construction}
\begin{figure*}[tb]
    \centering
    \includegraphics[width=1\linewidth]{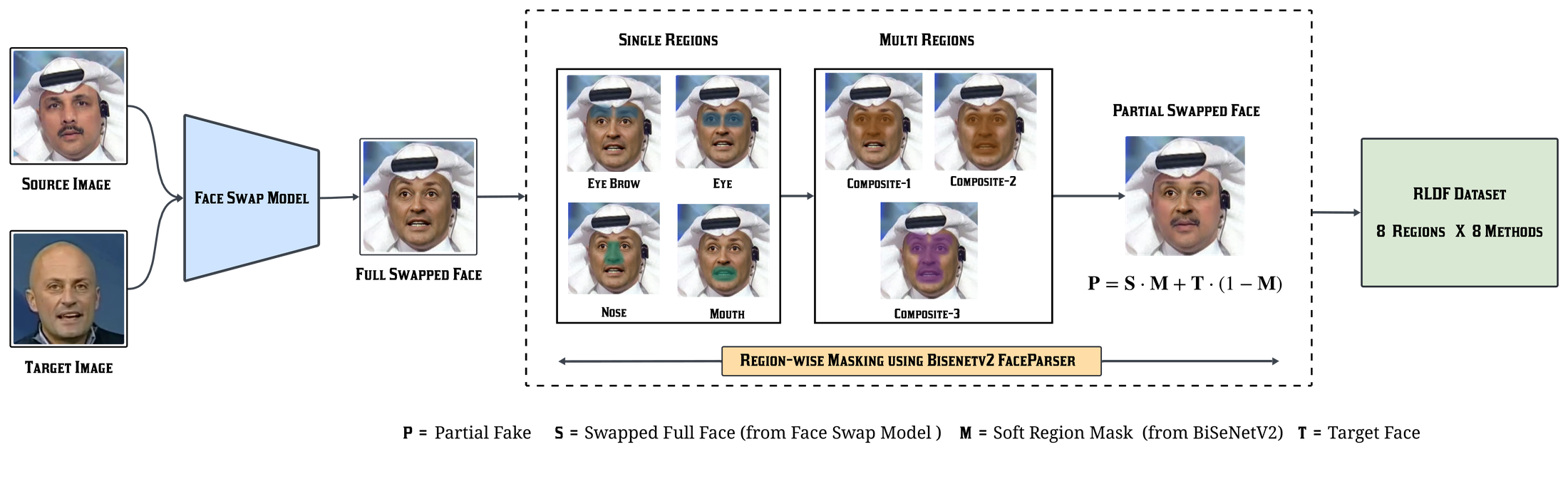}
    \caption{Overview of the proposed \textbf{region-aware partial swap synthesis:} Given source–target video pairs, (1) aligned face crops are extracted and passed through eight face-swapping methods to generate full face-swaps, (2) a BiSeNetV2 facial parser derives soft region masks from the target frame, and (3) the designated facial region from the full swap is composited back onto the target image, producing partial fakes across eight facial regions. Representative samples for every region and swap method are provided in Appendix~F.}
    \label{fig:placeholder}
\end{figure*}

\vspace{-3mm}
Let $\varphi: \mathcal{I}_s \times \mathcal{I}_t \rightarrow \hat{\mathcal{I}}$
denote a face-swapping method that transfers the identity of a source image
$I_s \in \mathbb{R}^{3\times H\times W}$ to a target image
$I_t \in \mathbb{R}^{3\times H\times W}$, producing a complete face-swap output
$\hat{\mathbf{I}}_f = \varphi(I_s, I_t)$ that preserves the target's pose,
expression, and background while replacing the facial identity.
We consider a diverse set of eight state-of-the-art face-manipulation methods: $\Phi$ = {\text{BlendFace},\, \text{E4S},\, \text{FaceDancer},\, \text{FSGAN},\,\text{GHOST},\, \text{InSwapper},\, \text{SimSwap},\, \text{UniFace}}. These are based on GAN-based, latent-space, reenactment, and diffusion-based generation paradigms. No method in $\Phi$ supports partial manipulation; all operate exclusively on the full facial region. Modifying these methods for
localized manipulation is infeasible at scale across eight architecturally distinct
systems. We therefore introduce a \textbf{region-aware partial swap synthesis} that
composites only the designated facial region from $\hat{\mathbf{I}}_f$ back onto
$I_t$ using a semantically derived soft mask, producing partial fakes without any
modification to $\varphi$. Figure~\ref{fig:placeholder} illustrates the pipeline and
Algorithm~\ref{alg:pipeline} formalizes the complete procedure.

\begin{algorithm}[t]
\small
\caption{Region-Aware Partial Swap Synthesis}
\label{alg:pipeline}
\begin{algorithmic}[1]
\Require Video pairs $\{(v_s^{(k)}, v_t^{(k)})\}_{k=1}^{1000}$, swap methods $\Phi$,
         regions $\mathcal{R}$, frames per pair $T=32$
\Ensure Dataset $\mathcal{D}_{\text{RLDF}}$ with tuples $(\mathbf{I}_i, r_i, y_i)$
\For{each pair $(v_s, v_t)$}
    \State Sample $T$ frames via linspace, detect and align~\cite{insightface}, crop to $224{\times}224$
    \For{each target frame $I_t$}
        \State Add $(I_t,\, \texttt{full},\, 0)$ to $\mathcal{D}_{\text{RLDF}}$ \Comment{$y=0$: real}
    \EndFor
    \For{each method $\varphi \in \Phi$}
        \State Generate full swap: $\hat{\mathbf{I}}_f = \varphi(I_s,\, I_t)$
        \State Parse $I_t$ via BiSeNetV2, dilate per region, Gaussian smooth ($\sigma{=}3.0$) $\rightarrow \mathbf{M}_r$
        \For{each region $r \in \mathcal{R}$}
            \State $\mathbf{I}_{p}^{(r)} \leftarrow \mathbf{M}_r \odot \hat{\mathbf{I}}_f + (1 - \mathbf{M}_r) \odot I_t$
            \State Add $(\mathbf{I}_{p}^{(r)},\, r,\, 1)$ to $\mathcal{D}_{\text{RLDF}}$ \Comment{$y=1$: fake}
        \EndFor
    \EndFor
\EndFor
\State Partition into train/val/test (720/140/140 pairs, no identity overlap)
\end{algorithmic}
\end{algorithm}
From the FaceForensics++ corpus~\cite{ff++}, we form 1,000 source–target pairs
$(v_s, v_t)$ and sample $T = 32$ temporally aligned target frames via linspace
indexing: $t_j = \lfloor j(F_t - 1)/(T-1) \rfloor$, where $F_t$ is the total
frame count. For each frame, InsightFace Antelope~\cite{insightface} detects the
primary face at $640{\times}640$ resolution and produces a $224{\times}224$
similarity-transformed aligned crop normalizing pose and scale. Frames with no
detected face ($<\!3\%$) are discarded. The full swap is then generated per
Eq.~\eqref{eq:fullswap}, and all mask extraction and blending operate on this
aligned representation.
\begin{equation}
    \hat{\mathbf{I}}_f = \varphi(I_s,\, I_t), \qquad \varphi \in \Phi.
    \label{eq:fullswap}
\end{equation}
For each target frame $I_t$, region masks $\mathbf{M}_r \in [0,1]^{H \times W}$ are
obtained via BiSeNetV2~\cite{BiSeNet_V2} facial parsing, followed by
morphological dilation to compensate for under-segmentation in small regions
(eyes, eyebrow, nose, mouth) and Gaussian smoothing ($\sigma{=}3.0$) to produce natural soft
boundaries. Critically, masks are parsed from the unmodified target $I_t$,
so region boundaries follow real facial anatomy rather than
the swap-induced artifacts present in $\hat{\mathbf{I}}_f$.
The partial fake is then synthesized as:
\begin{equation}
    \mathbf{I}_{p}^{(r)} = \mathbf{M}_r \odot \hat{\mathbf{I}}_f\; +\;
    (1 - \mathbf{M}_r) \odot I_t,
    \label{eq:partial_swap}
\end{equation}

where pixels inside the region ($\mathbf{M}_r \approx 1$) carry swap artifacts and
pixels outside ($\mathbf{M}_r \approx 0$) are drawn exactly from $I_t$. Detection
is thus strictly harder than whole-face forgery: the model must localize artifacts
within a semantically confined sub-region while ignoring the surrounding
authentic context.
The eight regions $\mathcal{R}$ comprise four \textit{individual} regions
(\textit{eye, eyebrow, nose, mouth}) and four \textit{composite} regions of
progressively increasing spatial coverage:
\textit{Composite-1} (eye\,+\,nose),
\textit{Composite-2} (eye\,+\,nose\,+\,mouth), and
\textit{Composite-3} (eye\,+\,nose\,+\,mouth\,+\,eyebrow),
plus \textit{Full Face} as an upper bound, yielding eight manipulation in total.

Applying Eq.~\eqref{eq:partial_swap} across all $|\Phi|{=}8$ methods,
$|\mathcal{R}|{=}8$ regions, $T{=}32$ frames, and 1,000 pairs yields
$1{,}000 \times 32 \times 64 = 2{,}048{,}000$ fake images, balanced with an equal
number of unmodified target frames as real samples ($y{=}0$).

\subsection{Mask-Guided Cross-Attention Framework}
\label{sec:detector}

\begin{figure*}[tb]
    \centering
    \includegraphics[width=1\linewidth]{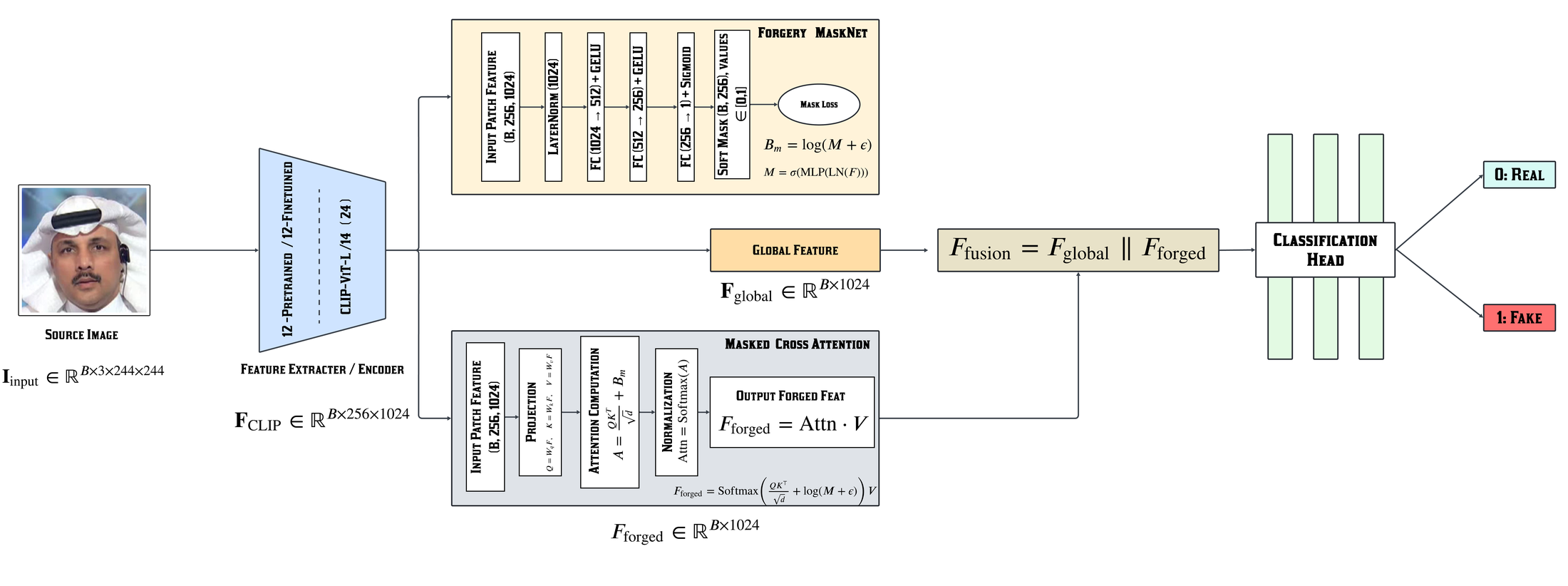}
    \caption{Overview of the proposed \textbf{mask-guided cross-attention framework}. A partially fine-tuned CLIP ViT-L/14 encoder extracts $N{=}256$ patch tokens. \textbf{Forgery MaskNet} assigns each patch a forgery score $\hat{m}_i \in [0,1]$. \textbf{Masked Cross-Attention} uses these scores as attention biases so the model focuses on suspicious patches. A global feature (mean-pooled) and the region-focused feature are concatenated and passed to a classifier.}
    \label{fig:architecture}
\end{figure*}

Existing detectors predict a binary label $y \in \{0,1\}$ from the entire image of \emph{the face, without knowing where} the manipulation occurred on the face.
This works for full face fake images, but fails for region-level fakes where the forged area is small.
To address this, we propose MCA that predicts both a binary label and a spatial forgery map:
\begin{equation}
    f: \mathbf{I} \;\longrightarrow\; \bigl(\hat{y} \in [0,1],\;\;
    \hat{\mathbf{m}} \in [0,1]^N\bigr),
\end{equation}
where $\hat{m}_i \approx 1$ means patch $i$ is likely forged.
The map $\hat{\mathbf{m}}$ is not a standalone output it guides a cross-attention module that builds a forgery-focused feature, which is then combined with a global feature for the final prediction.
Figure~\ref{fig:architecture} shows an overview.

Given an image $\mathbf{I} \in \mathbb{R}^{3 \times 224 \times 224}$, we use CLIP ViT-L/14~\cite{clip} to extract $N{=}256$ patch tokens $\mathbf{P} \in \mathbb{R}^{N \times D}$ ($D{=}1024$) from a $16{\times}16$ grid.
The first 12 of 24 transformer blocks are frozen. The last 12 are fine-tuned with $\text{lr}_{\text{enc}} = 10^{-6}$.

\noindent\textbf{Forgery MaskNet:}
MaskNet assigns a forgery score to each of the $N{=}256$ patches independently.
A lightweight three-layer MLP processes each patch token and outputs a probability $\hat{m}_i \in [0,1]$:
\begin{equation}
\begin{aligned}
    \hat{\mathbf{m}} &= \sigma\!\left(\text{MLP}\bigl(\text{LN}(\mathbf{P})\bigr)\right),\\
    \mathbb{R}^{1024} &\xrightarrow{\text{GELU}} \mathbb{R}^{512}
                       \xrightarrow{\text{GELU}} \mathbb{R}^{256}
                       \xrightarrow{\sigma}      \mathbb{R}^{1}.
\end{aligned}
    \label{eq:masknet}
\end{equation}
Each patch is scored independently, without looking at neighbouring patches.
This is intentional, as sharing information between patches can cause the mask to collapse to a trivial all-zero or all-one output.
The scores are learned entirely from binary real/fake labels, without any spatial annotations.

\noindent\textbf{Masked Cross-Attention (MCA):}
A standard mean-pooled feature treats all patches equally, irrespective of whether they belong to a forged or authentic region.
To focus on the forged area, we introduce MCA, which builds a feature $\mathbf{F}_f$ that attends selectively to suspicious patches.

We use a single learnable query vector $\mathbf{q} \in \mathbb{R}^{1 \times D}$ that represents what a forged region looks like.
It is projected to $\mathbf{Q} = \mathbf{W}_Q\mathbf{q}$, and all patch tokens form keys $\mathbf{K} = \mathbf{W}_K\mathbf{P}$ and values $\mathbf{V} = \mathbf{W}_V\mathbf{P}$.
The forgery scores $\hat{\mathbf{m}}$ from MaskNet are added to the attention logits as a log-space bias, yielding $\mathbf{F}_f$ with $d{=}128$ (8 heads) and $\epsilon{=}10^{-6}$ defined in Eq.~\eqref{eq:mca}:
\begin{equation}
    \mathbf{F}_f = \mathrm{LN}\!\left(\mathbf{W}_O \cdot
    \mathrm{Softmax}\!\left(
    \frac{\mathbf{Q}\mathbf{K}^\top}{\sqrt{d}}
    + \log\!\left(\hat{\mathbf{m}} + \epsilon\right)
    \right) \cdot \mathbf{V}\right).
    \label{eq:mca}
\end{equation}
The bias acts as a soft mask over the attention logits, where a forged patch ($\hat{m}_i \approx 1$) adds $\approx 0$ and is included normally, while an authentic patch ($\hat{m}_i \approx 0$) adds $-\infty$ and is suppressed.
As a result, $\mathbf{F}_f$ captures only the forged regions.
This soft mask is fully differentiable, so MaskNet and MCA improve each other during training without any spatial supervision.
On real images, all scores are near zero, so $\mathbf{F}_f \approx \mathbf{F}_g$ and no false forgery signal is produced.

\noindent\textbf{Dual-Path Fusion and Classification:}
We compute a global feature $\mathbf{F}_g = \frac{1}{N}\sum_i \mathbf{P}_i$ by averaging all patch tokens.
This captures the overall global feature from the input image.
We concatenate $\mathbf{F}_g$ and $\mathbf{F}_f$ and pass the combined feature to a classifier $\mathcal{C}$:
\begin{equation}
\begin{aligned}
    \hat{y} &= \sigma\!\left(\text{clamp}\bigl(\mathcal{C}([\mathbf{F}_g;\,\mathbf{F}_f]),
              -15, 15\bigr)\right),\\
    \mathbb{R}^{2048} &\xrightarrow{\text{GELU}} \mathbb{R}^{1024}
                       \xrightarrow{\text{GELU}} \mathbb{R}^{512}\\
                      &\xrightarrow{\text{GELU}} \mathbb{R}^{256}
                       \xrightarrow{\sigma}      \mathbb{R}^{1},
\end{aligned}
    \label{eq:fusion}
\end{equation}
where $\mathcal{C}$ is a four-layer MLP with LayerNorm, GELU, and Dropout(0.3).
Concatenation keeps both signals separate: $\mathbf{F}_g$ provides full face context, $\mathbf{F}_f$ provides localised forgery feature.

\noindent\textbf{Training Stability:}
At the start of training, MaskNet scores are random, which makes $\mathbf{F}_f$ noisy and unstable.
To handle this, we blend $\mathbf{F}_f$ toward $\mathbf{F}_g$ during an early warmup phase:
$\tilde{\mathbf{F}}_f = w\,\mathbf{F}_g + (1{-}w)\,\mathbf{F}_f$,
where $w = \max(1{-}e/T_w,\;1{-}0.67)$ decreases over $T_w{=}3$ epochs.
After warmup, $\mathbf{F}_f$ contributes up to 67\% of the fused signal.
This blending applies only during training.
We also maintain an exponential moving average (EMA, $\rho{=}0.999$) of all weights and use the EMA model for validation.

\noindent\textbf{Loss Function:}
The total loss combines binary classification with three mask regularisation terms:
\begin{equation}
    \mathcal{L} = \mathcal{L}_{\text{cls}}
               + 0.1\,\mathcal{L}_{\text{ent}}
               + 0.5\,\mathcal{L}_{\text{real}}
               + 0.3\,\mathcal{L}_{\text{fake}}.
    \label{eq:total_loss}
\end{equation}
$\mathcal{L}_{\text{cls}}$ is binary cross-entropy with label smoothing ($\alpha{=}0.1$).
$\mathcal{L}_{\text{ent}}$ penalises masks that are all-zero or all-one, keeping them informative.
$\mathcal{L}_{\text{real}}$ pushes mask scores toward zero on real images.
$\mathcal{L}_{\text{fake}}$ pushes them toward one on fake images.
Without the last two terms, the mask collapses and $\mathbf{F}_f$ degenerates to a copy of $\mathbf{F}_g$.
\section{Benchmarking and Findings}
\label{sec:experiments}
\noindent\textbf{Implementation Details:}
All models process $224{\times}224$ crops (InsightFace Antelope alignment~\cite{insightface}).
Backbone: CLIP ViT-L/14~\cite{clip}, $N{=}256$ patch tokens, $D{=}1024$; first 12 of 24 blocks frozen, last 12 fine-tuned ($\eta_{\text{enc}}{=}10^{-6}$).
Task heads use $\eta_{\text{heads}}{=}5{\times}10^{-5}$, AdamW, cosine schedule ($T_{\max}{=}30$, $\eta_{\min}{=}10^{-7}$), weight decay $0.05$, gradient clip $1.0$, batch $64$, single NVIDIA A100.
Loss: $\lambda_{\text{ent}}{=}0.1$, $\lambda_{\text{real}}{=}0.5$, $\lambda_{\text{fake}}{=}0.3$, label smoothing $\alpha{=}0.1$.
All detectors are trained exclusively on FF++ (C23)~\cite{ff++} ($720/140/140$ split) and tested with the official checkpoint provided in the paper, with no additional data or tuning. We evaluate on three protocols: \textbf{P\#1}: five classical benchmarks (CelebDF-v1/v2, DFDC, DFDCP, UADFV); \textbf{P\#2}: eight DF40 generation methods~\cite{df40}; \textbf{P\#3}: our RLDF test set (${\sim}286$K images, 8 face-swap methods $\times$ 8 regions).
All cells report ACC\,/\,AUC; weighted means are image-count weighted.

\noindent\textbf{Protocol Results.}
\textbf{P\#1} (Table~\ref{tab:protocol1}): Our 12-Block achieves the highest weighted mean AUC ($0.86$, ACC $0.78$), outperforming all baselines including ForAda~\cite{ForAda}, which leads on DFDC and DFDCP individually (AUC $0.84$/$0.89$) but ranks second overall at $0.87$.
Frequency-domain detectors (LGrad, FreqNet) collapse to near-random performance (AUC ${\approx}0.50$), confirming that artefacts introduced by FF++ C23 compression are dataset-specific and do not transfer across benchmarks.

\textbf{P\#2} (Table~\ref{tab:protocol2}): Our 12-Block leads on weighted ACC ($0.80$), with particularly strong AUC on DeepFaceLab ($0.97$) and HeyGen ($0.90$) generation methods that produce localised blending boundaries well-suited to MCA's patch-level attention.
Mid-Journey6 proves universally challenging (AUC $0.55$), as it synthesises faces from scratch without any identity transfer, leaving no spatial forgery cues for detection.

\textbf{P\#3} (Table~\ref{tab:protocol3}): Unlike P\#1 and P\#2 where the entire face is manipulated, P\#3 restricts forgery to individual facial sub-regions, which makes the detection task progressively harder as the manipulated area shrinks.
Our 8-Block leads all methods (ACC $0.81$, AUC $0.87$).
Notably, it outperforms the 12-Block, in contrast to P\#1 and P\#2, revealing a fundamental trade-off in encoder fine-tuning depth.
The 12-Blocks, fine-tuned more aggressively on full face FF++ forgeries, learns features biased toward holistic manipulation patterns and loses sensitivity to localized patch-level cues.
The 8-Blocks, with the, majority of CLIP layers frozen, retains more general spatial representations that generalize better to partial-face manipulation.
Across methods, E4S is the most detectable (AUC ${\ge}0.96$) owing to its pronounced style-transfer boundaries, while BlendFace is the hardest (AUC $0.85$) as its latent-space blending produces minimal visible seam artifacts. Table~\ref{tab:generalization_summary} consolidates per-protocol weighted means, extended analysis including ROC curves, AUC heatmaps, confusion matrices, t-SNE projections, MaskNet activation maps are provided in Appendices~G and~H.
\begin{figure*}[tb]
    \centering
    \includegraphics[width=1\linewidth]{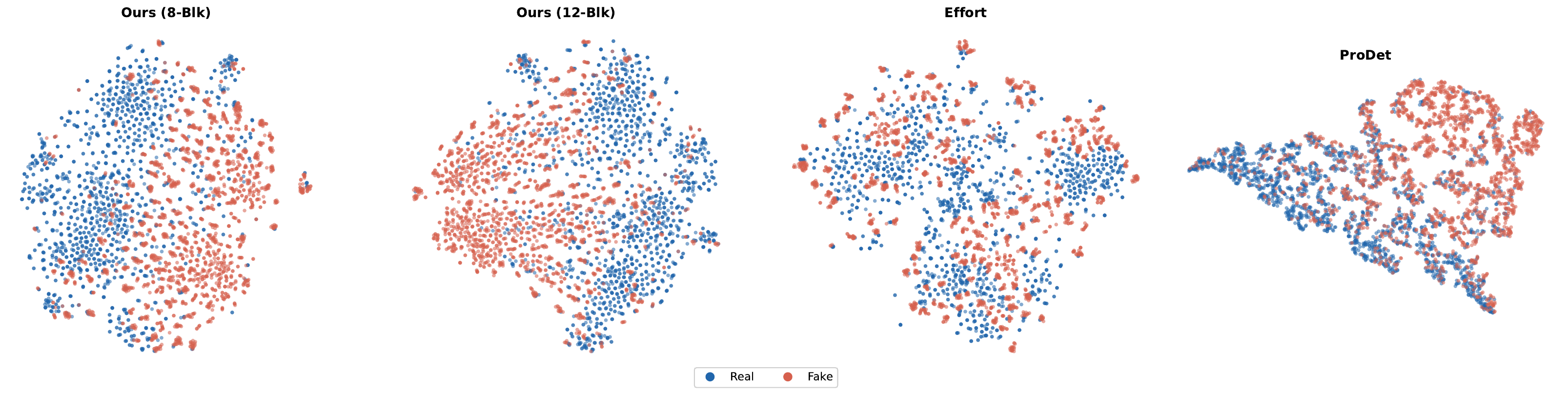}
    \caption{t-SNE visualization of penultimate features on the RLDF test set
    (Protocol~\#3). \textcolor{blue}{Blue} = real, \textcolor{red}{red} = fake.}
    \label{fig:tsne}
\end{figure*}

\noindent\textbf{Finding 1: No existing method generalises across all three protocols.}
ProDet~\cite{prodet} (AUC $0.89$ on P\#3) collapses to $0.57$ on P\#2, FatFormer~\cite{fatformer} ($0.85$ on P\#2) drops to $0.74$ on P\#3, ForAda~\cite{ForAda} leads P\#1 ($0.87$) but falls to $0.73$ on P\#2.
Among all evaluated detectors, our 12-Block baseline consistently maintains strong performance across all three protocols (mean ACC $0.78$, AUC $0.85$), suggesting that explicit spatial grounding via MaskNet and MCA offers a generalization advantage orthogonal to encoder pre-training and frequency analysis.

\noindent\textbf{Finding 2: The three protocols are orthogonal.}
Rank ordering of methods differs substantially across protocols: strong cross-dataset performance on P\#1 does not predict region-level sensitivity on P\#3, confirming that RLDF probes a capability not captured by any existing benchmark.

\noindent\textbf{Finding 3: Region-level evaluation reveals systematic failure.}
Detectors that score above $0.82$ AUC on P\#1 drop by up to $10.5$\% when evaluated on P\#3, validating the RLDF benchmark's role in exposing a critical and previously invisible generalization gap.

Overall, existing detectors fail to generalize consistently across partial facial regions, with an average drop of \textbf{10.5\%} AUC and \textbf{10\%} accuracy under region-level evaluation. This degradation is not uniform, it varies across swap methods and facial regions, with some region method pairs proving significantly harder than others. A full per-region and per-method breakdown is provided in Appendices~G and~H.
\begin{table*}[tb]
\tiny
\centering
\caption{\textbf{Protocol \#1 (Cross-Dataset, Classic Deepfake Benchmarks): Frame-level performance.}
Models are trained on FF++ and tested on five widely-used deepfake benchmarks.
\textbf{Bold} = best; \underline{underline} = second-best (per metric, per column).
Weighted mean is image-count weighted across five datasets
(total 297,914 images; DFDC+DFDCP account for 92.5\% of weight).}
\label{tab:protocol1}
\vspace{1mm}
\setlength{\tabcolsep}{4pt}
\renewcommand{\arraystretch}{1.45}
\resizebox{\linewidth}{!}{%
\begin{tabular}{l c cc cc cc cc cc cc}
\toprule
\textbf{Detector} & \textbf{Venue}
  & \multicolumn{2}{c}{\small\textbf{CelebDF-v1}}
  & \multicolumn{2}{c}{\small\textbf{CelebDF-v2}}
  & \multicolumn{2}{c}{\small\textbf{DFDC}}
  & \multicolumn{2}{c}{\small\textbf{DFDCP}}
  & \multicolumn{2}{c}{\small\textbf{UADFV}}
  & \multicolumn{2}{c}{\shortstack{\small\textbf{Wtd.}\\\small\textbf{Mean}}} \\
\cmidrule(lr){3-4}\cmidrule(lr){5-6}\cmidrule(lr){7-8}\cmidrule(lr){9-10}\cmidrule(lr){11-12}\cmidrule(lr){13-14}
  & & \scriptsize ACC & \scriptsize AUC & \scriptsize ACC & \scriptsize AUC
  & \scriptsize ACC & \scriptsize AUC & \scriptsize ACC & \scriptsize AUC
  & \scriptsize ACC & \scriptsize AUC & \scriptsize ACC & \scriptsize AUC \\
\midrule
\small Xception~\cite{ff++}
  & \small\scriptsize ICCV'19
  & \small 0.69 & \small 0.77 & \small 0.71 & \small 0.74 & \small 0.64 & \small 0.71 & \small 0.68 & \small 0.73 & \small 0.79 & \small 0.94 & \small 0.67 & \small 0.73 \\
\small F3Net~\cite{F3Net}
  & \small\scriptsize ECCV'20
  & \small 0.68 & \small 0.75 & \small 0.61 & \small 0.73 & \small 0.63 & \small 0.70 & \small 0.64 & \small 0.70 & \small 0.81 & \small 0.89 & \small 0.63 & \small 0.71 \\
\small RECCE~\cite{RECCE}
  & \small\scriptsize CVPR'22
  & \small 0.69 & \small 0.75 & \small 0.70 & \small 0.75 & \small 0.62 & \small 0.68 & \small 0.66 & \small 0.70 & \small 0.78 & \small 0.89 & \small 0.64 & \small 0.70 \\
\small SBI~\cite{SBI}
  & \small\scriptsize CVPR'22
  & \small 0.75 & \small 0.81 & \small 0.71 & \small 0.77 & \small 0.62 & \small 0.71 & \small 0.74 & \small 0.83 & \small 0.84 & \small 0.94 & \small 0.69 & \small 0.77 \\
\small LGrad~\cite{LGrad}
  & \small\scriptsize CVPR'23
  & \small 0.39 & \small 0.54 & \small 0.35 & \small 0.53 & \small 0.48 & \small 0.50 & \small 0.35 & \small 0.50 & \small 0.50 & \small 0.51 & \small 0.41 & \small 0.50 \\
\small UCF~\cite{yan2023ucf}
  & \small\scriptsize ICCV'23
  & \small 0.72 & \small 0.81 & \small 0.66 & \small 0.77 & \small 0.66 & \small 0.73 & \small 0.60 & \small 0.69 & \small 0.82 & \small 0.92 & \small 0.63 & \small 0.72 \\
\small FatFormer~\cite{fatformer}
  & \small\scriptsize CVPR'24
  & \small 0.40 & \small 0.74 & \small 0.36 & \small 0.60 & \small 0.56 & \small 0.63 & \small 0.59 & \small 0.59 & \small 0.60 & \small 0.78 & \small 0.56 & \small 0.61 \\
\small FreqNet~\cite{FreqNet}
  & \small\scriptsize AAAI'24
  & \small 0.44 & \small 0.44 & \small 0.38 & \small 0.56 & \small 0.48 & \small 0.51 & \small 0.40 & \small 0.53 & \small 0.45 & \small 0.45 & \small 0.43 & \small 0.52 \\
\small ProDet~\cite{prodet}
  & \small\scriptsize NeurIPS'24
  & \small 0.77 & \small 0.86 & \small 0.77 & \small 0.83 & \small 0.64 & \small 0.70 & \small 0.72 & \small 0.77 & \small 0.82 & \small 0.95 & \small 0.69 & \small 0.75 \\
\small DFD-FCG~\cite{dfd_fcg}
  & \small\scriptsize CVPR'25
  & \small 0.70 & \small 0.89 & \small 0.74 & \small 0.87 & \small 0.65 & \small 0.80 & \small 0.73 & \small 0.83 & \small 0.66 & \small 0.98 & \small 0.69 & \small 0.82 \\
\small Effort~\cite{effort}
  & \small\scriptsize ICML'25
  & \small 0.79 & \small 0.92 & \small 0.80 & \small 0.87 & \small 0.74 & \small 0.82 & \small 0.72 & \small 0.84 & \small 0.83 & \small 0.97 & \small 0.74 & \small 0.83 \\
\small ForAda~\cite{ForAda}
  & \small\scriptsize CVPR'25
  & \small 0.82 & \small 0.91 & \small 0.82 & \small\underline{0.90} & \small 0.75 & \small\textbf{0.84} & \small 0.73 & \small\textbf{0.89} & \small 0.89 & \small\underline{0.98} & \small\underline{0.75} & \small\textbf{0.87} \\
\midrule
\rowcolor{rldfblue}
\small Ours (8-Blocks)
  & \small —
  & \small 0.83 & \small\underline{0.92} & \small 0.80 & \small 0.89 & \small 0.73 & \small 0.81 & \small 0.74 & \small 0.81 & \small 0.89 & \small 0.98 & \small 0.74 & \small 0.82 \\
\rowcolor{rldfblue}
\small\textbf{Ours (12-Blocks)}
  & \small —
  & \small\textbf{0.84} & \small\textbf{0.93} & \small\textbf{0.84} & \small\textbf{0.92} & \small\textbf{0.75} & \small\underline{0.84} & \small\textbf{0.79} & \small\underline{0.86} & \small\textbf{0.90} & \small\textbf{0.98} & \small\textbf{0.78} & \small\underline{0.86} \\
\bottomrule
\end{tabular}}
\end{table*}

\begin{figure*}[tb]
    \centering
    \begin{minipage}[t]{0.32\textwidth}
        \centering
        \includegraphics[width=\linewidth]{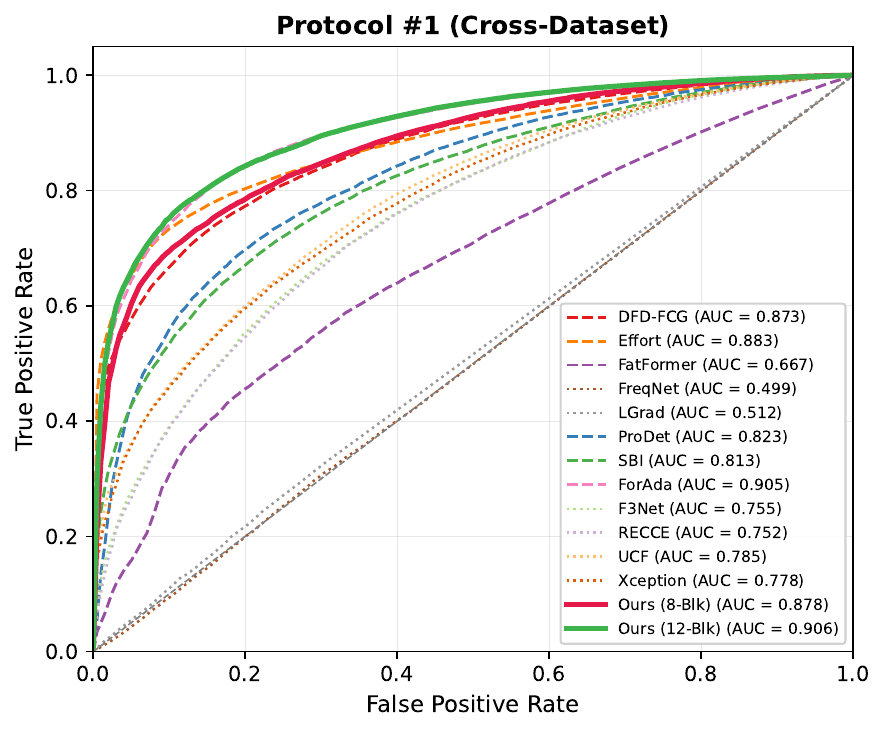}
        \small{(a) Protocol \#1}
    \end{minipage}
    \hfill
    \begin{minipage}[t]{0.32\textwidth}
        \centering        \includegraphics[width=\linewidth]{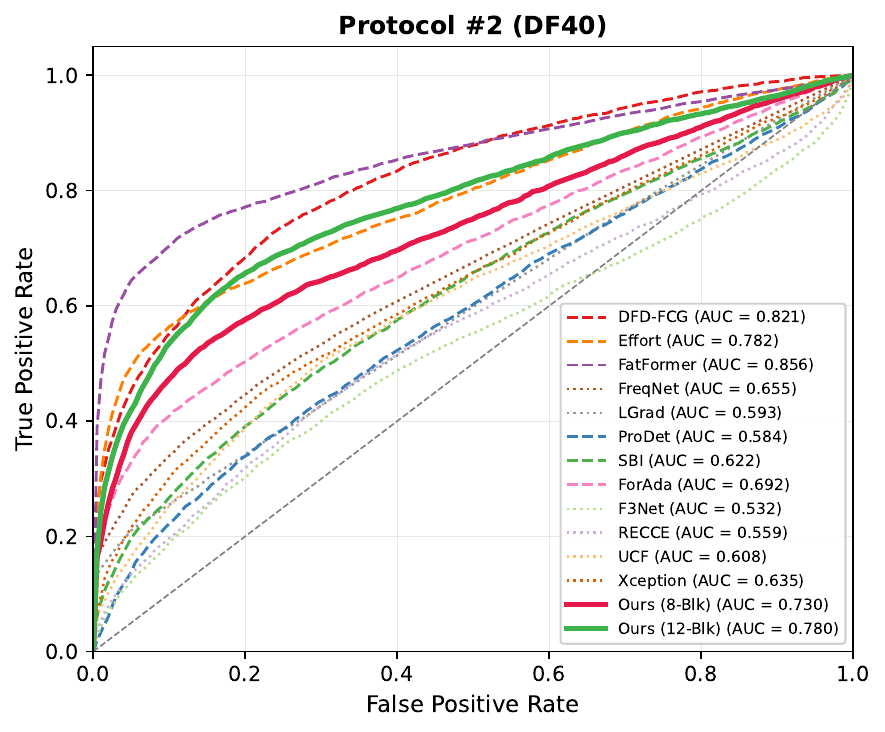}
        \small{(b) Protocol \#2}
    \end{minipage}
    \hfill
    \begin{minipage}[t]{0.32\textwidth}
        \centering
        \includegraphics[width=\linewidth]{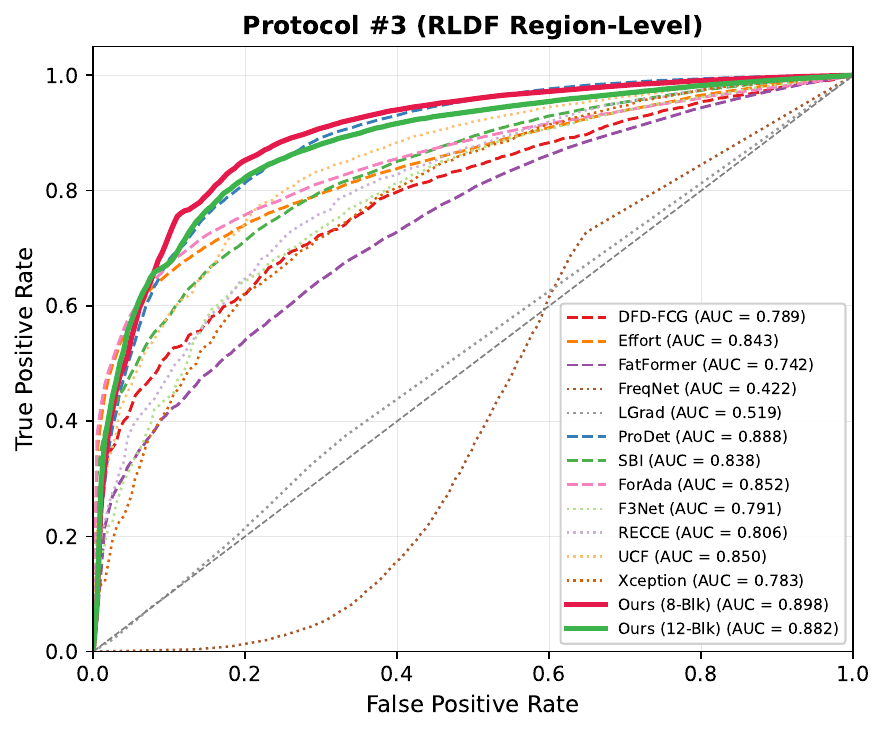}
        \small{(c) Protocol \#3}
    \end{minipage}
   \caption{ROC curves across all three evaluation protocols. Our baseline (8-Blk, 12-Blk) 
achieves competitive AUC across all protocols (P\protect\#1: $0.906$, P\protect\#2: $0.788$, 
P\protect\#3: $0.88$), while existing detectors show inconsistent performance under 
region-level evaluation.}
    \label{fig:roc_all}
\end{figure*}
\begin{table*}[tb]
\centering
\caption{\textbf{Protocol \#2 (Cross-Dataset, DF40 Test Set): Frame-level performance.}
Models are trained on FF++ and tested on eight generation methods from DF40.
\textbf{Bold} = best; \underline{underline} = second-best (per metric, per column).
Weighted mean is image-count weighted across the eight sub-datasets (total 31,238 images).}
\label{tab:protocol2}
\vspace{2mm}
\setlength{\tabcolsep}{4pt}
\renewcommand{\arraystretch}{1.45}
\resizebox{\linewidth}{!}{%
\begin{tabular}{l c cc cc cc cc cc cc cc cc cc cc}
\toprule
\textbf{Detector} & \textbf{Venue}
  & \multicolumn{2}{c}{\shortstack{\small\textbf{Deep-}\\\small\textbf{FaceLab}}}
  & \multicolumn{2}{c}{\shortstack{\small\textbf{Mid-}\\\small\textbf{Journey6}}}
  & \multicolumn{2}{c}{\shortstack{\small\textbf{Hey-}\\\small\textbf{Gen}}}
  & \multicolumn{2}{c}{\shortstack{\small\textbf{Star-}\\\small\textbf{GAN}}}
  & \multicolumn{2}{c}{\shortstack{\small\textbf{Star-}\\\small\textbf{GAN2}}}
  & \multicolumn{2}{c}{\shortstack{\small\textbf{Style-}\\\small\textbf{CLIP}}}
  & \multicolumn{2}{c}{\shortstack{\small\textbf{Collab-}\\\small\textbf{Diff}}}
  & \multicolumn{2}{c}{\shortstack{\small\textbf{Which-}\\\small\textbf{FaceReal}}}
  & \multicolumn{2}{c}{\shortstack{\small\textbf{Wtd.}\\\small\textbf{Mean}}} \\
\cmidrule(lr){3-4}\cmidrule(lr){5-6}\cmidrule(lr){7-8}\cmidrule(lr){9-10}
\cmidrule(lr){11-12}\cmidrule(lr){13-14}\cmidrule(lr){15-16}\cmidrule(lr){17-18}\cmidrule(lr){19-20}
  & & \small ACC & \small AUC & \small ACC & \small AUC & \small ACC & \small AUC
  & \small ACC & \small AUC & \small ACC & \small AUC & \small ACC & \small AUC
  & \small ACC & \small AUC & \small ACC & \small AUC & \small ACC & \small AUC \\
\midrule
Xception~\cite{ff++}
  & \small\scriptsize ICCV'19
  & 0.66 & 0.72 & 0.62 & 0.57 & 0.79 & 0.77
  & 0.57 & 0.62 & 0.48 & 0.42 & 0.65 & 0.78
  & 0.69 & 0.75 & 0.47 & 0.45 & 0.65 & 0.67 \\
F3Net~\cite{F3Net}
  & \small\scriptsize ECCV'20
  & 0.58 & 0.71 & 0.72 & 0.78 & 0.50 & 0.50
  & 0.59 & 0.67 & 0.48 & 0.39 & 0.42 & 0.44
  & 0.54 & 0.42 & 0.44 & 0.34 & 0.52 & 0.53 \\
RECCE~\cite{RECCE}
  & \small\scriptsize CVPR'22
  & 0.57 & 0.64 & 0.68 & 0.72 & 0.71 & 0.55
  & 0.55 & 0.61 & 0.48 & 0.38 & 0.50 & 0.61
  & 0.60 & 0.56 & 0.47 & 0.40 & 0.59 & 0.56 \\
SBI~\cite{SBI}
  & \small\scriptsize CVPR'22
  & 0.72 & 0.80 & 0.55 & 0.57 & 0.67 & 0.71
  & 0.71 & 0.81 & 0.52 & 0.52 & 0.67 & 0.68
  & 0.53 & 0.55 & 0.40 & 0.34 & 0.63 & 0.67 \\
LGrad~\cite{LGrad}
  & \small\scriptsize CVPR'23
  & 0.33 & 0.48 & 0.99 & \underline{1.00} & 0.75 & 0.44
  & 0.50 & 0.53 & 0.50 & 0.49 & 0.26 & 0.46
  & 0.51 & 0.61 & 0.67 & 0.76 & 0.56 & 0.53 \\
UCF~\cite{yan2023ucf}
  & \small\scriptsize ICCV'23
  & 0.64 & 0.74 & 0.70 & 0.76 & 0.61 & 0.73
  & 0.61 & 0.70 & 0.49 & 0.44 & 0.51 & 0.62
  & 0.60 & 0.58 & 0.47 & 0.41 & 0.58 & 0.65 \\
FatFormer~\cite{fatformer}
  & \small\scriptsize CVPR'24
  & 0.35 & 0.53 & 0.95 & 0.85 & 0.86 & \textbf{0.94}
  & 0.98 & \textbf{1.00} & 0.68 & \underline{0.82} & 0.79 & \textbf{0.89}
  & 0.68 & 0.75 & 0.87 & \textbf{0.98} & 0.76 & \underline{0.85} \\
FreqNet~\cite{FreqNet}
  & \small\scriptsize AAAI'24
  & 0.32 & 0.42 & 0.97 & \textbf{1.00} & 0.86 & 0.71
  & 0.50 & 0.50 & 0.50 & 0.50 & 0.26 & 0.49
  & 0.61 & \underline{0.85} & 0.71 & \underline{0.79} & 0.59 & 0.61 \\
ProDet~\cite{prodet}
  & \small\scriptsize NeurIPS'24
  & 0.73 & 0.77 & 0.54 & 0.64 & 0.31 & 0.44
  & 0.58 & 0.63 & 0.48 & 0.47 & 0.68 & 0.66
  & 0.52 & 0.61 & 0.47 & 0.44 & 0.51 & 0.57 \\
DFD-FCG~\cite{dfd_fcg}
  & \small\scriptsize CVPR'25
  & 0.76 & 0.90 & 0.75 & 0.85 & 0.59 & 0.89
  & 0.74 & 0.97 & 0.61 & 0.71 & 0.81 & 0.86
  & 0.72 & \textbf{0.86} & 0.53 & 0.53 & 0.68 & \textbf{0.85} \\
Effort~\cite{effort}
  & \small\scriptsize ICML'25
  & 0.84 & 0.93 & 0.58 & 0.57 & 0.81 & 0.80
  & 0.83 & \underline{0.99} & 0.79 & \textbf{0.85} & 0.82 & \underline{0.88}
  & 0.65 & 0.71 & 0.50 & 0.52 & 0.77 & 0.82 \\
ForAda~\cite{ForAda}
  & \small\scriptsize CVPR'25
  & 0.85 & 0.92 & 0.57 & 0.46 & 0.87 & 0.68
  & 0.89 & 0.95 & 0.66 & 0.73 & 0.52 & 0.69
  & 0.57 & 0.64 & 0.48 & 0.47 & 0.74 & 0.73 \\
\midrule
\rowcolor{rldfblue}
Ours (8-Blocks)
  & —
  & 0.88 & \underline{0.94} & 0.54 & 0.41 & 0.87 & 0.85
  & 0.83 & 0.92 & 0.70 & 0.76 & 0.75 & 0.81
  & 0.65 & 0.69 & 0.55 & 0.47 & \underline{0.78} & 0.80 \\
\rowcolor{rldfblue}
\textbf{Ours (12-Blocks)}
  & —
  & 0.92 & \textbf{0.97} & 0.60 & 0.55 & 0.89 & \underline{0.90}
  & 0.84 & 0.93 & 0.72 & 0.78 & 0.80 & 0.86
  & 0.68 & 0.73 & 0.56 & 0.52 & \textbf{0.80} & 0.84 \\
\bottomrule
\end{tabular}}
\end{table*}

\vspace{-4mm}
\begin{table*}[tb]
\centering
\caption{\textbf{Protocol \#3 (RLDF Region-Level Test Set): Frame-level performance.}
Models are trained on FF++ and evaluated on our proposed RLDF benchmark across eight face-swap methods.
\textbf{Bold} = best; \underline{underline} = second-best (per metric, per column).
Weighted mean is image-count weighted (total 286,456 images;
counts are near-uniform so weighting is effectively equal).}
\label{tab:protocol3}
\vspace{2mm}
\setlength{\tabcolsep}{3pt}
\renewcommand{\arraystretch}{1.45}
\resizebox{\linewidth}{!}{%
\begin{tabular}{l c cc cc cc cc cc cc cc cc cc cc}
\toprule
\textbf{Detector} & \textbf{Venue}
  & \multicolumn{2}{c}{\small\textbf{BlendFace}}
  & \multicolumn{2}{c}{\small\textbf{E4S}}
  & \multicolumn{2}{c}{\small\textbf{FaceDancer}}
  & \multicolumn{2}{c}{\small\textbf{FSGAN}}
  & \multicolumn{2}{c}{\small\textbf{Ghost}}
  & \multicolumn{2}{c}{\small\textbf{InSwapper}}
  & \multicolumn{2}{c}{\small\textbf{SimSwap}}
  & \multicolumn{2}{c}{\small\textbf{UniFace}}
  & \multicolumn{2}{c}{\small\textbf{RLDF: All}}
  & \multicolumn{2}{c}{\shortstack{\small\textbf{Wtd.}\\\small\textbf{Mean}}} \\
\cmidrule(lr){3-4}\cmidrule(lr){5-6}\cmidrule(lr){7-8}\cmidrule(lr){9-10}\cmidrule(lr){11-12}
\cmidrule(lr){13-14}\cmidrule(lr){15-16}\cmidrule(lr){17-18}\cmidrule(lr){19-20}\cmidrule(lr){21-22}
  & & \small ACC & \small AUC & \small ACC & \small AUC & \small ACC & \small AUC
  & \small ACC & \small AUC & \small ACC & \small AUC & \small ACC & \small AUC
  & \small ACC & \small AUC & \small ACC & \small AUC & \small ACC & \small AUC
  & \small ACC & \small AUC \\
\midrule
Xception~\cite{ff++}
  & \small\scriptsize ICCV'19
  & 0.66 & 0.73 & 0.70 & 0.78 & 0.69 & 0.76 & 0.80 & 0.88 & 0.69 & 0.77
  & 0.79 & 0.87 & 0.66 & 0.74 & 0.68 & 0.74 & 0.71 & 0.78 & 0.71 & 0.78 \\
F3Net~\cite{F3Net}
  & \small\scriptsize ECCV'20
  & 0.59 & 0.69 & 0.69 & 0.79 & 0.61 & 0.72 & 0.81 & 0.92 & 0.66 & 0.80
  & 0.84 & \underline{0.93} & 0.63 & 0.77 & 0.59 & 0.69 & 0.68 & 0.79 & 0.68 & 0.79 \\
RECCE~\cite{RECCE}
  & \small\scriptsize CVPR'22
  & 0.62 & 0.71 & 0.75 & 0.84 & 0.66 & 0.75 & 0.82 & 0.91 & 0.70 & 0.80
  & 0.84 & 0.92 & 0.67 & 0.77 & 0.64 & 0.73 & 0.71 & 0.80 & 0.71 & 0.81 \\
SBI~\cite{SBI}
  & \small\scriptsize CVPR'22
  & 0.70 & 0.81 & 0.93 & 0.98 & 0.62 & 0.74 & 0.77 & 0.88 & 0.70 & 0.79
  & 0.81 & 0.89 & 0.67 & 0.78 & 0.71 & 0.82 & 0.74 & 0.84 & 0.74 & 0.84 \\
LGrad~\cite{LGrad}
  & \small\scriptsize CVPR'23
  & 0.50 & 0.54 & 0.50 & 0.57 & 0.50 & 0.53 & 0.49 & 0.48 & 0.50 & 0.57
  & 0.50 & 0.59 & 0.50 & 0.58 & 0.50 & 0.53 & 0.50 & 0.55 & 0.50 & 0.55 \\
UCF~\cite{yan2023ucf}
  & \small\scriptsize ICCV'23
  & 0.64 & 0.73 & 0.82 & 0.90 & 0.68 & 0.78 & 0.87 & \underline{0.95} & 0.77 & \underline{0.87}
  & \textbf{0.89} & \textbf{0.96} & \textbf{0.74} & \textbf{0.85} & 0.64 & 0.73 & 0.75 & 0.84 & 0.76 & 0.85 \\
FatFormer~\cite{fatformer}
  & \small\scriptsize CVPR'24
  & 0.53 & 0.61 & 0.96 & 0.98 & 0.57 & 0.72 & 0.57 & 0.70 & 0.57 & 0.76
  & 0.55 & 0.75 & 0.56 & 0.72 & 0.56 & 0.71 & 0.61 & 0.75 & 0.61 & 0.74 \\
FreqNet~\cite{FreqNet}
  & \small\scriptsize AAAI'24
  & 0.38 & 0.45 & 0.38 & 0.46 & 0.39 & 0.47 & 0.32 & 0.36 & 0.38 & 0.46
  & 0.38 & 0.47 & 0.38 & 0.46 & 0.39 & 0.46 & 0.38 & 0.45 & 0.38 & 0.45 \\
ProDet~\cite{prodet}
  & \small\scriptsize NeurIPS'24
  & \textbf{0.81} & \textbf{0.89} & 0.87 & 0.97 & 0.74 & 0.82 & \textbf{0.86} & \textbf{0.95} & 0.78 & 0.86
  & 0.79 & 0.86 & 0.77 & \underline{0.84} & \textbf{0.79} & \underline{0.88} & \textbf{0.80} & \textbf{0.88} & \underline{0.80} & \textbf{0.89} \\
DFD-FCG~\cite{dfd_fcg}
  & \small\scriptsize CVPR'25
  & 0.61 & 0.67 & 0.71 & 0.98 & 0.66 & 0.79 & 0.69 & 0.87 & 0.63 & 0.74
  & 0.66 & 0.80 & 0.57 & 0.62 & 0.65 & 0.78 & 0.67 & 0.81 & 0.65 & 0.78 \\
Effort~\cite{effort}
  & \small\scriptsize ICML'25
  & 0.62 & 0.67 & 0.89 & \underline{0.99} & 0.77 & 0.84 & 0.84 & 0.93 & 0.78 & 0.85
  & 0.80 & 0.88 & 0.68 & 0.74 & 0.76 & 0.83 & 0.76 & 0.84 & 0.77 & 0.84 \\
ForAda~\cite{ForAda}
  & \small\scriptsize CVPR'25
  & 0.68 & 0.75 & \textbf{0.92} & \textbf{0.99} & 0.79 & 0.86 & 0.87 & 0.94 & 0.74 & 0.81
  & 0.80 & 0.86 & 0.66 & 0.72 & 0.78 & 0.85 & 0.78 & 0.85 & 0.78 & 0.85 \\
\midrule
\rowcolor{rldfblue}
\textbf{Ours (8-Blocks)}
  & —
  & \underline{0.75} & \underline{0.85} & 0.92 & 0.96 & \textbf{0.81} & \textbf{0.90} & 0.84 & 0.91 & \textbf{0.80} & \textbf{0.88}
  & 0.80 & 0.88 & 0.80 & 0.70 & 0.79 & \textbf{0.88} & \underline{0.80} & \underline{0.88} & \textbf{0.81} & \underline{0.87} \\
\rowcolor{rldfblue}
Ours (12-Blocks)
  & —
  & 0.69 & 0.82 & 0.92 & 0.97 & \underline{0.79} & \underline{0.89} & 0.81 & 0.90 & 0.74 & 0.85
  & 0.74 & 0.85 & 0.63 & 0.75 & 0.77 & 0.88 & 0.76 & 0.86 & 0.76 & 0.86 \\
\bottomrule
\end{tabular}}
\end{table*}

\enlargethispage{8mm}
\vspace{-4mm}
\begin{table*}[tb]

\centering
\tiny
\caption{\textbf{Cross-Protocol Generalization.}
Our detector is the only method maintaining strong performance across all three regimes.
\textbf{Bold} = best, \underline{underline} = second-best (per metric, per column).}
\label{tab:generalization_summary}
\vspace{1mm}
\setlength{\tabcolsep}{5pt}
\renewcommand{\arraystretch}{1.45}
\resizebox{0.78\linewidth}{!}{%
\begin{tabular}{l c cc cc cc cc}
\toprule
\textbf{Detector} & \textbf{Venue}
  & \multicolumn{2}{c}{\small\textbf{P\#1 (Classic)}}
  & \multicolumn{2}{c}{\small\textbf{P\#2 (DF40)}}
  & \multicolumn{2}{c}{\small\textbf{P\#3 (RLDF)}}
  & \multicolumn{2}{c}{\small\textbf{Mean}} \\
\cmidrule(lr){3-4}\cmidrule(lr){5-6}\cmidrule(lr){7-8}\cmidrule(lr){9-10}
  & & \scriptsize ACC & \scriptsize AUC & \scriptsize ACC & \scriptsize AUC
  & \scriptsize ACC & \scriptsize AUC & \scriptsize ACC & \scriptsize AUC \\
\midrule
\small Xception~\cite{ff++}
  & \small\scriptsize ICCV'19
  & \small 0.67 & \small 0.73 & \small 0.65 & \small 0.67 & \small 0.71 & \small 0.78 & \small 0.68 & \small 0.73 \\
\small F3Net~\cite{F3Net}
  & \small\scriptsize ECCV'20
  & \small 0.63 & \small 0.71 & \small 0.52 & \small 0.53 & \small 0.68 & \small 0.79 & \small 0.61 & \small 0.68 \\
\small RECCE~\cite{RECCE}
  & \small\scriptsize CVPR'22
  & \small 0.64 & \small 0.70 & \small 0.59 & \small 0.56 & \small 0.71 & \small 0.81 & \small 0.65 & \small 0.69 \\
\small SBI~\cite{SBI}
  & \small\scriptsize CVPR'22
  & \small 0.69 & \small 0.77 & \small 0.63 & \small 0.67 & \small 0.74 & \small 0.84 & \small 0.69 & \small 0.76 \\
\small LGrad~\cite{LGrad}
  & \small\scriptsize CVPR'23
  & \small 0.41 & \small 0.50 & \small 0.56 & \small 0.53 & \small 0.50 & \small 0.55 & \small 0.49 & \small 0.53 \\
\small UCF~\cite{yan2023ucf}
  & \small\scriptsize ICCV'23
  & \small 0.63 & \small 0.72 & \small 0.58 & \small 0.65 & \small 0.76 & \small 0.85 & \small 0.66 & \small 0.74 \\
\small FatFormer~\cite{fatformer}
  & \small\scriptsize CVPR'24
  & \small 0.56 & \small 0.61 & \small 0.76 & \small\underline{0.85} & \small 0.61 & \small 0.74 & \small 0.64 & \small 0.73 \\
\small FreqNet~\cite{FreqNet}
  & \small\scriptsize AAAI'24
  & \small 0.43 & \small 0.52 & \small 0.59 & \small 0.61 & \small 0.38 & \small 0.45 & \small 0.47 & \small 0.53 \\
\small ProDet~\cite{prodet}
  & \small\scriptsize NeurIPS'24
  & \small 0.69 & \small 0.75 & \small 0.51 & \small 0.57 & \small\underline{0.80} & \small\textbf{0.89} & \small 0.67 & \small 0.74 \\
\small DFD-FCG~\cite{dfd_fcg}
  & \small\scriptsize CVPR'25
  & \small 0.69 & \small 0.82 & \small 0.68 & \small\textbf{0.85} & \small 0.65 & \small 0.78 & \small 0.67 & \small 0.82 \\
\small Effort~\cite{effort}
  & \small\scriptsize ICML'25
  & \small 0.74 & \small 0.83 & \small 0.77 & \small 0.82 & \small 0.77 & \small 0.84 & \small 0.76 & \small\underline{0.83} \\
\small ForAda~\cite{ForAda}
  & \small\scriptsize CVPR'25
  & \small\underline{0.75} & \small\textbf{0.87} & \small 0.74 & \small 0.73 & \small 0.78 & \small 0.85 & \small 0.76 & \small 0.82 \\
\midrule
\rowcolor{rldfblue}
\small Ours (8-Blocks)
  & \small —
  & \small 0.74 & \small 0.82 & \small\underline{0.78} & \small 0.80 & \small\textbf{0.81} & \small\underline{0.87} & \small\underline{0.78} & \small 0.83 \\
\rowcolor{rldfblue}
\small\textbf{Ours (12-Blocks)}
  & \small —
  & \small\textbf{0.78} & \small\underline{0.86} & \small\textbf{0.80} & \small 0.84 & \small 0.76 & \small 0.86 & \small\textbf{0.78} & \small\textbf{0.85} \\
\bottomrule
\end{tabular}}
\end{table*}

\subsection{Ablation Study}
\label{sec:ablation}
Table~\ref{tab:ablation} evaluates each component by removing it while keeping the rest fixed.
Removing \textbf{mask guidance} or the \textbf{forged path} $\mathbf{F}_f$ has negligible impact on P\#1 (${\le}0.001$ ACC) but causes the largest drops on P\#3 ($0.155$ and $0.157$ ACC), confirming that detecting region-level forgery demands spatial localization, not just a global feature. Removing the \textbf{global path} $\mathbf{F}_g$ hurts P\#2 most ($0.061$ AUC), where generators like Mind-Journey6 leave no local forgery cue and whole-face context is the only reliable signal. Replacing the \textbf{soft mask} with a hard binary threshold drops P\#3 by $0.113$ ACC, as binarisation severs the gradient between MCA and MaskNet, preventing them from jointly refining spatial localization. These results validate all four design choices, each targeting a distinct failure mode, and removing any one degrades performance on at least one protocol.
\vspace{-2mm}
\begin{table*}[tb]
\centering
\caption{\textbf{Ablation Study.}
Component-wise analysis of the proposed Mask-Guided Cross-Attention (MCA) framework
across three evaluation protocols: P1 (classic cross-dataset), P2 (DF40), and
P3 (RLDF region-level benchmark). $\Delta$ denotes the difference from the full
model (negative indicates performance drop).}
\label{tab:ablation}
\small
\setlength{\tabcolsep}{5pt}
\renewcommand{\arraystretch}{1.35}
\resizebox{\linewidth}{!}{%
\begin{tabular}{l ccc ccc ccc ccc}
\toprule
\multirow{3}{*}{\textbf{Variant}}
  & \multicolumn{6}{c}{\textbf{Absolute Performance}}
  & \multicolumn{6}{c}{\textbf{$\Delta$ vs Full Model}} \\
\cmidrule(lr){2-7}\cmidrule(lr){8-13}
  & \multicolumn{2}{c}{\textbf{P1 (Classic)}}
  & \multicolumn{2}{c}{\textbf{P2 (DF40)}}
  & \multicolumn{2}{c}{\textbf{P3 (RLDF)}}
  & \multicolumn{2}{c}{\textbf{P1 (Classic)}}
  & \multicolumn{2}{c}{\textbf{P2 (DF40)}}
  & \multicolumn{2}{c}{\textbf{P3 (RLDF)}} \\
\cmidrule(lr){2-3}\cmidrule(lr){4-5}\cmidrule(lr){6-7}
\cmidrule(lr){8-9}\cmidrule(lr){10-11}\cmidrule(lr){12-13}
  & ACC & AUC & ACC & AUC & ACC & AUC
  & $\Delta$ACC & $\Delta$AUC
  & $\Delta$ACC & $\Delta$AUC
  & $\Delta$ACC & $\Delta$AUC \\
  \midrule
w/o mask guidance
  & 0.772 & 0.852 & 0.749 & 0.780 & 0.607 & 0.850
  & $+$0.001 & $+$0.000
  & $-$0.051 & $-$0.060
  & $\mathbf{-0.155}$ & $-$0.011 \\
w/o forged path ($\mathbf{F}_f{=}\mathbf{F}_g$)
  & 0.771 & 0.851 & 0.748 & 0.774 & 0.605 & 0.850
  & $+$0.001 & $-$0.001
  & $-$0.052 & $-$0.066
  & $\mathbf{-0.157}$ & $-$0.012 \\
w/o global path ($\mathbf{F}_g{=}\mathbf{F}_f$)
  & 0.770 & 0.850 & 0.749 & 0.779 & 0.617 & 0.850
  & $-$0.001 & $-$0.001
  & $-$0.051 & $-$0.061
  & $-$0.144 & $-$0.011 \\
Hard mask (binary $\hat{\mathbf{m}}$)
  & 0.764 & 0.851 & 0.748 & 0.780 & 0.648 & 0.851
  & $-$0.007 & $-$0.001
  & $-$0.052 & $-$0.060
  & $-$0.113 & $-$0.011 \\
  \midrule
  \rowcolor{rldfblue}
  Full model (baseline)
  & 0.771 & 0.852 & 0.800 & 0.840 & 0.761 & 0.861
  & — & — & — & — & — & — \\
\midrule
\end{tabular}}
\end{table*}
\vspace{-8mm}
\section{Limitations and Future Work}
RLDF is deliberately scoped to face swapping: all eight generators transfer
identity, so region-level behaviour under reenactment, entire-face synthesis and
attribute editing remains open. Partial fakes are produced by compositing a region
of a full swap back onto the target through a smoothed parsing mask rather than by
region-native generation, so the resulting boundaries reflect both the generator and
the compositing step, and the region masks inherit any errors of the face parser.
Evaluation is frame-level, leaving the temporal cues available in video unused by
both the benchmark and the MCA baseline. The perceptual study is indicative rather
than conclusive, covering 14 images rated by 37 non-specialist participants.
Extending RLDF to reenactment methods and to video-level temporal analysis are
therefore the most immediate directions for future work.

\section{Conclusion}
We presented \textbf{RLDF}, the first dataset and benchmark for evaluating deepfake detectors at the region level. RLDF contains over 2M images produced by 8 face-swapping methods applied to 8 distinct facial regions, built using a \textbf{region-aware partial swap synthesis}. We evaluated 12 detectors across three protocols. First finding is that, no single detector performs well across all three protocols, indicating that existing methods are not generalizable when the evaluation paradigm change. Second, the three protocols are orthogonal and perform well on classic or DF40 benchmarks, but exhibit a significant performance drop when evaluated at the region level. Third, existing detectors fail to generalize consistently across partial facial regions. Performance drops by up to \textbf{10.5\%} AUC and \textbf{10\%} accuracy when the manipulated area shrinks from full-face to a single region. To provide a starting point, we propose a \textbf{mask-guided cross-attention (MCA)} baseline that grounds forgery detection at the patch level. Our ablation shows that spatial localization, not just global image-level features, drives gains in region-level evaluation.

\noindent\textbf{Ethical considerations.}
RLDF is built from FaceForensics++~\cite{ff++} real videos under its academic
research licence and is released strictly for non-commercial detection research;
no new identities were collected. Users must respect that licence and those of the
eight swapping methods used, and must not employ the data for impersonation or any
purpose harming the privacy or reputation of the source individuals (Appendix~E).
Data, code and checkpoints are released under the same non-commercial terms.

\clearpage
{
    \small
    \bibliographystyle{ieeenat_fullname}
    \bibliography{rldf_references}
}

% ===============================================================
% Supplementary material, appended for the arXiv version.
% Single-column, as originally laid out; floats renumbered S1, S2, ...
% so they do not clash with the main paper's numbering.
% ===============================================================
\clearpage
\onecolumn
\setcounter{figure}{0}
\setcounter{table}{0}
\setcounter{equation}{0}
\renewcommand{\thefigure}{S\arabic{figure}}
\renewcommand{\thetable}{S\arabic{table}}
\renewcommand{\theequation}{S\arabic{equation}}

% Reproduces the visual header of wacv.sty's \maketitlesupplementary,
% but in one-column so the appendix figures keep their original measure.
\begin{center}
  {\Large\bfseries RLDF: A Region-Level Deepfake Dataset and Benchmark for\\
   Partial-Face Manipulation Detection}\\[0.5em]
  {\Large Supplementary Material}\\[1.0em]
\end{center}

\vspace{1mm}
\subsection*{A.\; Content Overview}
\vspace{1mm}
{\small
This appendix provides supplementary material for \textbf{RLDF}. Due to space constraints in the main
paper, detailed results, visualisations, and analyses are presented here. The content is organised as follows:

\begin{itemize}[leftmargin=2.2em, itemsep=2pt, topsep=3pt, parsep=1pt, partopsep=0pt]
  \item \textbf{Section B: Datasheet for RLDF.}
    Full documentation of the dataset following \emph{Datasheets for Datasets},
    covering motivation, composition, collection process, preprocessing, recommended
    and discouraged uses, distribution terms, and maintenance.

  \item \textbf{Section C: Face-Swapping Method Selection.}
    Criteria and characteristics of the eight face-swapping methods used in RLDF construction,
    covering technical paradigm, publication venue, and visual artefact profile.

         \item \textbf{Section D: Human Perceptual Realism Study.}
    A 37-participant survey rating 14 region-level deepfake failure cases on a 0 to 5 fakeness scale,
    confirming that partial-face forgeries are perceptually ambiguous even to non-specialist humans.
    
  \item \textbf{Section E: Ethical Statement.}
    Usage constraints, licence compliance, and data provenance for RLDF.
    
  \item \textbf{Section F: Dataset Visualisations.}
    Representative partial-face swap samples across all eight methods and
    seven manipulation regions (Eye, Nose, Mouth, Composite-1/2/3, Full Face),
    illustrating the region isolation pipeline and soft region masks.

  \item \textbf{Section G: Region-Level Performance Breakdown.}
    Comprehensive quantitative analysis under Protocol~\#3, including:
    \begin{itemize}[leftmargin=1.8em, itemsep=1pt, topsep=1pt, parsep=0pt, partopsep=0pt]
      \item ROC curves averaged per region and per swap method;
      \item Per-detector AUC heatmaps across all method$\times$region pairs;
      \item Confusion matrices analysing false-positive and false-negative patterns;
      \item t-SNE feature space projections per detector and per region.
    \end{itemize}

  \item \textbf{Section H: MaskNet Spatial Analysis and Score Distributions.}
    Analysis of MaskNet's emergent spatial attention including:
    \begin{itemize}[leftmargin=1.8em, itemsep=1pt, topsep=1pt, parsep=0pt, partopsep=0pt]
      \item Spatial activation maps for BlendFace across all 7 regions
            (validating unsupervised spatial grounding);
      \item Activation maps for Eye-region fakes across all 8 swap methods
            (showing intensity correlates with detectability).
      \item Cross-dataset activation maps on CelebDF, DFDC, DFDCP, and UADFV
            (demonstrating dataset-agnostic generalization).
      \item Real/fake score distribution analysis across regions and detectors.
    \end{itemize}

  \item \textbf{Section I: Per-Method Region-Level Analysis (Protocol~\#3).}
    Complete ACC\,/\,AUC tables for all 13 detectors across seven manipulation
    regions, reported separately for each of the eight face-swapping methods.

  \item \textbf{Section J: Qualitative Failure Cases.}
    False-negative examples (fake images misclassified as real) from
    Ours~(12-Blk), organised by manipulation region and sorted by
    detection confidence.
\end{itemize}
}

\vspace{2mm}
\clearpage
% ─────────────────────────────────────────────────────────────────────────────
% Datasheet for RLDF, following Gebru et al., "Datasheets for Datasets".
% Written for the one-column supplementary; do not add \twocolumn here.
% ─────────────────────────────────────────────────────────────────────────────
\definecolor{darkblue}{RGB}{46,25,110}

\newcommand{\dssectionheader}[1]{%
   \noindent\framebox[\linewidth]{%
      {\fontfamily{phv}\selectfont \textbf{\textcolor{darkblue}{#1}}}
   }\smallskip
}
\newcommand{\dsquestion}[1]{%
    {\noindent \fontfamily{phv}\selectfont \textcolor{darkblue}{\textbf{#1}}}
}
\newcommand{\dsquestionex}[2]{%
    {\noindent \fontfamily{phv}\selectfont \textcolor{darkblue}{\textbf{#1} #2}}
}
\newcommand{\dsanswer}[1]{%
   {\noindent #1 \smallskip}
}

\subsection*{B.\; Datasheet for RLDF}
\label{ap:datasheet}
\small

%%%%%%%%%%%%%%%%%%%%%%%%%%%%%%%%%%%%%%%%%%
\dssectionheader{Motivation}

\dsquestionex{For what purpose was the dataset created?}{Was there a specific task in mind? Was there a specific gap that needed to be filled?}
\dsanswer{
RLDF was created to evaluate deepfake detection when manipulation is confined to a
localised facial sub-region rather than the whole face. Existing benchmarks
(FaceForensics++, Celeb-DF, DFDC, DFDCP, UADFV, DF40) annotate manipulation at the
global image level with a binary label, and every forgery they contain replaces or
synthesises the entire face. No prior dataset supports region-level evaluation, so it
was not previously possible to ask whether detectors remain reliable as the
manipulated area shrinks. RLDF fills that gap by providing partial face swaps with
per-region annotations.
}

\dsquestion{Who created this dataset and on behalf of which entity?}
\dsanswer{Anonymised for double-blind review.}

\dsquestionex{Who funded the creation of the dataset?}{If there is an associated grant, please provide the name of the grantor and the grant name and number.}
\dsanswer{Anonymised for double-blind review.}

\dsquestion{Any other comments?}
\dsanswer{None.}

%%%%%%%%%%%%%%%%%%%%%%%%%%%%%%%%%%%%%%%%%%
\medskip
\dssectionheader{Composition}

\dsquestionex{What do the instances that comprise the dataset represent?}{Are there multiple types of instances?}
\dsanswer{
Each instance is a single $224{\times}224$ aligned face crop, labelled real or fake. A
fake instance is additionally annotated with the face-swapping method used and the
facial region that was manipulated. There is one instance type, and real and fake
instances differ only in their label and annotations.
}

\dsquestion{How many instances are there in total?}
\dsanswer{
Applying the region-aware partial swap synthesis across $|\Phi|{=}8$ face-swapping
methods, $|\mathcal{R}|{=}8$ facial regions, $T{=}32$ frames per pair and 1{,}000
source--target video pairs yields $1{,}000 \times 32 \times 64 = 2{,}048{,}000$ fake
images, balanced with an equal number of unmodified target frames as real samples.
The held-out test split contains approximately 286K images.
}

\dsquestionex{Does the dataset contain all possible instances or is it a sample?}{If the dataset is a sample, what is the larger set? Is the sample representative?}
\dsanswer{
RLDF is a sample. The real source footage is drawn from FaceForensics++, and 1{,}000
source--target video pairs were used, with 32 frames sampled per pair by linear
spacing. Frames on which the face detector failed were discarded. Representativeness
relative to the full FaceForensics++ distribution has not been formally validated, and
RLDF inherits the demographic and capture-condition composition of that corpus.
}

\dsquestionex{What data does each instance consist of?}{``Raw'' data or features?}
\dsanswer{
Raw image data. Each instance is an RGB face crop, together with a metadata record
giving its binary label, and for fakes the swap method and manipulated region. The
soft region masks derived during construction are released alongside the images.
}

\dsquestionex{Is there a label or target associated with each instance?}{If so, please provide a description.}
\dsanswer{
Yes. Every instance carries a binary authenticity label ($y{=}1$ fake, $y{=}0$ real).
Fake instances additionally carry a region label drawn from four discrete regions
(eye, nose, mouth, eyebrow) and four composite regions (Composite-1: eye$+$nose,
Composite-2: $+$mouth, Composite-3: $+$eyebrow, and Full Face), plus the identity of
the generating method.
}

\dsquestionex{Is any information missing from individual instances?}{If so, please explain why.}
\dsanswer{
No fields are missing for retained instances. Frames on which face detection or
alignment failed were excluded during construction rather than retained with partial
annotation.
}

\dsquestionex{Are relationships between individual instances made explicit?}{If so, please describe how.}
\dsanswer{
Yes. Instances derived from the same source--target video pair share an identifier
prefix, and each fake is explicitly linked to the unmodified target frame it was
composited onto. This makes the 64 method$\times$region variants of a given frame
identifiable as a group.
}

\dsquestionex{Are there recommended data splits?}{If so, please describe them and the rationale.}
\dsanswer{
Yes. RLDF uses a 720/140/140 train/validation/test split defined over source video
pairs, with no identity overlap between splits. Splitting at the pair level rather
than the frame level prevents frames of the same subject appearing in both training
and evaluation.
}

\dsquestionex{Are there any errors, sources of noise, or redundancies?}{If so, please describe.}
\dsanswer{
Yes. Region masks are produced by an automatic BiSeNetV2 face parser, so parsing
errors propagate into the composited boundary. Because partial fakes are composited
from a full swap through a smoothed mask rather than generated region-natively, blend
boundaries reflect both the generator and the compositing step. Consecutive sampled
frames from the same clip are visually redundant.
}

\dsquestionex{Is the dataset self-contained, or does it rely on external resources?}{If it links to external resources, are there guarantees they will remain constant, and are there restrictions?}
\dsanswer{
RLDF depends on FaceForensics++ for its real source footage, which must be obtained
separately under that dataset's own licence, and on the eight public face-swapping
implementations used in construction (Appendix~C). Those repositories are third-party
and carry no guarantee of permanence.
}

\dsquestionex{Does the dataset contain data that might be considered confidential?}{If so, please describe.}
\dsanswer{
No. RLDF is derived entirely from FaceForensics++ footage, which is publicly released
for research, and inherits whatever protections that release applied.
}

\dsquestionex{Does the dataset contain data that, if viewed directly, might be offensive, insulting, threatening, or otherwise cause anxiety?}{If so, please describe why.}
\dsanswer{
RLDF consists of manipulated facial imagery. No content-moderation pass was performed
beyond that already applied by FaceForensics++. The dataset depicts identifiable people
whose likenesses have been altered without their involvement, which some viewers may
find uncomfortable, and its purpose is to support detection of exactly that harm.
}

\dsquestionex{Does the dataset relate to people?}{If not, you may skip the remaining questions in this section.}
\dsanswer{Yes. Every instance is a human face.}

\dsquestionex{Does the dataset identify any subpopulations?}{If so, please describe their distributions.}
\dsanswer{
No. RLDF adds no demographic annotation. The underlying distribution is that of
FaceForensics++, which has not been rebalanced here. Consequently, per-region detection
results reported in this paper should not be read as uniform across demographic groups.
}

\dsquestionex{Is it possible to identify individuals, directly or indirectly?}{If so, please describe how.}
\dsanswer{
Yes, visually. RLDF adds no names or explicit identifiers, but the imagery depicts
identifiable individuals to the same extent as FaceForensics++, and the real target
frames are unmodified.
}

\dsquestionex{Does the dataset contain data that might be considered sensitive?}{If so, please describe.}
\dsanswer{
Facial imagery is biometric data. RLDF annotates no other sensitive attribute, though
apparent characteristics could in principle be inferred from the images themselves, as
in the source corpus.
}

\dsquestion{Any other comments?}
\dsanswer{None.}

%%%%%%%%%%%%%%%%%%%%%%%%%%%%%%%%%%%%%%%%%%
\medskip
\dssectionheader{Collection Process}

\dsquestionex{How was the data associated with each instance acquired?}{Was it directly observable, reported by subjects, or inferred/derived? If inferred, was it validated?}
\dsanswer{
Derived. Real footage was taken from FaceForensics++. Fakes were produced by the
region-aware partial swap synthesis described in the paper: a full face swap is
generated with each method, a soft region mask is derived from the target frame, and
only the designated region is composited back onto the target. Region labels follow
deterministically from the mask used, so no annotation judgement was required. Realism
was assessed by a human perceptual study (Appendix~D).
}

\dsquestionex{What mechanisms or procedures were used to collect the data?}{How were these validated?}
\dsanswer{
Faces were detected and aligned with InsightFace Antelope at $640{\times}640$ and
cropped to $224{\times}224$. Region masks were produced by a BiSeNetV2 face parser,
dilated per region and Gaussian-smoothed ($\sigma{=}3.0$) to avoid hard seams. The
eight swap methods were used through their public reference implementations without
modification (Appendix~C).
}

\dsquestion{If the dataset is a sample from a larger set, what was the sampling strategy?}
\dsanswer{
Deterministic. 1{,}000 source--target pairs were used, and 32 frames per pair were
sampled by linear spacing across the clip rather than at random, so the sample is
reproducible.
}

\dsquestion{Who was involved in the data collection process and how were they compensated?}
\dsanswer{
Construction was carried out by the authors. The perceptual realism study
(Appendix~D) involved 37 volunteer participants from academic backgrounds, none of
them deepfake specialists. Participants were fellow research scholars who assisted
voluntarily, and no compensation was provided.
}

\dsquestionex{Over what timeframe was the data collected?}{Does this match the creation timeframe of the underlying data?}
\dsanswer{
Content generation was carried out between November 2025 and May 2026. This does not
match the creation timeframe of the underlying footage, which was recorded for
FaceForensics++ considerably earlier.
}

\dsquestionex{Were any ethical review processes conducted?}{If so, please describe the process and outcome.}
\dsanswer{
No formal institutional ethics review was sought, as none was required for a study of
this kind: participation was voluntary, no personal or demographic data were collected,
only anonymous ratings were recorded, and the imagery shown was already public within
FaceForensics++. The dataset's intended use, licence constraints and prohibited uses
are set out in Appendix~E.
}

\dsquestionex{Does the dataset relate to people?}{If not, you may skip the remaining questions in this section.}
\dsanswer{Yes.}

\dsquestion{Did you collect the data from the individuals in question directly, or via third parties?}
\dsanswer{
Via a third party. All footage originates from FaceForensics++, and no individual was
recorded for this work.
}

\dsquestionex{Were the individuals in question notified about the data collection?}{If so, how was notice provided?}
\dsanswer{Not by the authors. Notification was a matter for the original FaceForensics++ collection.}

\dsquestionex{Did the individuals consent to the collection and use of their data?}{If so, how was consent requested and provided?}
\dsanswer{
Not to the authors directly. FaceForensics++ is derived from consented, publicly
available video content, and RLDF relies on that consent and adds no separate process.
Individuals did not specifically consent to having their likeness partially swapped.
}

\dsquestionex{If consent was obtained, were individuals provided with a mechanism to revoke it?}{If so, please describe.}
\dsanswer{
Not by the authors. Any revocation mechanism is that of FaceForensics++, and a withdrawal
upstream would require the corresponding RLDF instances to be removed.
}

\dsquestionex{Has an analysis of the potential impact of the dataset on data subjects been conducted?}{If so, please describe the outcomes.}
\dsanswer{
No formal data protection impact assessment was conducted. Ethical considerations and
prohibited uses are stated in Appendix~E.
}

\dsquestion{Any other comments?}
\dsanswer{None.}

%%%%%%%%%%%%%%%%%%%%%%%%%%%%%%%%%%%%%%%%%%
\medskip
\dssectionheader{Preprocessing / cleaning / labeling}

\dsquestionex{Was any preprocessing/cleaning/labeling done?}{If so, please describe. If not, you may skip the remainder of this section.}
\dsanswer{
Yes. Faces were detected, aligned and cropped to $224{\times}224$, and frames with no
detected face were discarded. Region masks were dilated and Gaussian-smoothed before
compositing. Splits were formed at the video-pair level to prevent identity leakage.
}

\dsquestionex{Was the ``raw'' data saved in addition to the preprocessed data?}{If so, please provide a link.}
\dsanswer{
The unmodified target frames are retained as the real half of the dataset. The original
FaceForensics++ videos are not redistributed and must be obtained from that project.
}

\dsquestionex{Is the software used to preprocess/clean/label the instances available?}{If so, please provide a link.}
\dsanswer{
Yes. The construction pipeline, including mask derivation and compositing, is released
with the code (Algorithm~1 in the main paper gives the procedure in full).
}

\dsquestion{Any other comments?}
\dsanswer{None.}

%%%%%%%%%%%%%%%%%%%%%%%%%%%%%%%%%%%%%%%%%%
\medskip
\dssectionheader{Uses}

\dsquestionex{Has the dataset been used for any tasks already?}{If so, please describe.}
\dsanswer{
Yes. RLDF forms Protocol~\#3 of the benchmark reported in this paper, on which 12
state-of-the-art detectors were evaluated alongside two classic cross-dataset protocols.
It was also used to train and evaluate the mask-guided cross-attention baseline.
}

\dsquestionex{Is there a repository that links to papers or systems that use the dataset?}{If so, please provide a link.}
\dsanswer{Not at present.}

\dsquestion{What (other) tasks could the dataset be used for?}
\dsanswer{
Forgery localisation and segmentation, since the region of manipulation is known,
analysis of detector sensitivity as a function of manipulated area, robustness studies
across swap methods, and as pretraining data for spatially grounded detectors.
}

\dsquestionex{Is there anything about the composition or collection process that might impact future uses?}{Is there anything a future user should know to avoid unfair treatment or other harms?}
\dsanswer{
Three points. First, partial fakes are composited rather than generated
region-natively, so a detector may learn the compositing signature instead of a
generator artefact, and conclusions should be checked against region-native manipulations
where possible. Second, RLDF covers face swapping only, not reenactment, entire-face
synthesis or attribute editing. Third, it inherits the demographic composition of
FaceForensics++ and carries no demographic labels, so per-group performance cannot be
measured on it and should not be assumed uniform.
}

\dsquestionex{Are there tasks for which the dataset should not be used?}{If so, please describe.}
\dsanswer{
RLDF must not be used to build or improve face-swapping or impersonation systems, for
political messaging, or for any application that could harm the privacy or reputation
of the individuals in the source footage (Appendix~E). Detectors trained on it should
not be deployed to make consequential decisions about people without independent
validation on the target domain.
}

\dsquestion{Any other comments?}
\dsanswer{None.}

%%%%%%%%%%%%%%%%%%%%%%%%%%%%%%%%%%%%%%%%%%
\medskip
\dssectionheader{Distribution}

\dsquestionex{Will the dataset be distributed to third parties outside of the entity on behalf of which it was created?}{If so, please describe.}
\dsanswer{Yes. RLDF is intended for public release to the research community.}

\dsquestionex{How will the dataset be distributed?}{Does the dataset have a digital object identifier (DOI)?}
\dsanswer{
Via Hugging Face, with the construction and evaluation code released separately. The
release URL is withheld here to preserve double-blind review and will be given in the
camera-ready version. No DOI has been minted.
}

\dsquestion{When will the dataset be distributed?}
\dsanswer{Publicly by the camera-ready deadline, pending acceptance.}

\dsquestionex{Will the dataset be distributed under a copyright or other IP license, and/or under applicable terms of use?}{If so, please describe.}
\dsanswer{
Access is conditioned on compliance with the FaceForensics++ academic research licence
and with the licences of all eight face-swapping methods used in construction. Use is
restricted to non-commercial research. The licence applied to RLDF's own contributed
content (region masks, annotations and composited images) will be stated on release,
upon acceptance.
}

\dsquestionex{Have any third parties imposed IP-based or other restrictions on the data?}{If so, please describe.}
\dsanswer{
Yes. FaceForensics++ and each of the eight swap-method implementations carry their own
licences, with which users must independently comply.
}

\dsquestionex{Do any export controls or other regulatory restrictions apply?}{If so, please describe.}
\dsanswer{None known to the authors.}

\dsquestion{Any other comments?}
\dsanswer{None.}

%%%%%%%%%%%%%%%%%%%%%%%%%%%%%%%%%%%%%%%%%%
\medskip
\dssectionheader{Maintenance}

\dsquestion{Who will be supporting/hosting/maintaining the dataset?}
\dsanswer{Anonymised for double-blind review.}

\dsquestion{How can the owner/curator/manager of the dataset be contacted?}
\dsanswer{Anonymised for double-blind review.}

\dsquestionex{Is there an erratum?}{If so, please provide a link.}
\dsanswer{No. Errata will be recorded in the release repository if any arise.}

\dsquestionex{Will the dataset be updated?}{If so, please describe how often, by whom, and how updates will be communicated.}
\dsanswer{
Corrections to region masks or labels will be issued if errors are identified. There is
no fixed release schedule, and updates will be published as revisions of the Hugging Face
repository, and a changelog will be maintained from the public release onward.
}

\dsquestionex{If the dataset relates to people, are there applicable limits on the retention of the data?}{If so, please describe and explain enforcement.}
\dsanswer{
RLDF collects no data directly from individuals, and retention is governed by the
FaceForensics++ terms. A withdrawal upstream would require removal of the corresponding
RLDF instances.
}

\dsquestionex{Will older versions of the dataset continue to be supported/hosted/maintained?}{If so, please describe how.}
\dsanswer{
Releases will be versioned so that published results remain reproducible. Because the
repository is git-backed, superseded revisions remain retrievable by their commit
identifier and will not be deleted.
}

\dsquestionex{If others want to extend/augment/build on/contribute to the dataset, is there a mechanism for them to do so?}{If so, please describe.}
\dsanswer{
The construction pipeline is released, so others can extend RLDF to further swap
methods or regions. Contributions are not centrally validated by the authors.
}

\dsquestion{Any other comments?}
\dsanswer{None.}

\medskip
\subsection*{C.\; Face-Swapping Method Selection}
\vspace{1mm}

{\small
RLDF uses eight face-swapping methods chosen to maximise diversity across three axes: \textit{technical paradigm}
(identity disentanglement, GAN-based blending, latent-space inversion, reenactment-guided, and lightweight mobile),
\textit{publication venue and recency} (2019--2023; ArXiv, CVPR, ICCV, ECCV, and WACV), and \textit{visual artifact profile}
(seam-based, style-transfer, colour-shift, and blending artefacts). None supports partial manipulation natively, all operate
on the full face, making them suitable for region-aware partial swap synthesis without architectural modification. All eight
are publicly available and widely benchmarked in deepfake detection literature.
}

\begin{table}[htbp]
\centering
\small
\caption{Face-swapping methods used in RLDF construction.}
\label{tab:supp_methods}
\setlength{\tabcolsep}{3pt}
\renewcommand{\arraystretch}{1.12}
\resizebox{\linewidth}{!}{%
\begin{tabular}{clllll}
\toprule
\# & \textbf{Method} & \textbf{Paradigm} & \textbf{Venue} & \textbf{Key characteristic} & \textbf{Code} \\
\midrule
1 & BlendFace~\cite{blendface}   & GAN, disentangle    & ICCV 2023  & Seam-minimising latent blending; hardest to detect   & \href{https://github.com/mapooon/BlendFace}{Link} \\
2 & E4S~\cite{e4s}               & Latent inversion    & CVPR 2023  & Style-transfer boundaries; most detectable           & \href{https://github.com/e4s2022/e4s}{Link} \\
3 & FaceDancer~\cite{facedancer} & GAN, disentangle    & WACV 2023  & Adaptive feature injection across identities         & \href{https://github.com/felixrosberg/FaceDancer}{Link} \\
4 & FSGAN~\cite{fsgan}           & Reenactment-guided  & ArXiv 2019 & Parsing-mask-based subject-agnostic swap             & \href{https://github.com/YuvalNirkin/fsgan}{Link} \\
5 & GHOST~\cite{ghost}           & GAN, lightweight    & ArXiv 2022 & Efficient one-shot swap with strong colour consistency & \href{https://github.com/ai-forever/ghost}{Link} \\
6 & InSwapper~\cite{inswapper}   & GAN (Roop)          & —          & Commercial-grade; widely used in real-world deepfakes  & \href{https://github.com/haofanwang/inswapper}{Link} \\
7 & SimSwap~\cite{SimSwap}       & GAN, disentangle    & ICCV 2019  & Weak feature matching preserves target attributes    & \href{https://github.com/neuralchen/SimSwap}{Link} \\
8 & UniFace~\cite{uniface}       & GAN, disentangle    & ECCV 2022  & Unified framework for arbitrary face swapping        & \href{https://github.com/xc-csc101/UniFace}{Link} \\
\bottomrule
\end{tabular}}
\end{table}

% ── H. Human Perceptual Study ─────────────────────────────────────────────────
\subsection*{D.\; Human Perceptual Realism Study}
\label{ap:C}
\vspace{1mm}

To validate that RLDF failure cases are genuinely difficult for humans not just detectors we conducted a perceptual realism survey.
Participants were shown 14 region-level deepfake images (2 samples per region across 7 regions: Eye, Nose, Mouth, C-1, C-2, C-3, Full Face) and asked to rate each image on a scale of \textbf{0} (looks completely real) to \textbf{5} (looks completely fake).
All 14 images were drawn from detector failure cases identified in Protocol~\#3 (Appendix~J).

37 participants completed the survey, recruited from diverse academic backgrounds spanning Computer Science, Electronics, Electrical Engineering, Mechanical Engineering, Civil Engineering, and Humanities \& Social Sciences.
Participants included undergraduate (BE), postgraduate (ME), and doctoral (PhD) scholars, none of whom were deepfake specialists.

\begin{figure}[htbp]
    \centering
\includegraphics[width=0.82\linewidth,height=0.9\textheight,keepaspectratio]{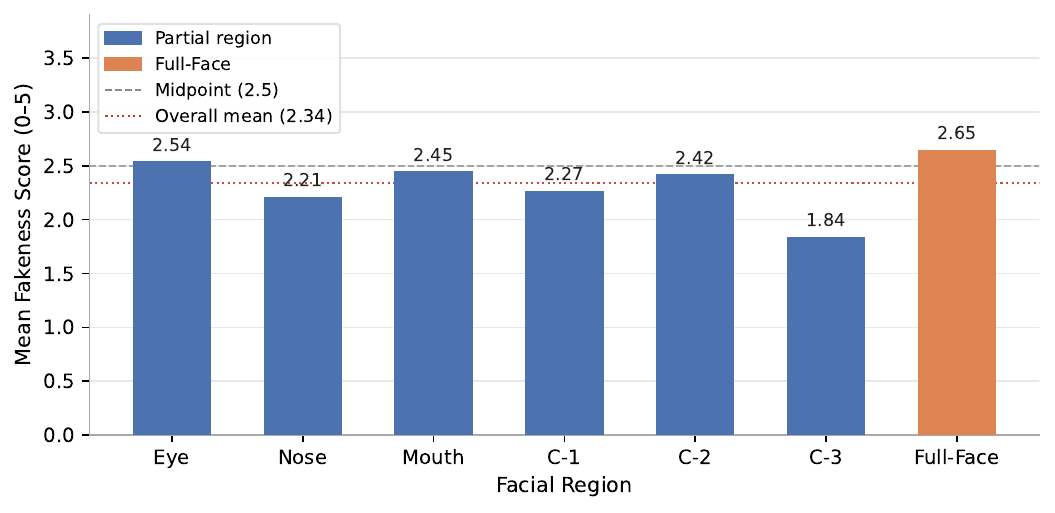}
\caption{\textbf{Human Perceptual Realism Survey.}
Per-region mean fakeness score (0 = real, 5 = obviously fake) rated by 37 participants across 14 region-level deepfake images.
Dashed line = midpoint (2.5); dotted red line = overall mean (2.34).
Partial-region fakes score below the midpoint; Full-Face fakes are the most perceptually detectable.}
\label{tab:survey}
\end{figure}

Figure~\ref{tab:survey} reports the mean and standard deviation of perceived fakeness per image.
The overall mean score is \textbf{2.34 / 5.0}, indicating that participants were largely uncertain leaning toward perceiving these images as \emph{real}.
Certain partial-region samples received the lowest scores (Nose S2: 1.11, C-3 S1: 1.24), confirming that small-region forgeries are perceptually near-indistinguishable from authentic faces.
In particular, Full-Face samples scored highest (2.92 and 2.38), consistent with the expectation that full-face swaps introduce more visible artifacts than partial-region fakes.
The high standard deviations (${\sim}1.7$) reflect genuine inter-rater ambiguity rather than systematic disagreement.
These results support the core motivation for RLDF: region-level fakes represent a qualitatively harder detection regime for both humans and detectors.

% ── I. Ethical Statement ──────────────────────────────────────────────────────
\subsection*{E.\; Ethical Statement}
RLDF is constructed from FaceForensics++~\cite{ff++} real videos under its academic research licence and is intended strictly for non-commercial deepfake detection research.
The dataset must not be used for impersonation, political messaging, or any application that could harm the privacy or reputation of individuals in the source videos.
Users must comply with the FaceForensics++ licence~\cite{ff++} and the licences of all face-swapping methods used in dataset construction.
All images in RLDF are derived from consented public-domain video content; no new identities were collected or recorded for this work.
% ── Dataset Visualizations ────────────────────────────────────────────────────
\medskip
\subsection*{F.\; RLDF: Dataset Visualizations}
\begin{figure}[htbp]
    \centering
\includegraphics[width=\linewidth,height=0.62\textheight,keepaspectratio]{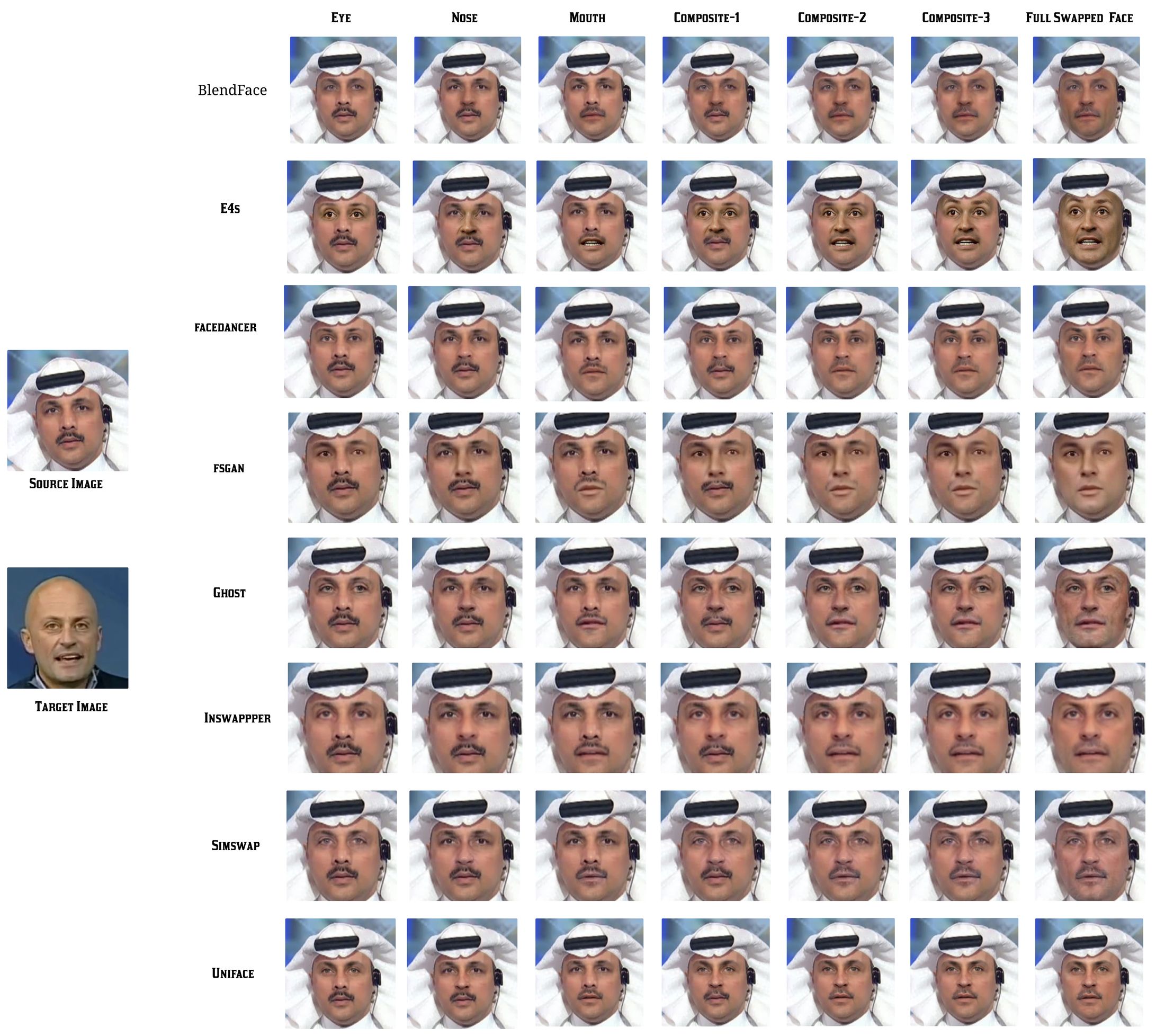}
    \captionof{figure}{RLDF sample images across  8 face-swapping methods (rows) and seven manipulation regions of increasing spatial coverage from single regions (Eye, Nose, Mouth) to full face swap (columns).}
    \label{fig:ab_RLDF_Dataset}
    \vspace{6pt}
\includegraphics[width=\linewidth,height=0.25\textheight,keepaspectratio]{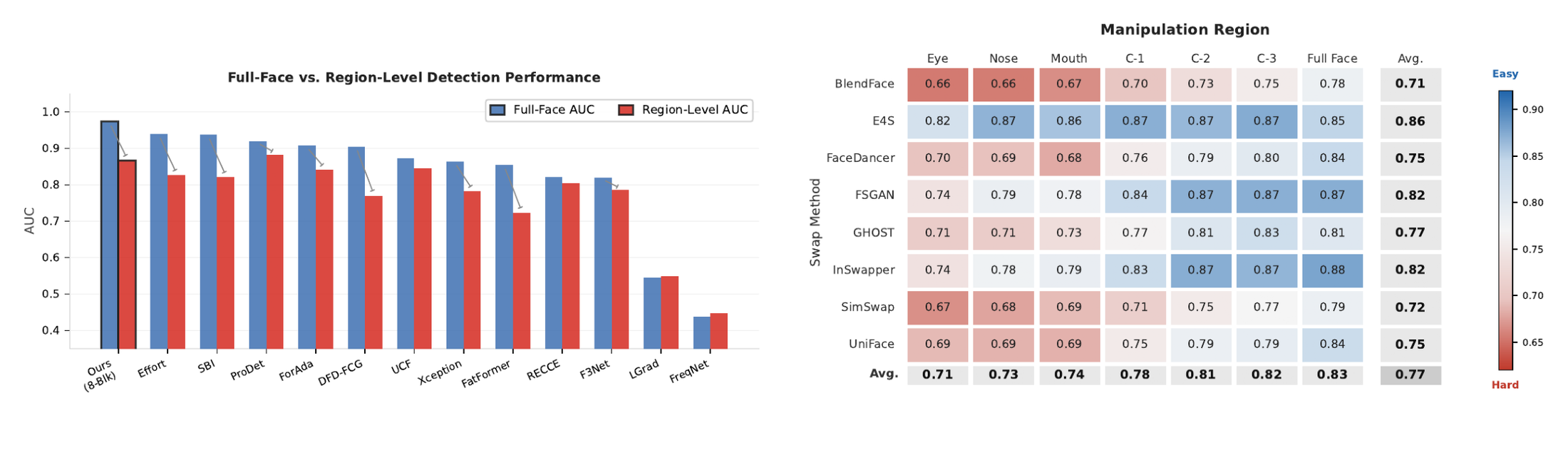}
    \captionof{figure}{\textbf{RLDF overview:} (a) full face vs.\ all region-level combined  AUC drop
  across 12 detectors, showing consistent degradation under 
  evaluation, and (b) per-region AUC heatmap averaged across all detectors
  and swap methods, exposing systematic failure on individual facial regions.}
    \label{fig:overview_supp}
\end{figure}

\begin{figure}[htbp]
    \centering
\includegraphics[width=\linewidth,height=0.9\textheight,keepaspectratio]{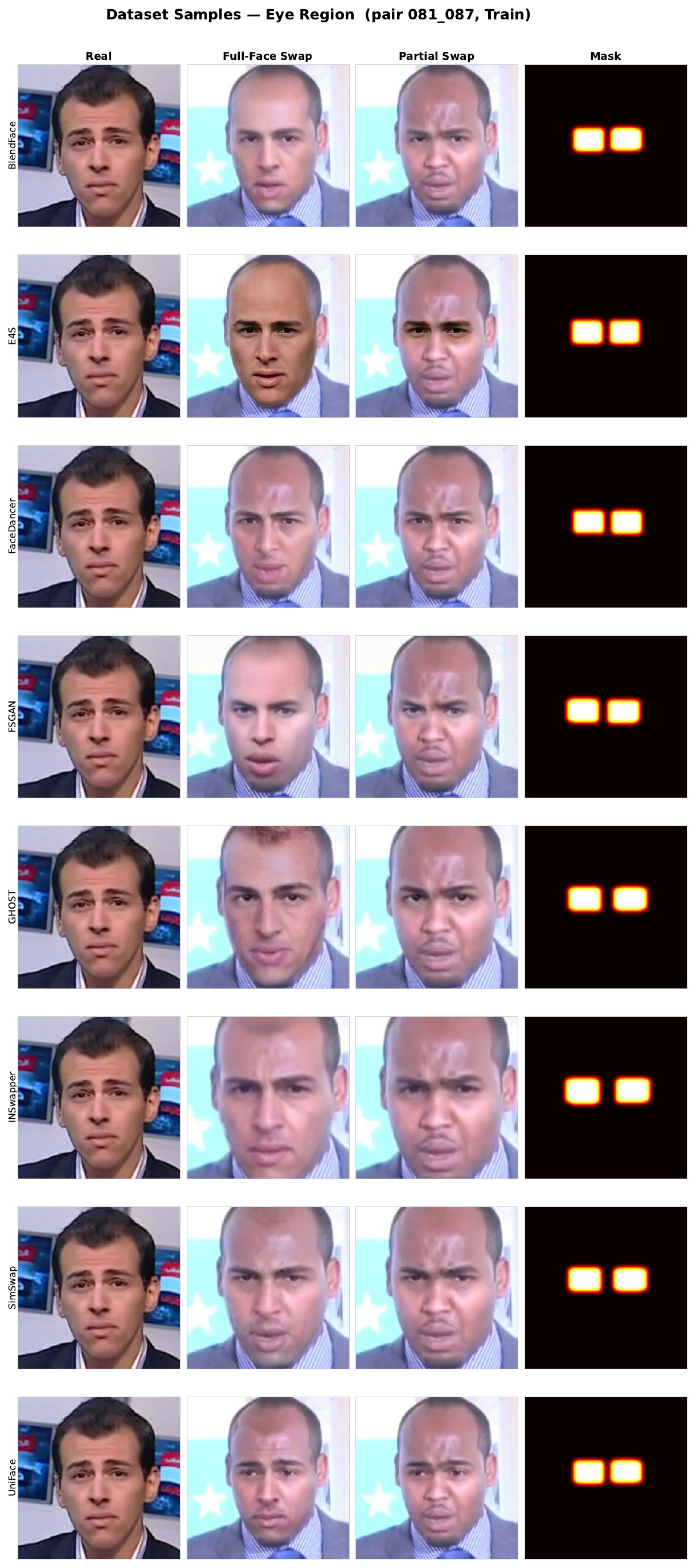}
    \caption{Dataset samples: \textbf{Eye} region (pair 081/087).}
    \label{fig:supp_samples_eye}
\end{figure}

\begin{figure}[htbp]
    \centering
\includegraphics[width=\linewidth,height=0.9\textheight,keepaspectratio]{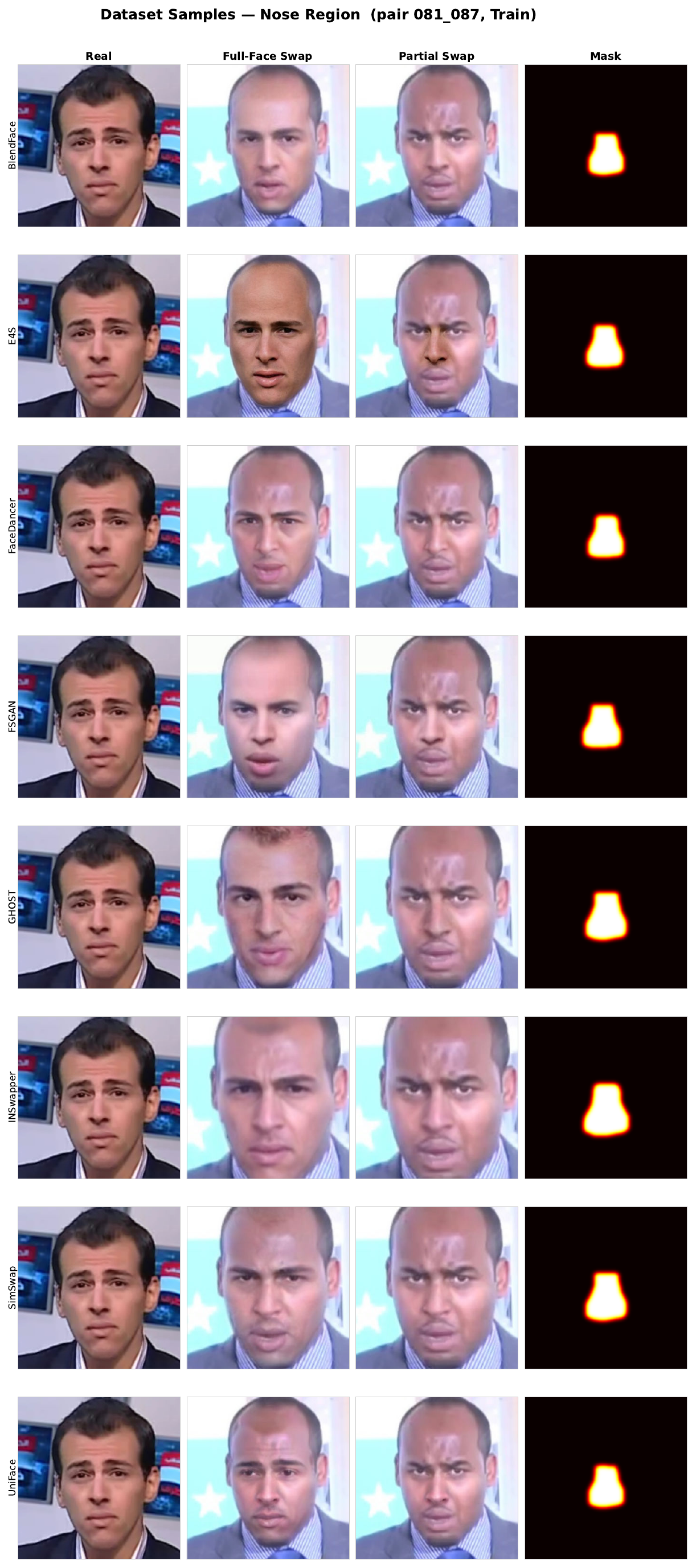}
    \caption{Dataset samples: \textbf{Nose} region (pair 081/087).}
    \label{fig:supp_samples_nose}
\end{figure}

\begin{figure}[htbp]
    \centering
\includegraphics[width=\linewidth,height=0.9\textheight,keepaspectratio]{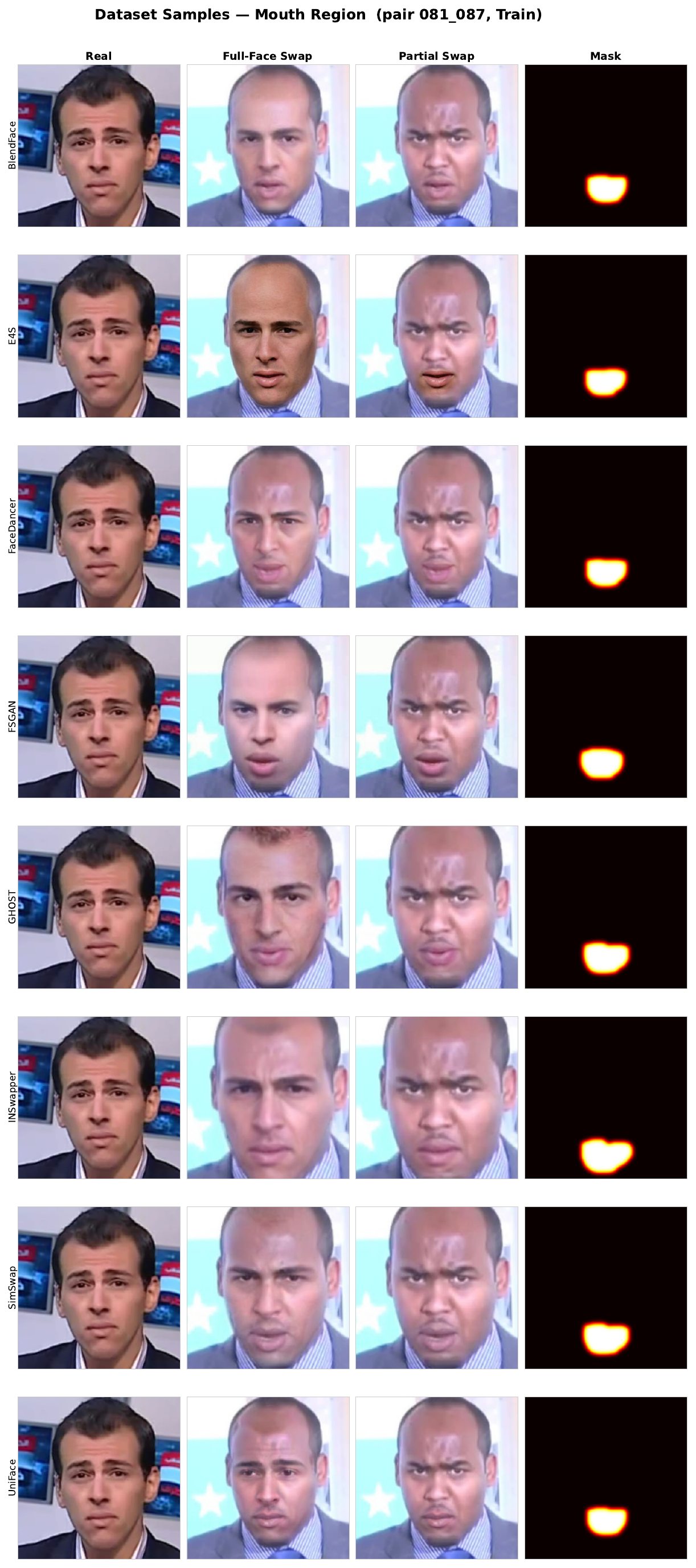}
    \caption{Dataset samples: \textbf{Mouth} region (pair 081/087).}
    \label{fig:supp_samples_mouth}
\end{figure}

\begin{figure}[htbp]
    \centering
\includegraphics[width=\linewidth,height=0.9\textheight,keepaspectratio]{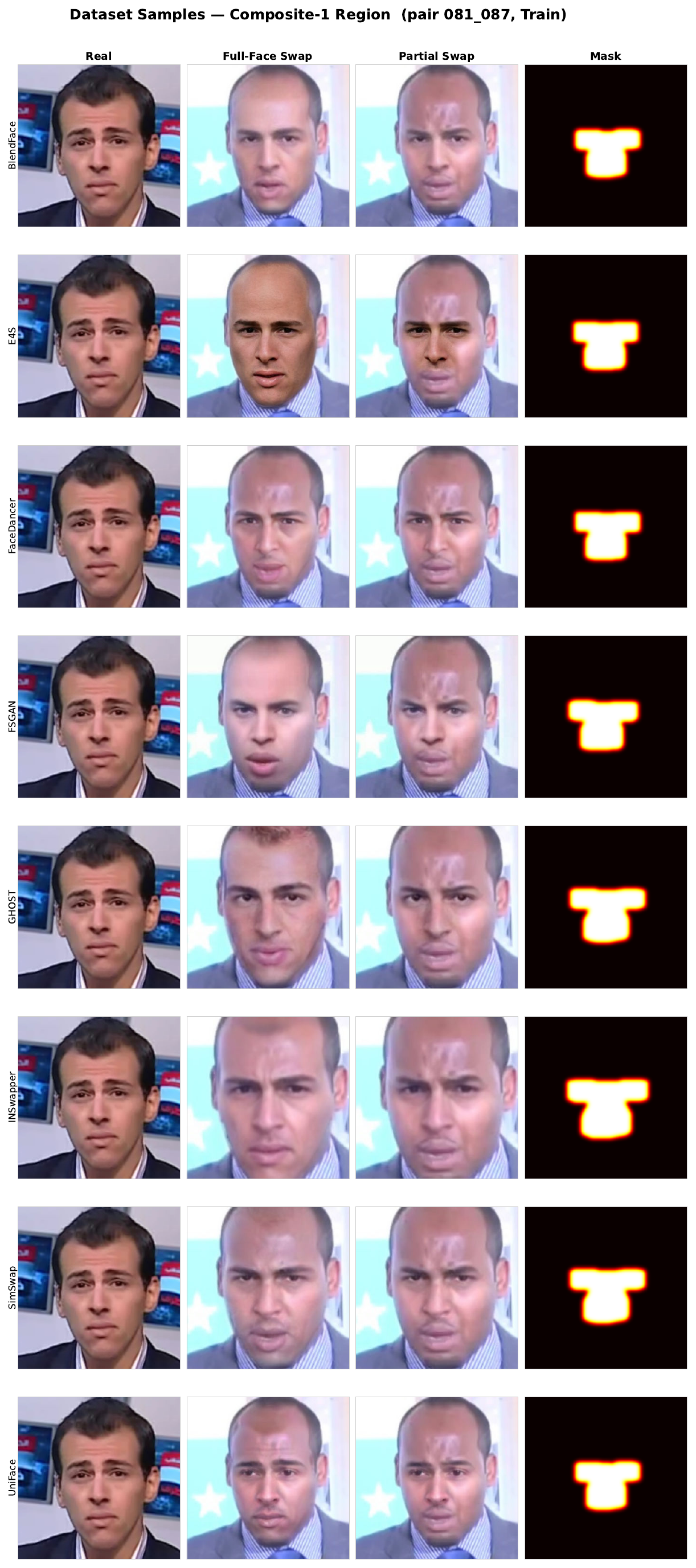}
    \caption{Dataset samples: \textbf{Composite-1} (Eye+Nose) region (pair 081/087).}
    \label{fig:supp_samples_c1}
\end{figure}

\begin{figure}[htbp]
    \centering
\includegraphics[width=\linewidth,height=0.9\textheight,keepaspectratio]{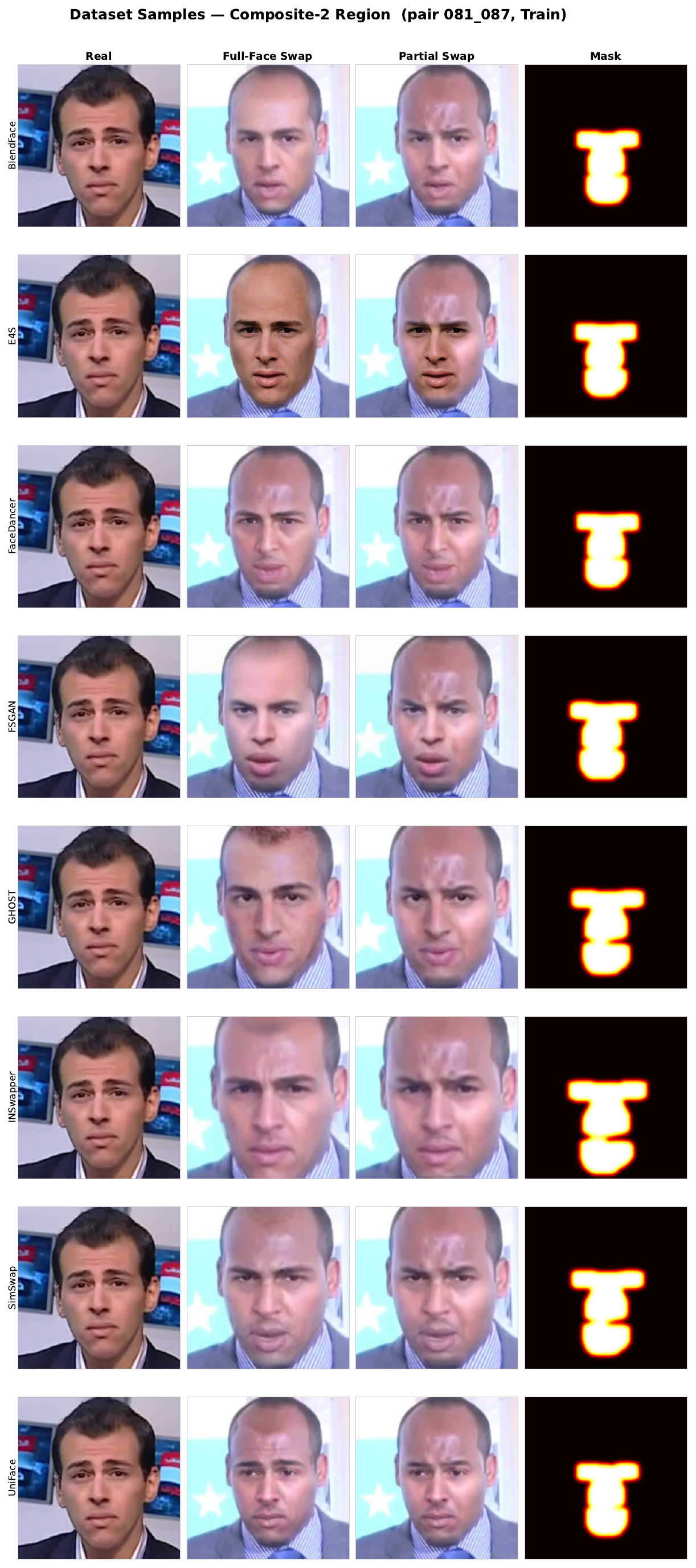}
    \caption{Dataset samples: \textbf{Composite-2} (Eye+Nose+Mouth) region (pair 081/087).}
    \label{fig:supp_samples_c2}
\end{figure}

\begin{figure}[htbp]
    \centering
\includegraphics[width=\linewidth,height=0.9\textheight,keepaspectratio]{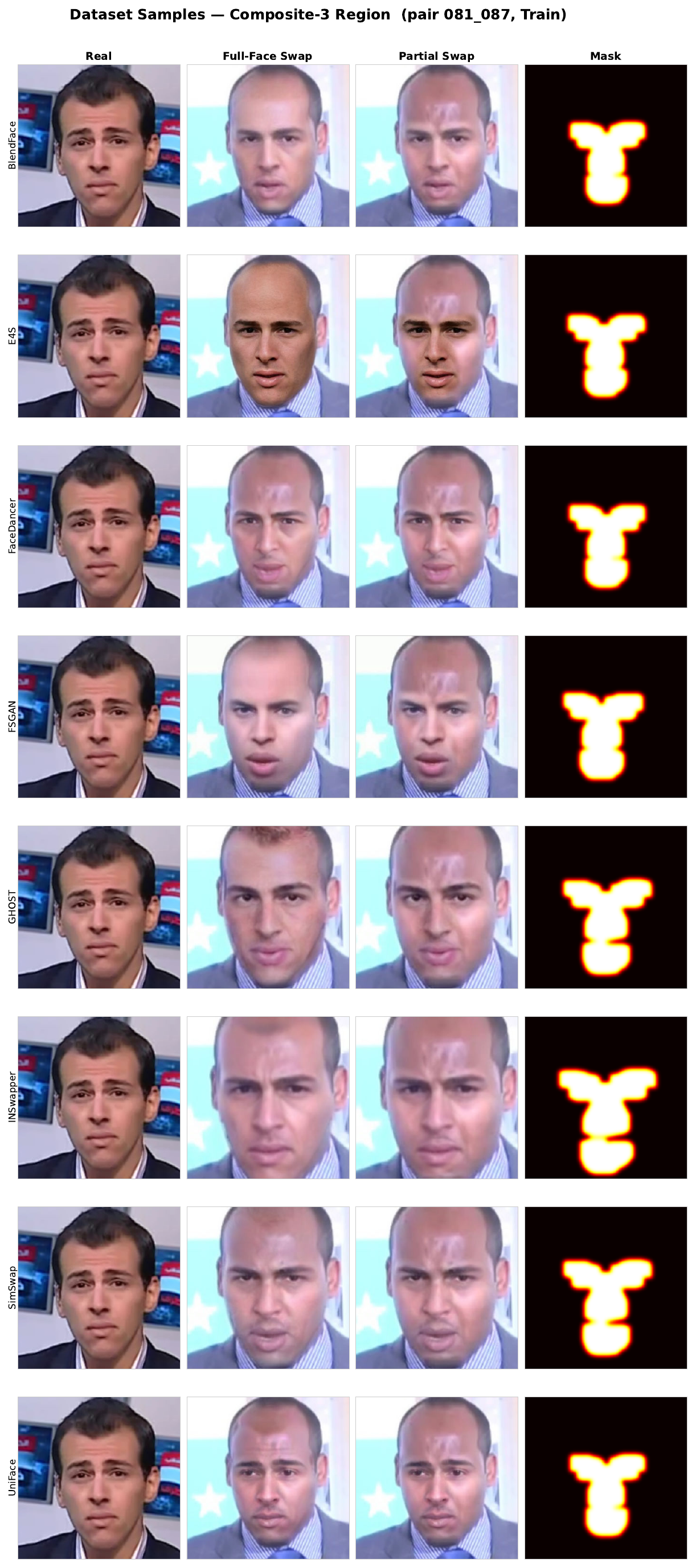}
    \caption{Dataset samples: \textbf{Composite-3} (Eye+Nose+Mouth+Eyebrow) region (pair 081/087).}
    \label{fig:supp_samples_c3}
\end{figure}

This section presents representative visual samples from the RLDF dataset, illustrating the region-isolation 
pipeline across all eight face-swapping methods and seven manipulation regions. 
Figures~\ref{fig:ab_RLDF_Dataset} to \ref{fig:supp_samples_c3} show sample images spanning single-region 
manipulations (Eye, Nose, Mouth) through composite regions (Composite-1/2/3) to full face swap, along 
with their corresponding soft region masks, demonstrating the progressive increase in spatial coverage 
and detection difficulty as the manipulated region shrinks.
\vspace{1mm}


% ── C. Region-Level Performance Breakdown ─────────────────────────────────────
\subsection*{G.\; Region-Level Performance Breakdown}
\label{ap:F}
\begin{sidewaysfigure}
    \centering
\includegraphics[width=\linewidth,height=0.92\textwidth,keepaspectratio]{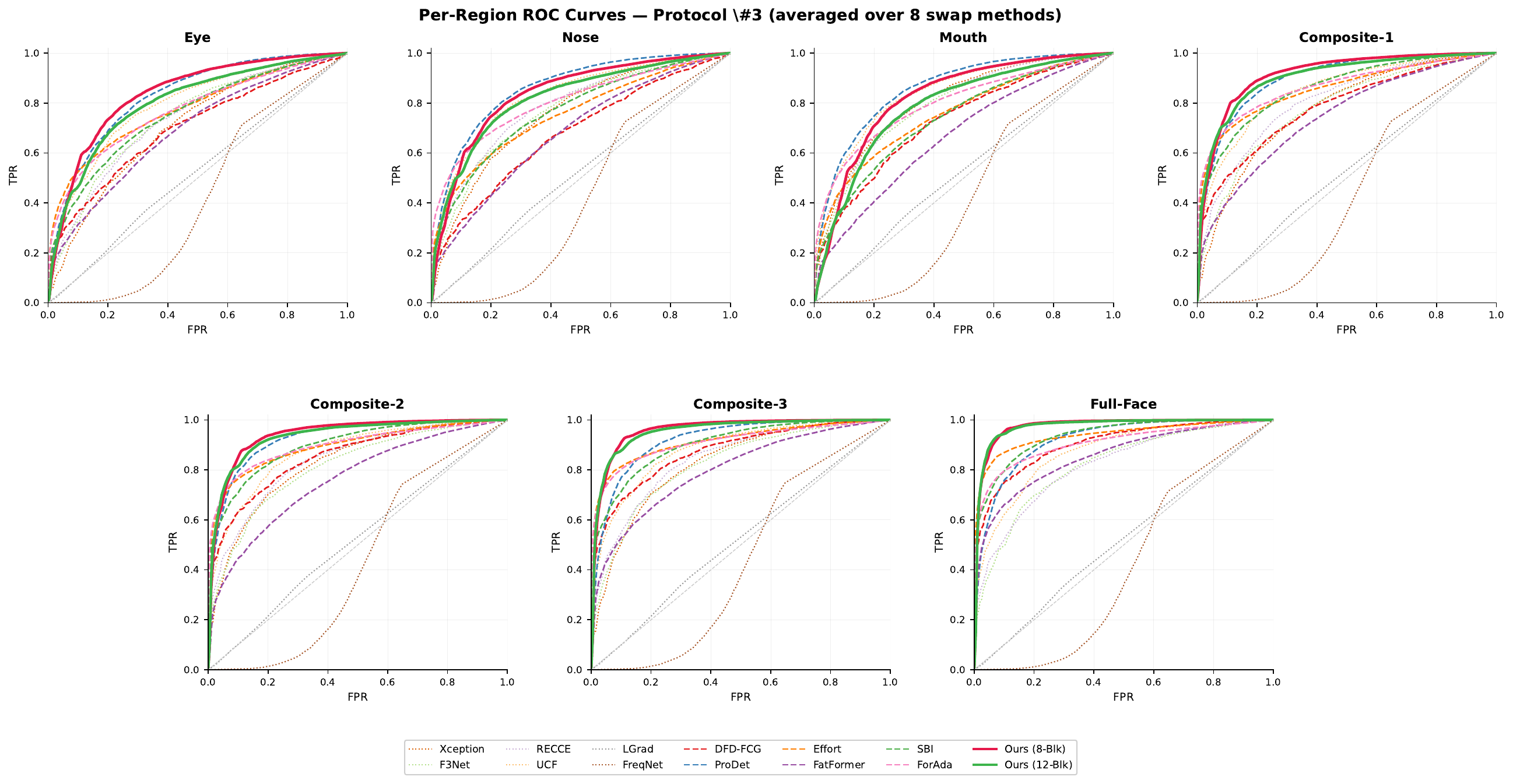}
    \caption{ROC curves averaged per manipulation region (Protocol~\#3). Detection AUC degrades as region size decreases, with Eye being the hardest.}
    \label{fig:supp_roc_region}
\end{sidewaysfigure}

\begin{sidewaysfigure}
    \centering
\includegraphics[width=\linewidth,height=0.92\textwidth,keepaspectratio]{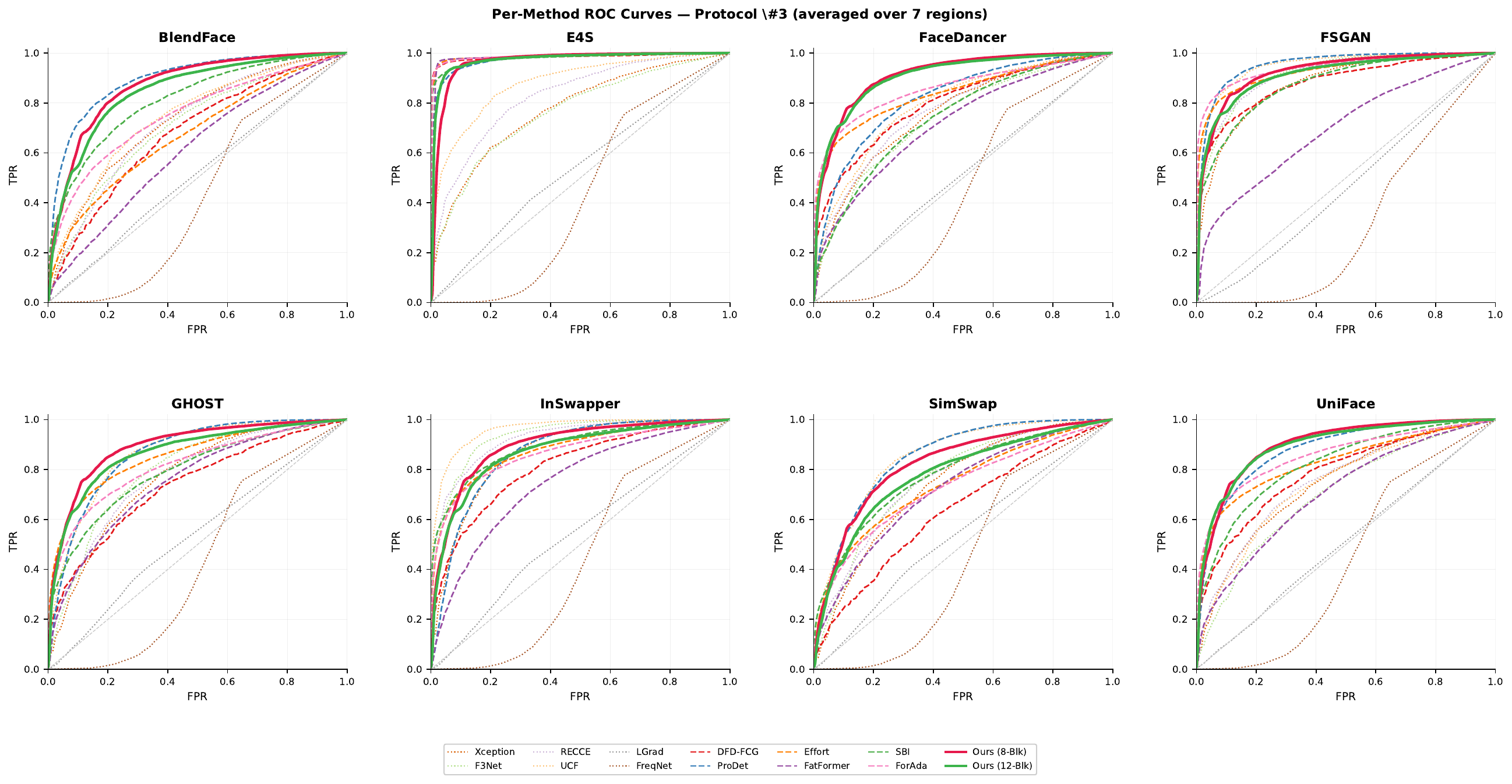}
    \caption{ROC curves averaged per face-swap method (Protocol~\#3). E4S is the most detectable; BlendFace is the hardest.}
    \label{fig:supp_roc_method}
\end{sidewaysfigure}

\noindent\textbf{ROC Curves:}
Figures~\ref{fig:supp_roc_region} and~\ref{fig:supp_roc_method} show ROC curves under Protocol~\#3 averaged per region and per swap method respectively.
Detection difficulty monotonically increases as the manipulated area shrinks: Full Face fakes are easiest (AUC~$\geq$~0.90 for most detectors) while single-region fakes, particularly the Eye region ($\approx$5\% of face area), push all detectors toward chance level.
Across swap methods, E4S is the most detectable owing to its visible style-transfer boundaries, while BlendFace is the hardest due to its seam-minimising latent blending.


\begin{figure}[htbp]
    \centering
    \includegraphics[width=0.78\linewidth,height=0.9\textheight,keepaspectratio]{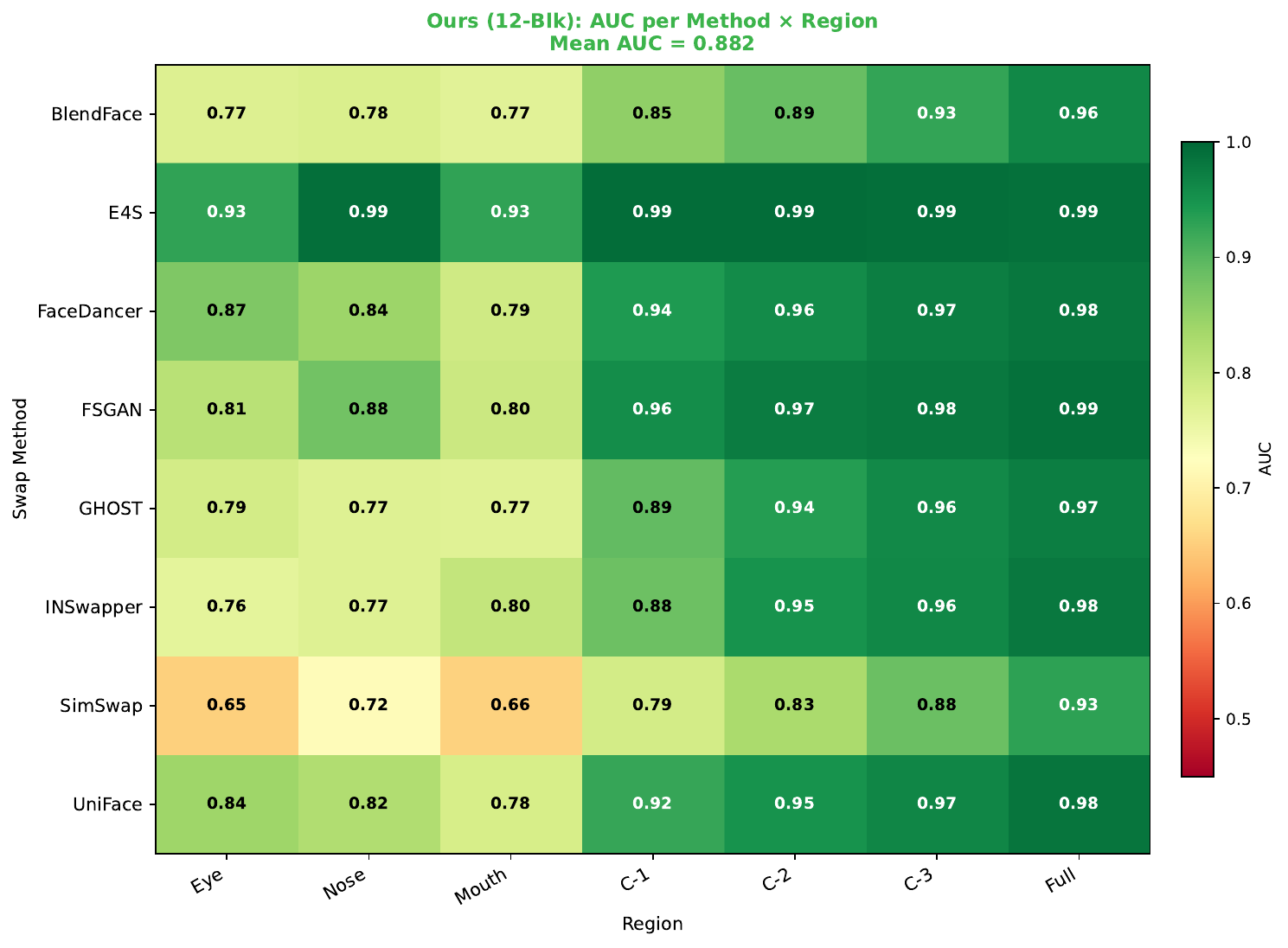}
    \caption{AUC heatmap: \textbf{Ours (12-Blk)}. Most uniform coverage.}
    \label{fig:auc_ours12}
\end{figure}

\begin{figure}[htbp]
    \centering
    \includegraphics[width=0.78\linewidth,height=0.9\textheight,keepaspectratio]{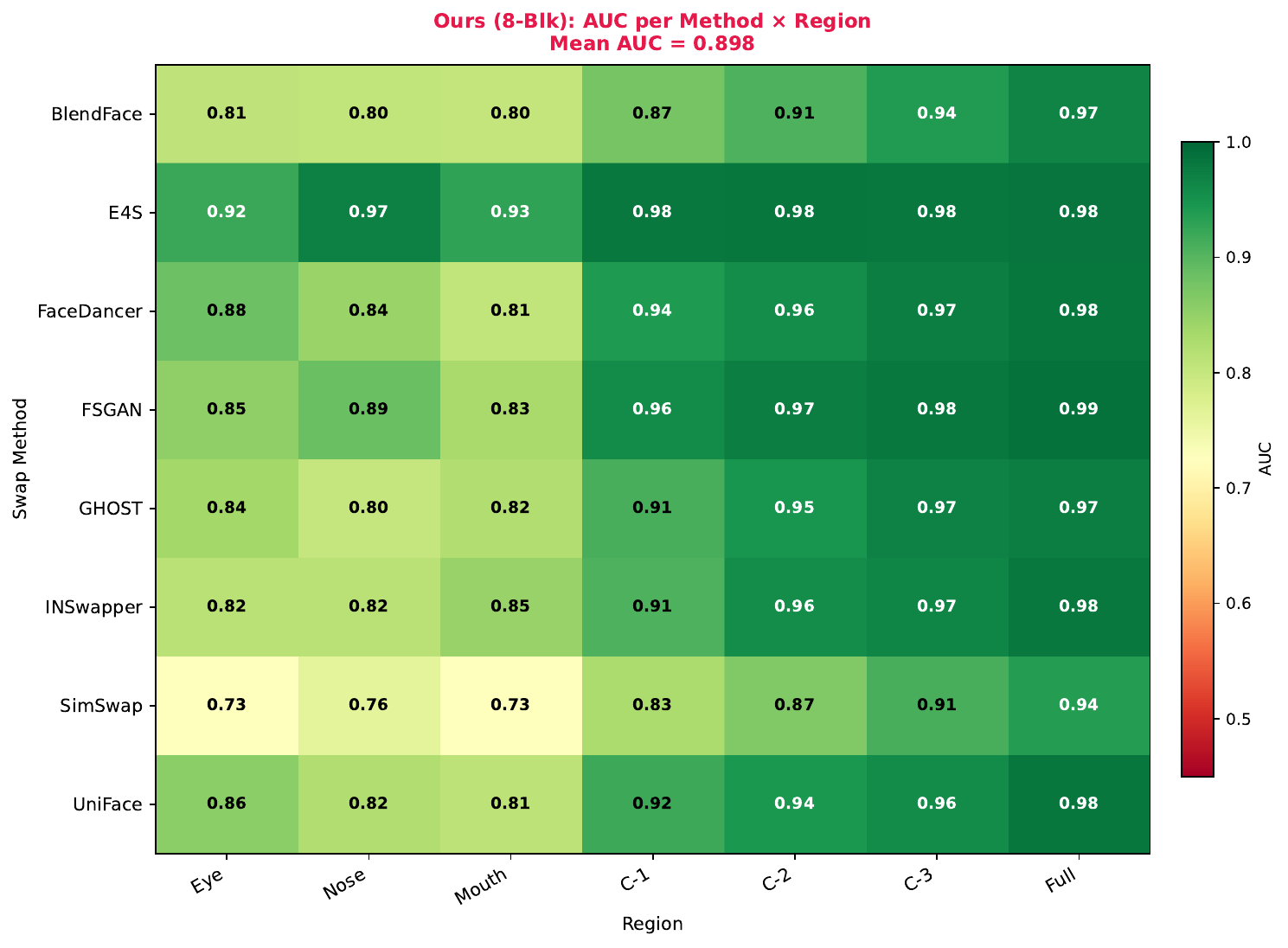}
    \caption{AUC heatmap: \textbf{Ours (8-Blk)}. Slightly stronger on small regions.}
    \label{fig:auc_ours8}
\end{figure}

\begin{figure}[htbp]
    \centering
    \includegraphics[width=0.78\linewidth,height=0.9\textheight,keepaspectratio]{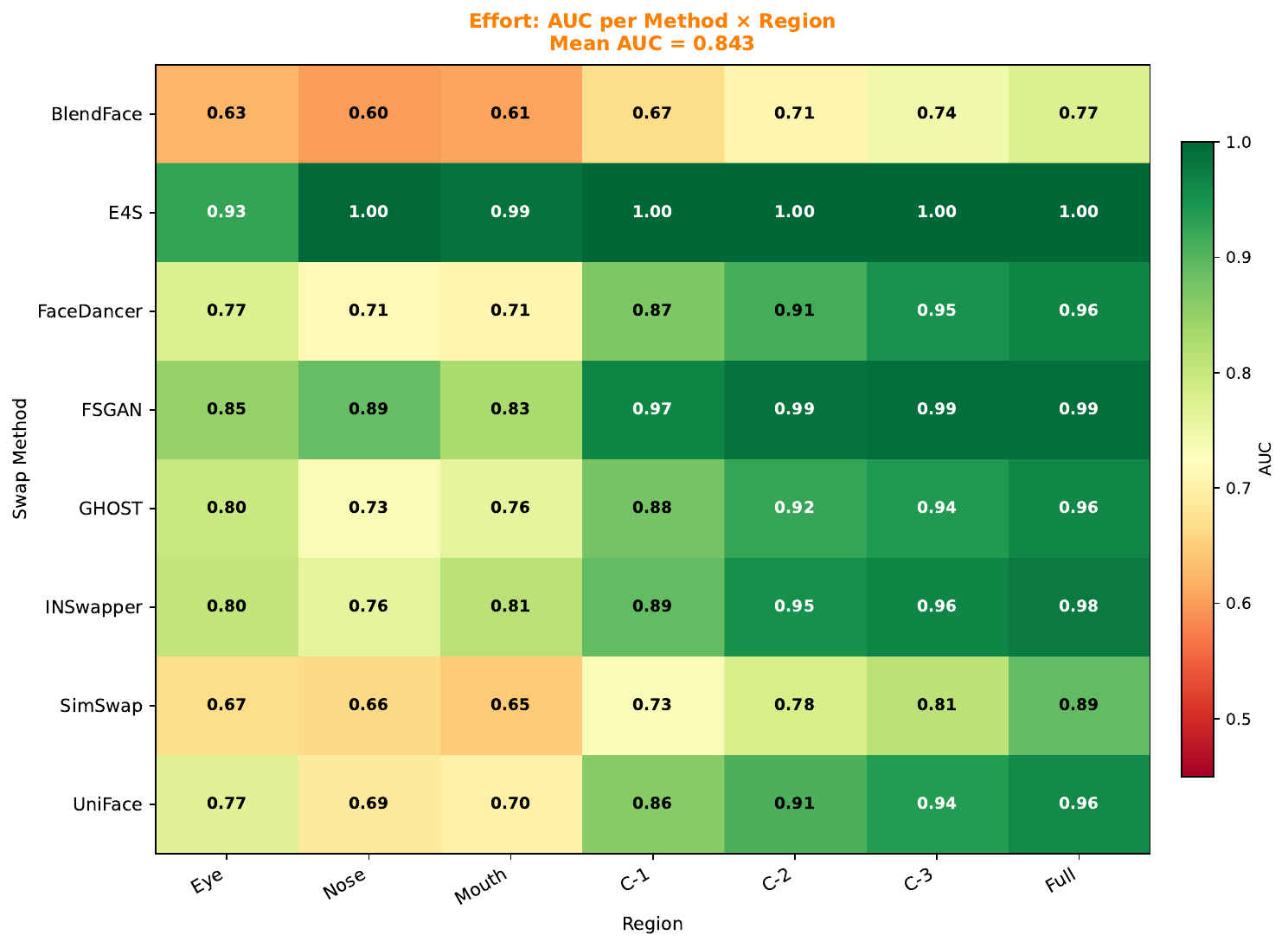}
    \caption{AUC heatmap: \textbf{Effort}. Strong on full face, weakens for single-region.}
    \label{fig:auc_effort}
\end{figure}

\begin{figure}[htbp]
    \centering
    \includegraphics[width=0.78\linewidth,height=0.9\textheight,keepaspectratio]{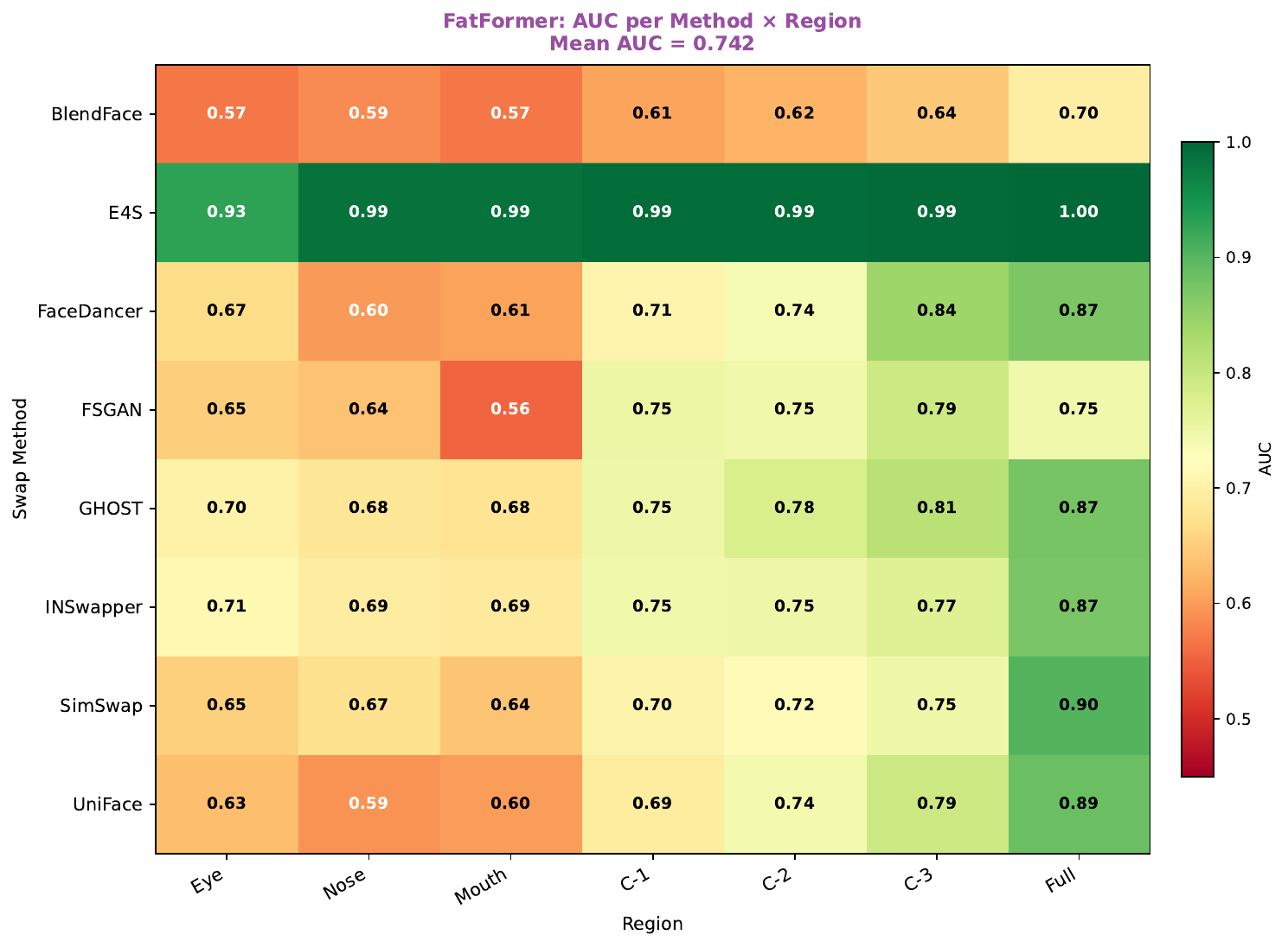}
    \caption{AUC heatmap: \textbf{FatFormer}. High on E4S, collapses on BlendFace.}
    \label{fig:auc_fatformer}
\end{figure}

\begin{figure}[htbp]
    \centering
    \includegraphics[width=0.78\linewidth,height=0.9\textheight,keepaspectratio]{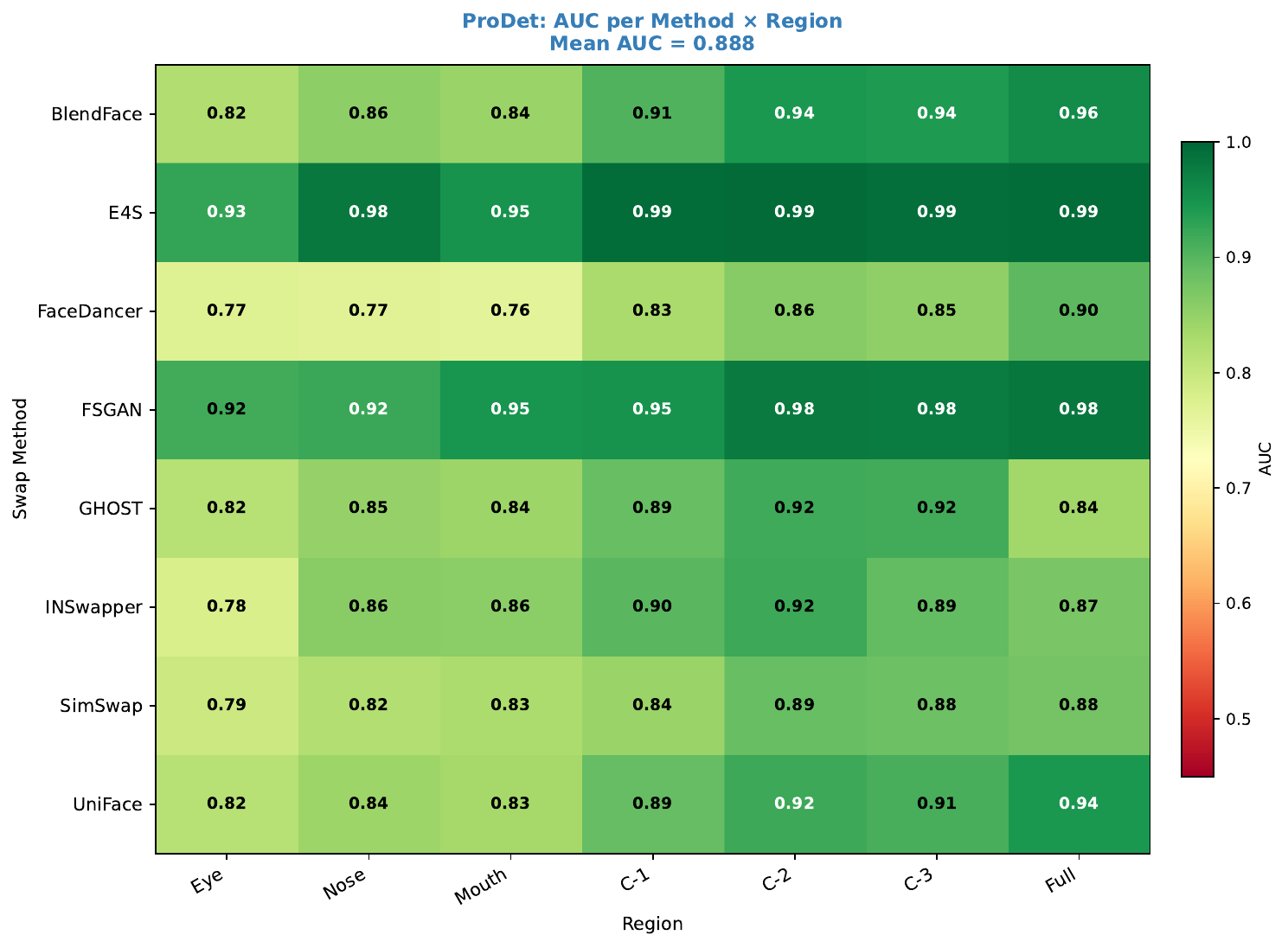}
    \caption{AUC heatmap: \textbf{ProDet}. Strong on Full Face, poor on small regions.}
    \label{fig:auc_prodet}
\end{figure}

\noindent\textbf{AUC Heatmaps:}
Figures~\ref{fig:auc_ours12}--\ref{fig:auc_prodet} show AUC heatmaps across all eight swap methods (rows) and seven manipulation regions (columns) for five detectors.
Each cell is the AUC on that specific (method, region) pair from Protocol~\#3.
Our 12-Block detector (Figure~\ref{fig:auc_ours12}) shows the most uniform AUC distribution, sustaining strong performance across all cells.
Specialist detectors such as ProDet (Figure~\ref{fig:auc_prodet}) achieve high AUC on full face and composite regions but degrade sharply for single-region (Eye, Nose, Mouth) fakes, revealing that their decision boundaries are calibrated for global forgery signals.
FatFormer (Figure~\ref{fig:auc_fatformer}) achieves high AUC on E4S across all regions due to its style-transfer artefacts being globally visible, but collapses on BlendFace entirely.


\begin{figure}[htbp]
    \centering
\includegraphics[width=\linewidth,height=0.9\textheight,keepaspectratio]{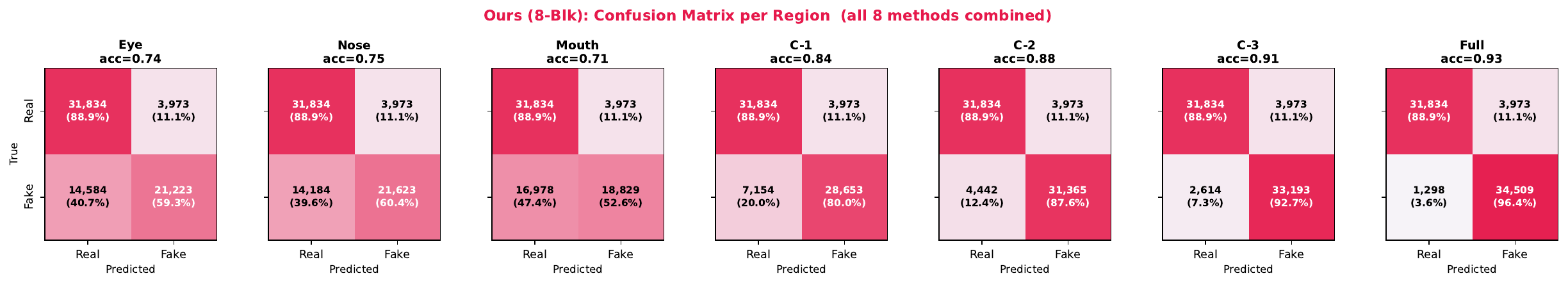}
    \caption{Confusion matrix: \textbf{Ours (8-Blk)}. Most balanced FP/FN across all regions.}
    \label{fig:cm_ours8}
\end{figure}

\begin{figure}[htbp]
    \centering
\includegraphics[width=\linewidth,height=0.9\textheight,keepaspectratio]{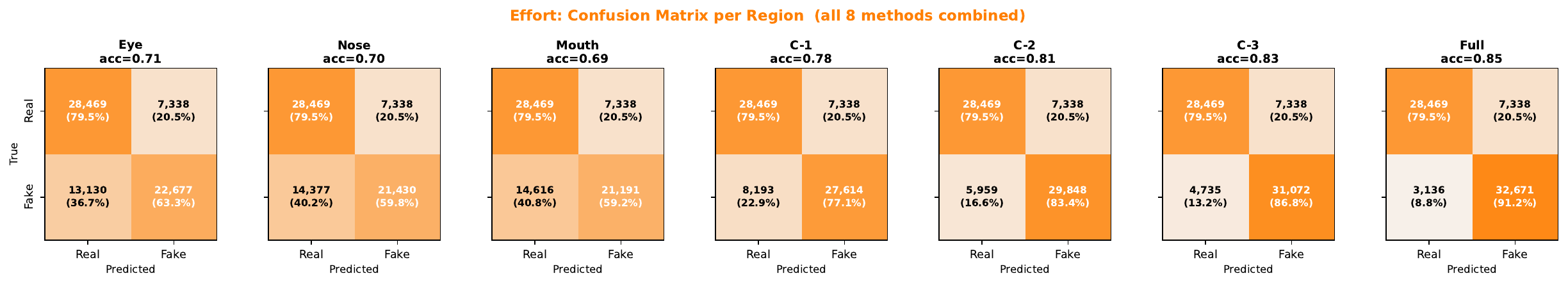}
    \caption{Confusion matrix: \textbf{Effort}. High false-negative rate on single-region fakes.}
    \label{fig:cm_effort}
\end{figure}

\begin{figure}[htbp]
    \centering
\includegraphics[width=\linewidth,height=0.9\textheight,keepaspectratio]{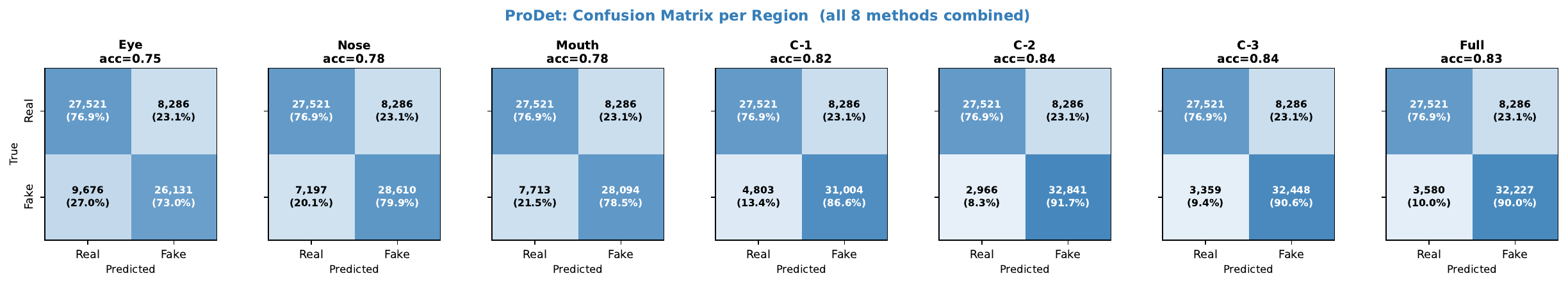}
    \caption{Confusion matrix: \textbf{ProDet}. High FN on sub-region fakes, strong on full face.}
    \label{fig:cm_prodet}
\end{figure}

\begin{figure}[htbp]
    \centering
\includegraphics[width=\linewidth,height=0.9\textheight,keepaspectratio]{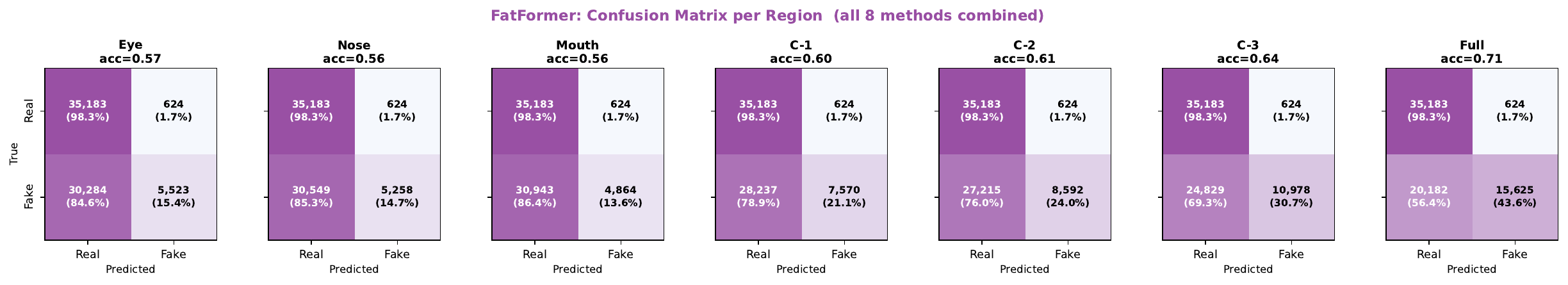}
    \caption{Confusion matrix: \textbf{FatFormer}. Elevated false-positive rate on real images.}
    \label{fig:cm_fatformer}
\end{figure}

\noindent\textbf{Confusion Matrices:}
Figures~\ref{fig:cm_ours8}--\ref{fig:cm_fatformer} show confusion matrices on the combined RLDF test set.
Our 8-Block model achieves the most balanced true positive and true negative rates.
ProDet and Effort exhibit high false-negative rates on region-level fakes consistently predicting partial-face manipulations as real consistent with training and design for full face forgery only.
FatFormer shows an elevated false-positive rate, flagging many real images as fake, likely because its frequency-sensitive features are triggered by natural facial texture variation.


\begin{figure}[htbp]
    \centering
\includegraphics[width=0.95\linewidth,height=0.9\textheight,keepaspectratio]{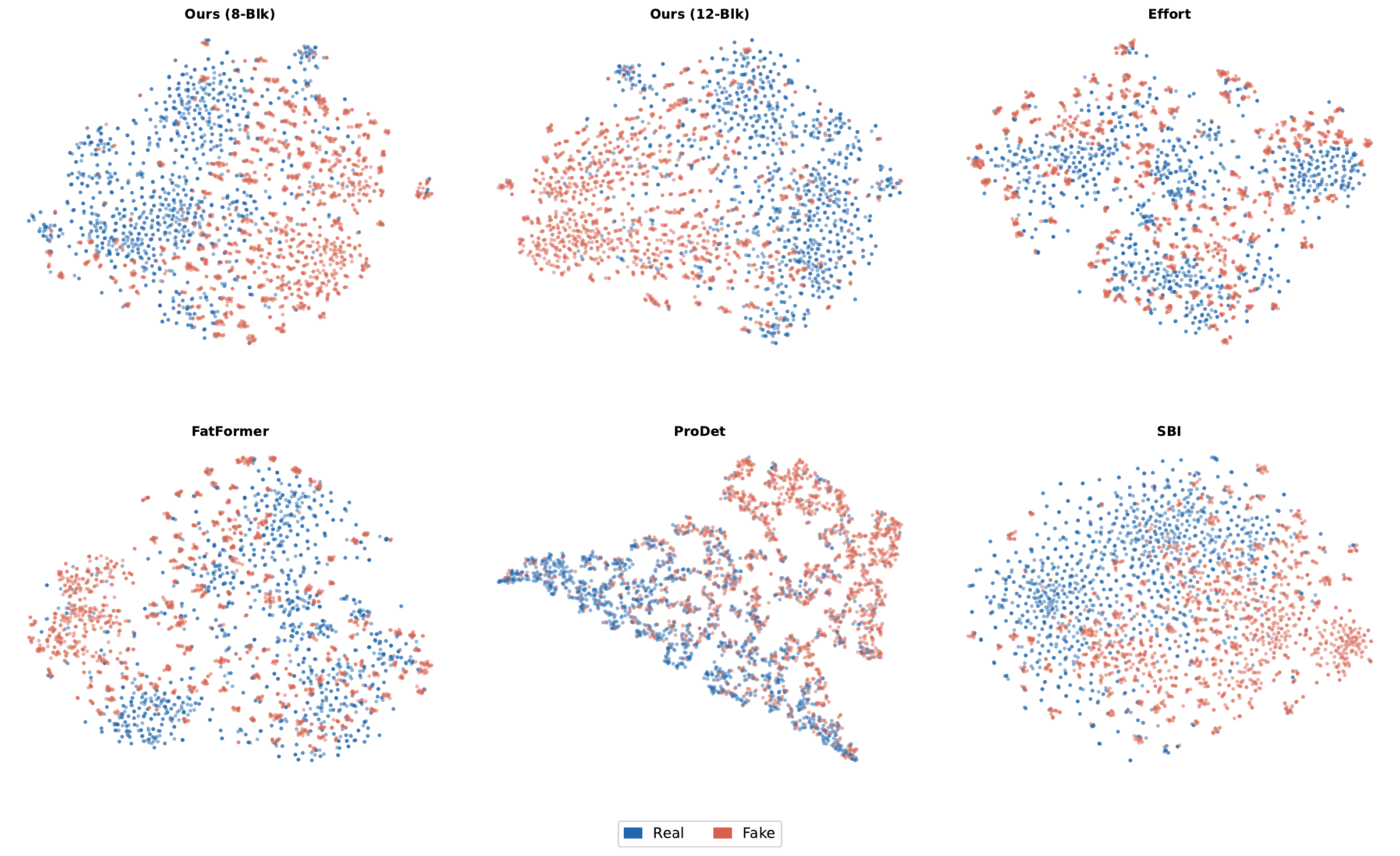}
    \caption{t-SNE of penultimate features on the RLDF test set. \textcolor{blue}{Blue}~=~Real, \textcolor{red}{Red}~=~Fake. Our models achieve cleaner separation than competitors.}
    \label{fig:supp_tsne_det}
\end{figure}

\begin{sidewaysfigure}
    \centering
\includegraphics[width=\linewidth,height=0.92\textwidth,keepaspectratio]{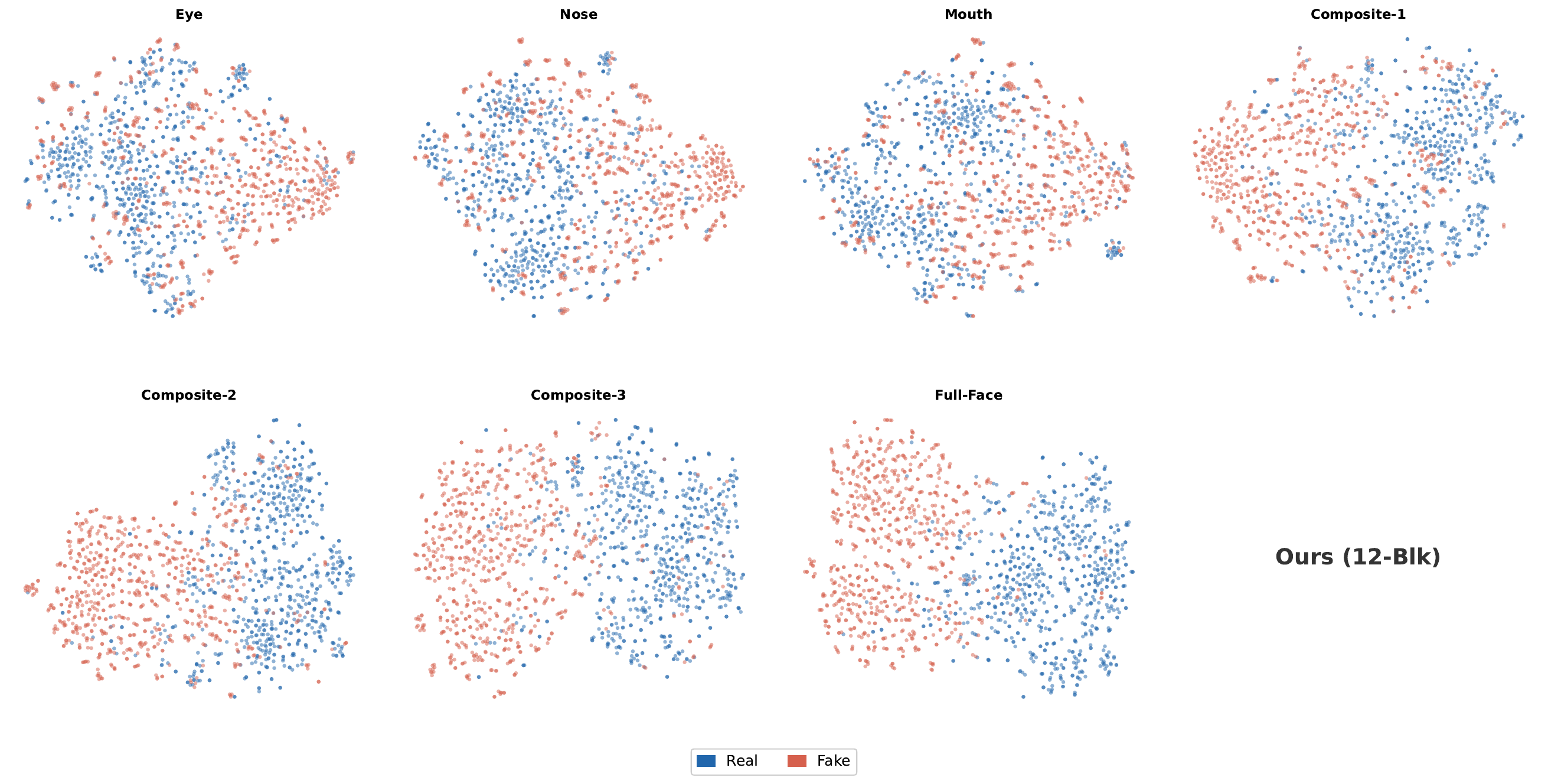}
    \caption{\textbf{Ours (12-Blk)}: Per-region t-SNE visualization}
    \label{fig:supp_tsne_ours12}
\end{sidewaysfigure}

\begin{sidewaysfigure}
    \centering
\includegraphics[width=\linewidth,height=0.92\textwidth,keepaspectratio]{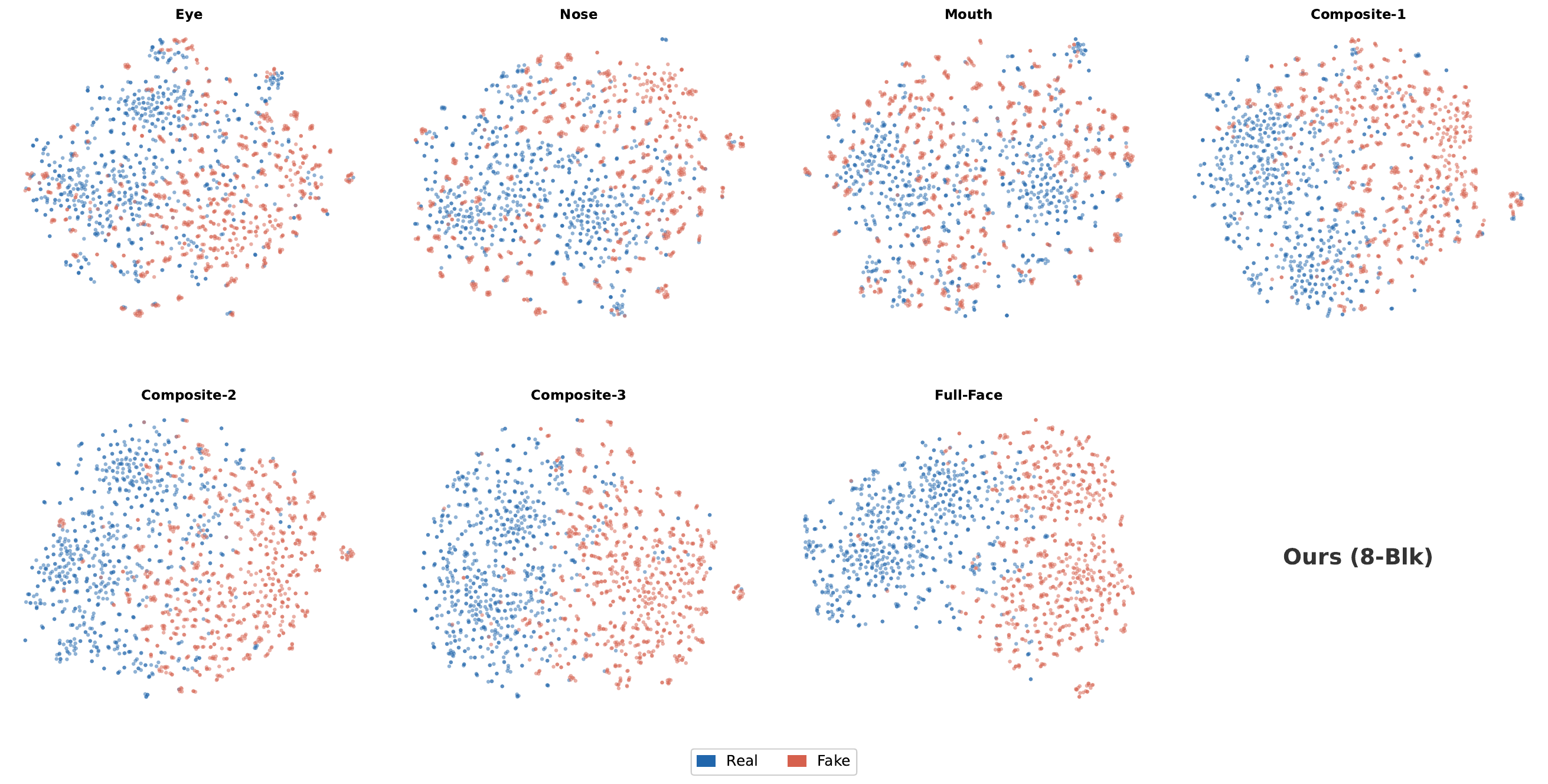}
    \caption{\textbf{Ours (8-Blk)}: Per-region t-SNE visualization}
    \label{fig:supp_tsne_ours8}
\end{sidewaysfigure}

\begin{sidewaysfigure}
    \centering
\includegraphics[width=\linewidth,height=0.92\textwidth,keepaspectratio]{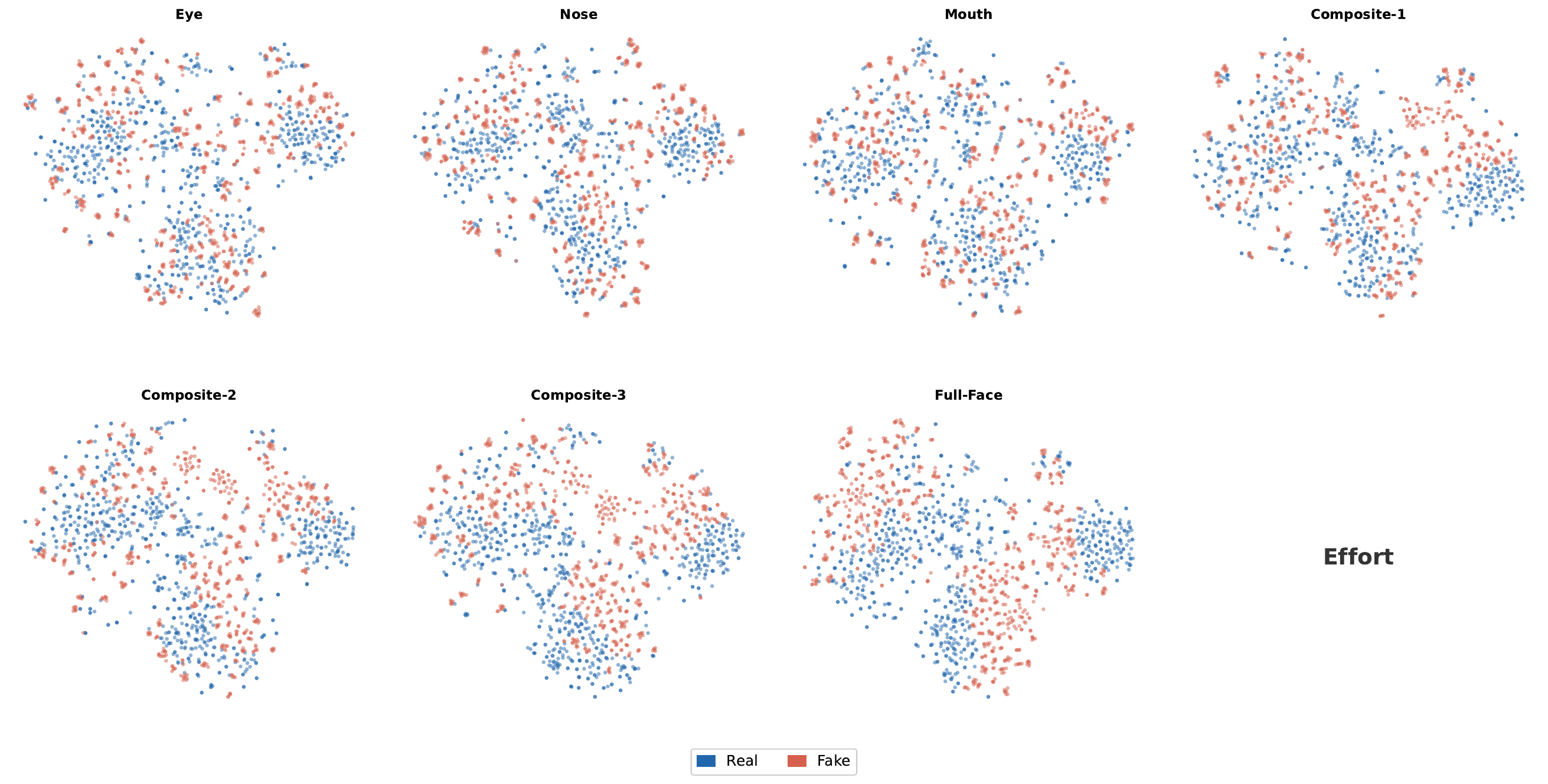}
    \caption{\textbf{Effort}: Per-region t-SNE visualization}
    \label{fig:supp_tsne_effort}
\end{sidewaysfigure}

\begin{sidewaysfigure}
    \centering
\includegraphics[width=\linewidth,height=0.92\textwidth,keepaspectratio]{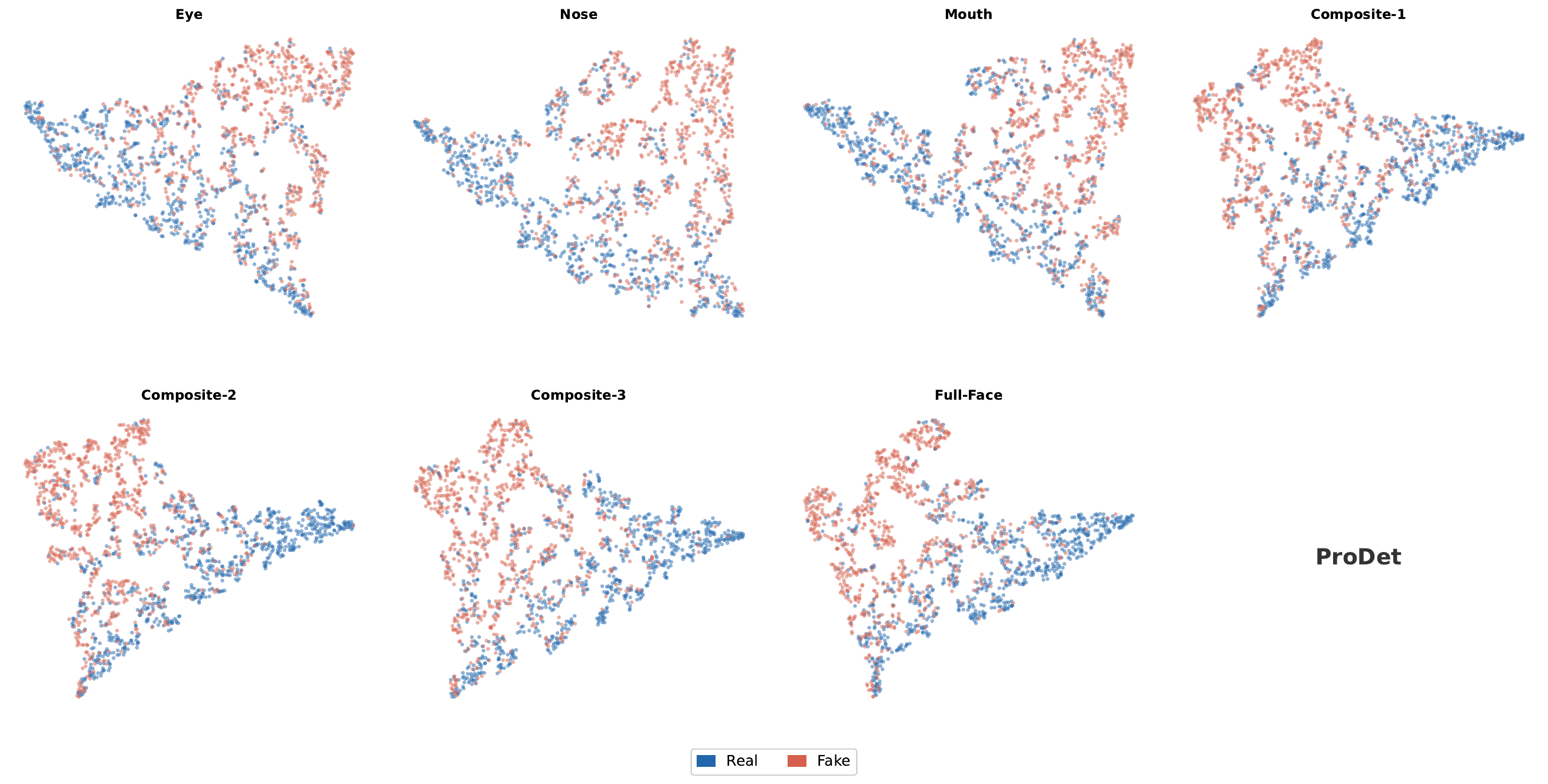}
    \caption{\textbf{ProDet}: Per-region t-SNE visualization}
    \label{fig:supp_tsne_prodet}
\end{sidewaysfigure}

\begin{sidewaysfigure}
    \centering
\includegraphics[width=\linewidth,height=0.92\textwidth,keepaspectratio]{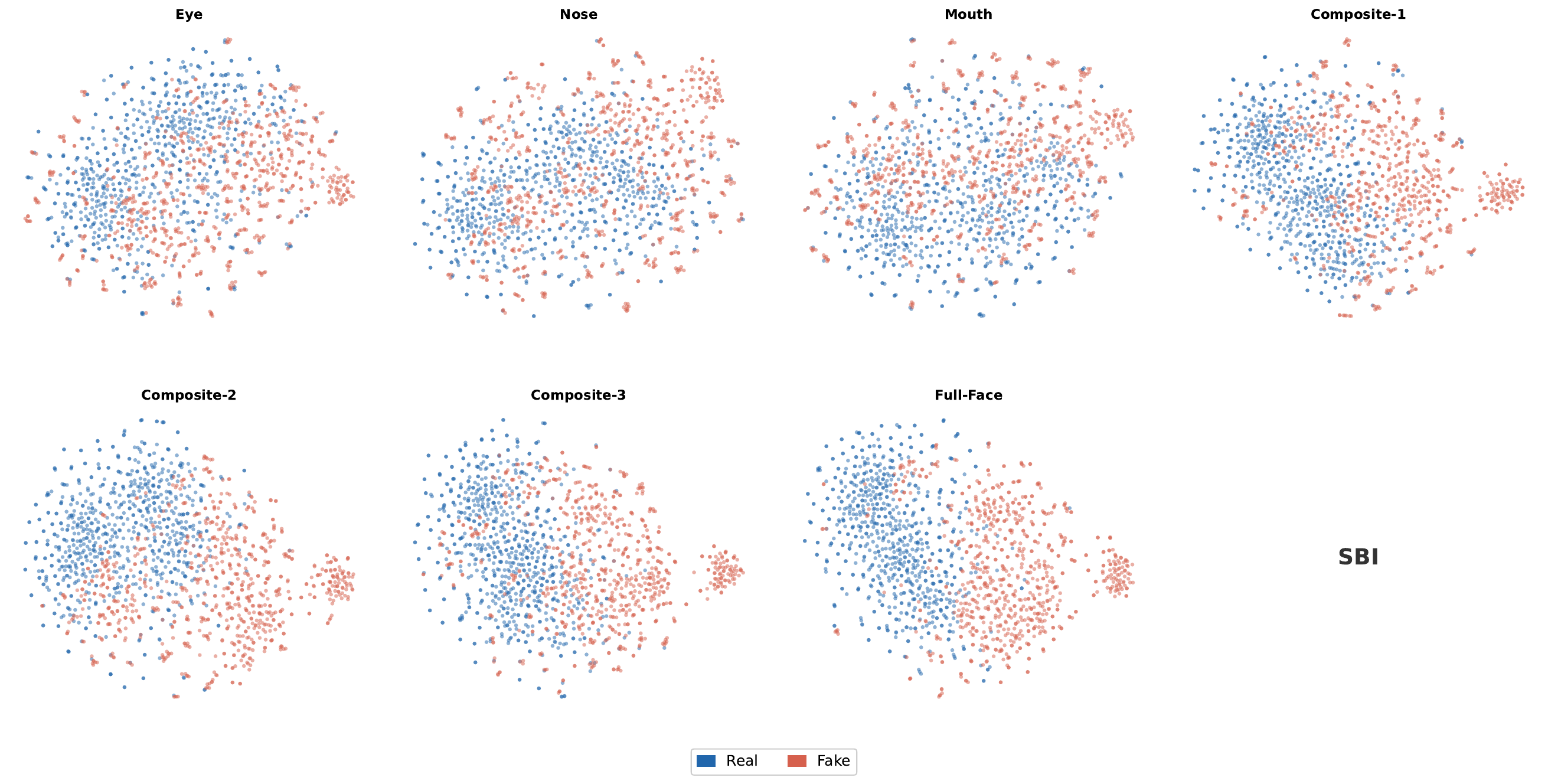}
    \caption{\textbf{SBI}: Per-region t-SNE visualization}
    \label{fig:supp_tsne_sbi}
\end{sidewaysfigure}

\begin{sidewaysfigure}
    \centering
\includegraphics[width=\linewidth,height=0.92\textwidth,keepaspectratio]{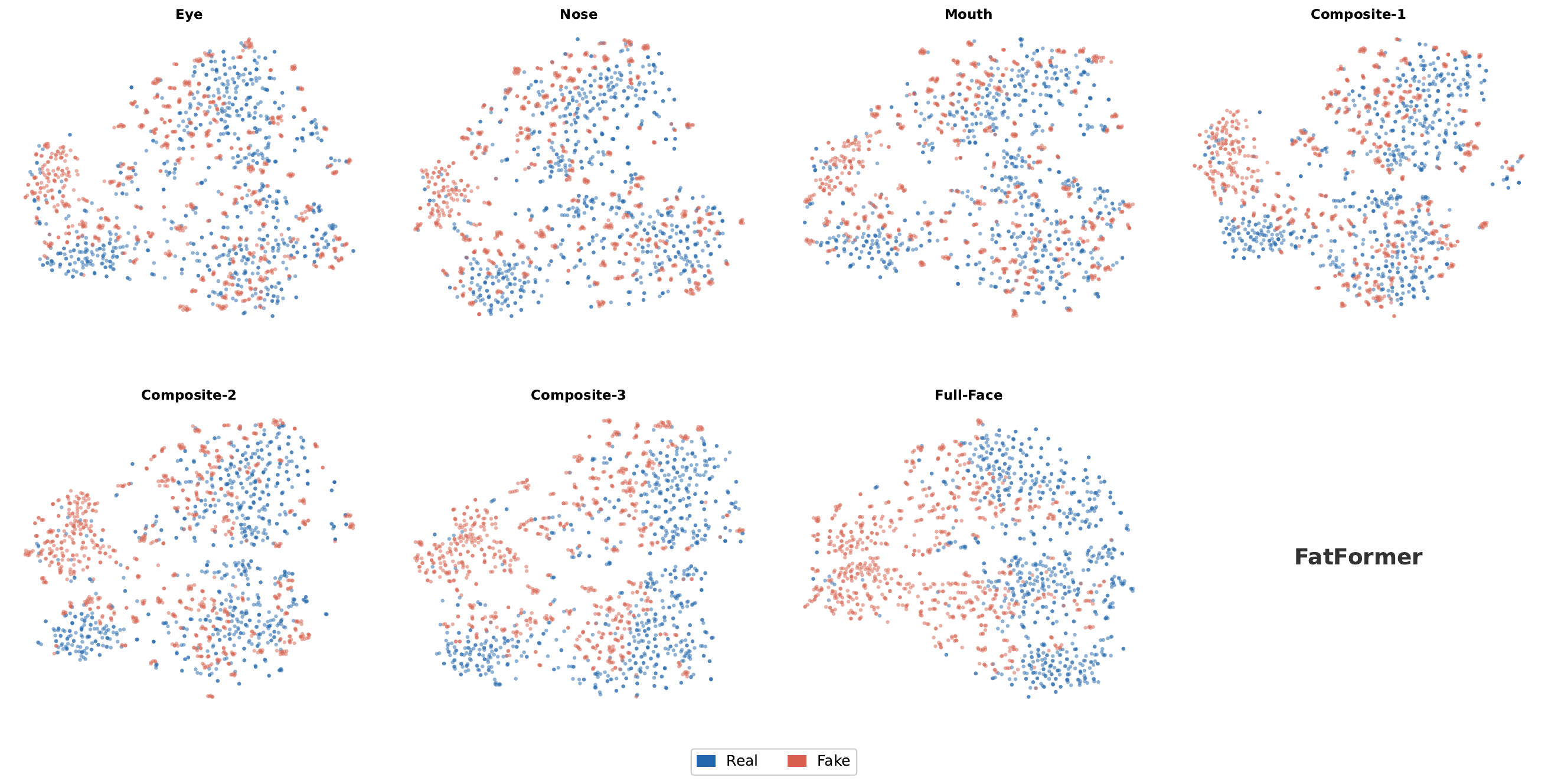}
    \caption{\textbf{FatFormer}: Per-region t-SNE visualization}
    \label{fig:supp_tsne_fatformer}
\end{sidewaysfigure}

\noindent\textbf{Feature Space Analysis (t-SNE):}
Figure~\ref{fig:supp_tsne_det} compares penultimate-layer t-SNE projections across all detectors on the RLDF test set.
Our models show cleaner real/fake separation than competitor methods, particularly for composite and full face regions.
Figures~\ref{fig:supp_tsne_ours12}--\ref{fig:supp_tsne_fatformer} show per-region breakdowns for each detector: separation quality degrades as region size decreases for all methods, but degrades most sharply for Effort and FatFormer at the Eye region, where their features are not spatially grounded.
\vspace{-1mm}


% ── E. MaskNet Spatial Analysis and Score Distributions ───────────────────────
\clearpage
\raggedbottom
\vspace{2mm}
\subsection*{H.\; MaskNet Spatial Analysis and Score Distributions}
\label{ap:G}
\begin{sidewaysfigure}
    \centering
\begin{minipage}[t]{0.50\linewidth}\centering
    \includegraphics[width=\linewidth,height=0.30\textwidth,keepaspectratio]{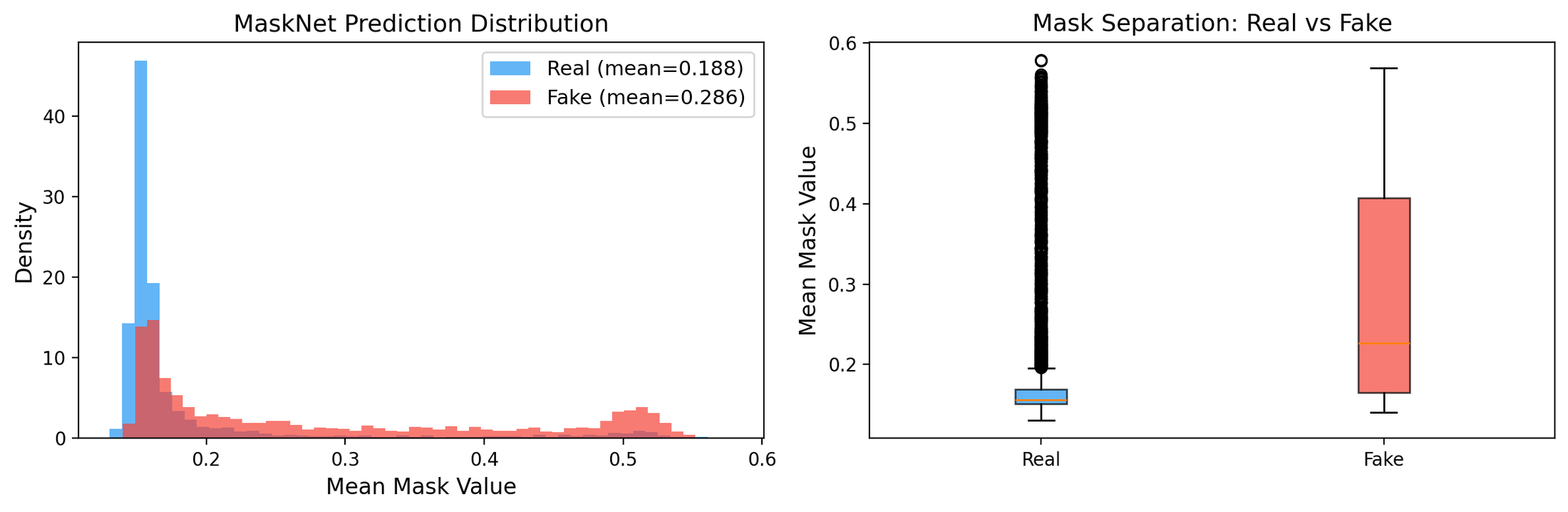}\\{\small Eye}
  \end{minipage}\hfill
  \begin{minipage}[t]{0.50\linewidth}\centering
    \includegraphics[width=\linewidth,height=0.30\textwidth,keepaspectratio]{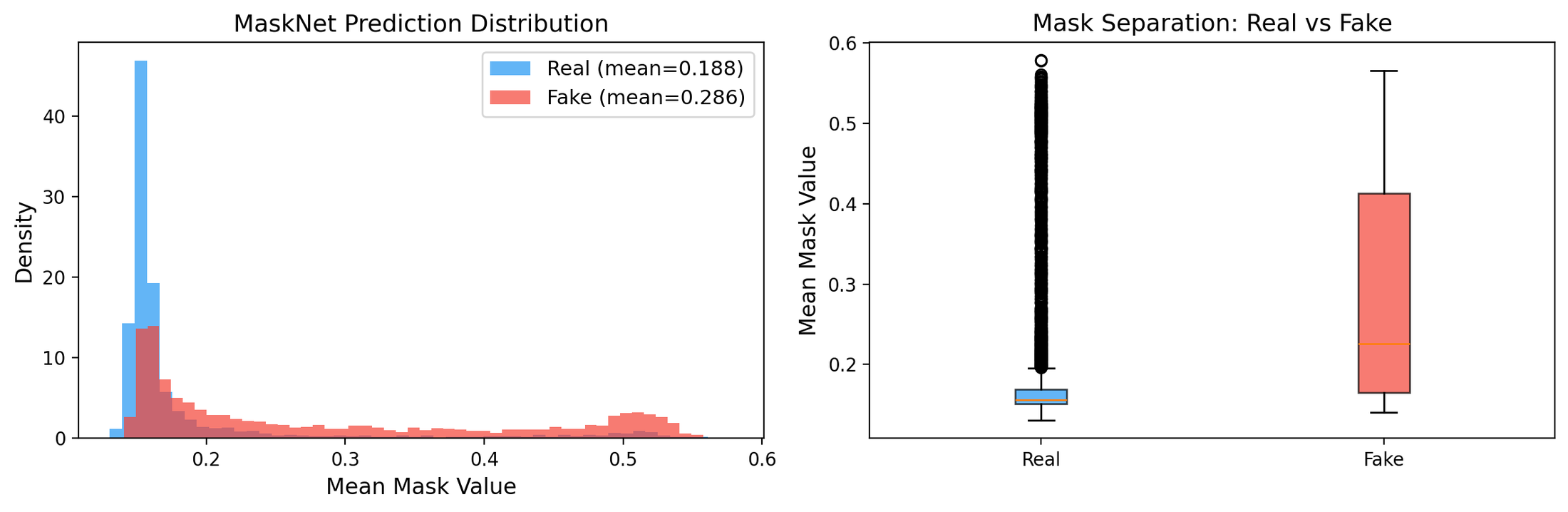}\\{\small Nose}
  \end{minipage}\\[6pt]
  \begin{minipage}[t]{0.50\linewidth}\centering
    \includegraphics[width=\linewidth,height=0.30\textwidth,keepaspectratio]{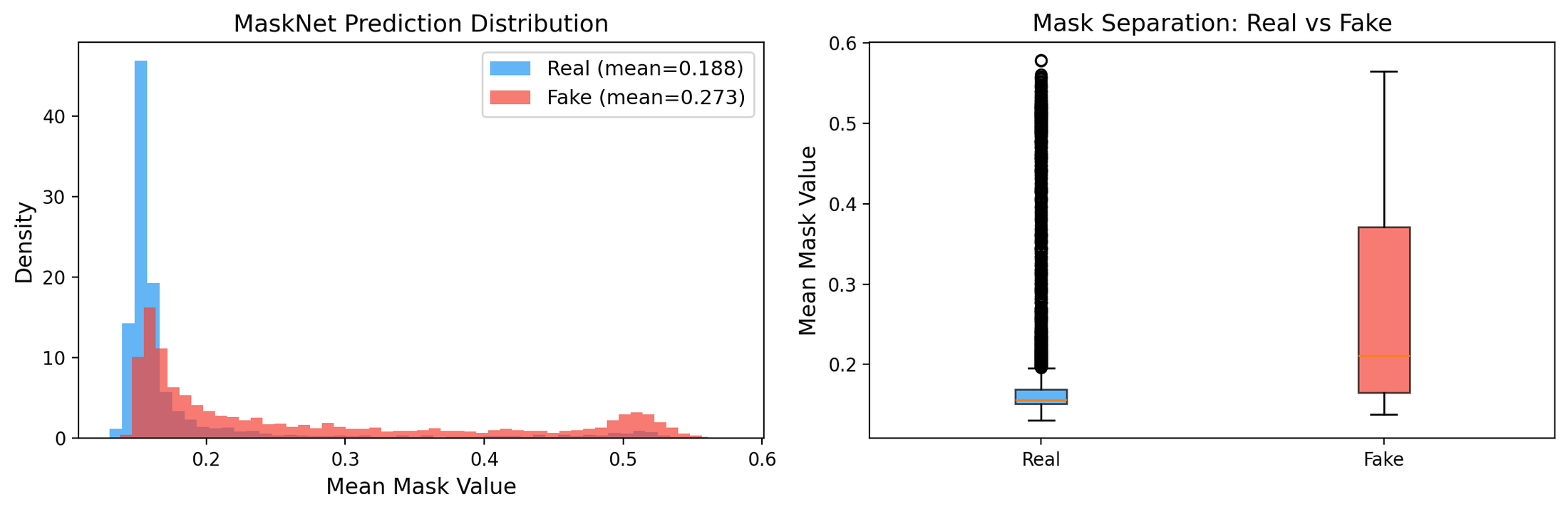}\\{\small Mouth}
  \end{minipage}\hfill
  \begin{minipage}[t]{0.50\linewidth}\centering
    \includegraphics[width=\linewidth,height=0.30\textwidth,keepaspectratio]{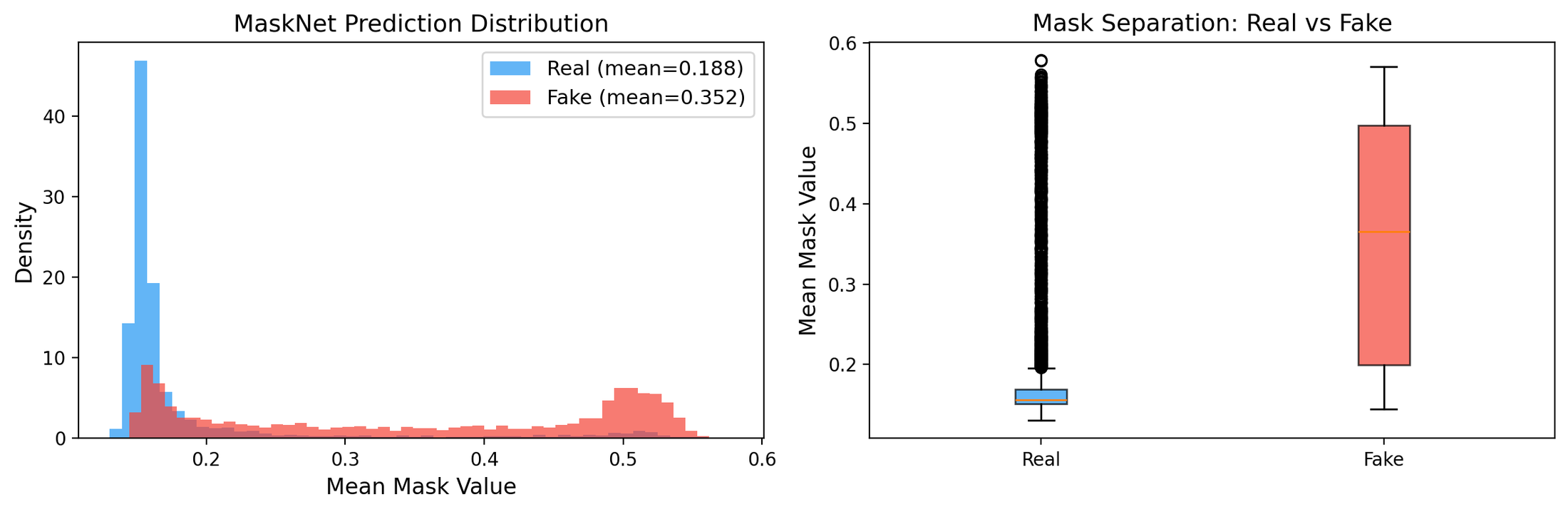}\\{\small Composite-1 (Eye+Nose)}
  \end{minipage}\\[6pt]
  \begin{minipage}[t]{0.50\linewidth}\centering
    \includegraphics[width=\linewidth,height=0.30\textwidth,keepaspectratio]{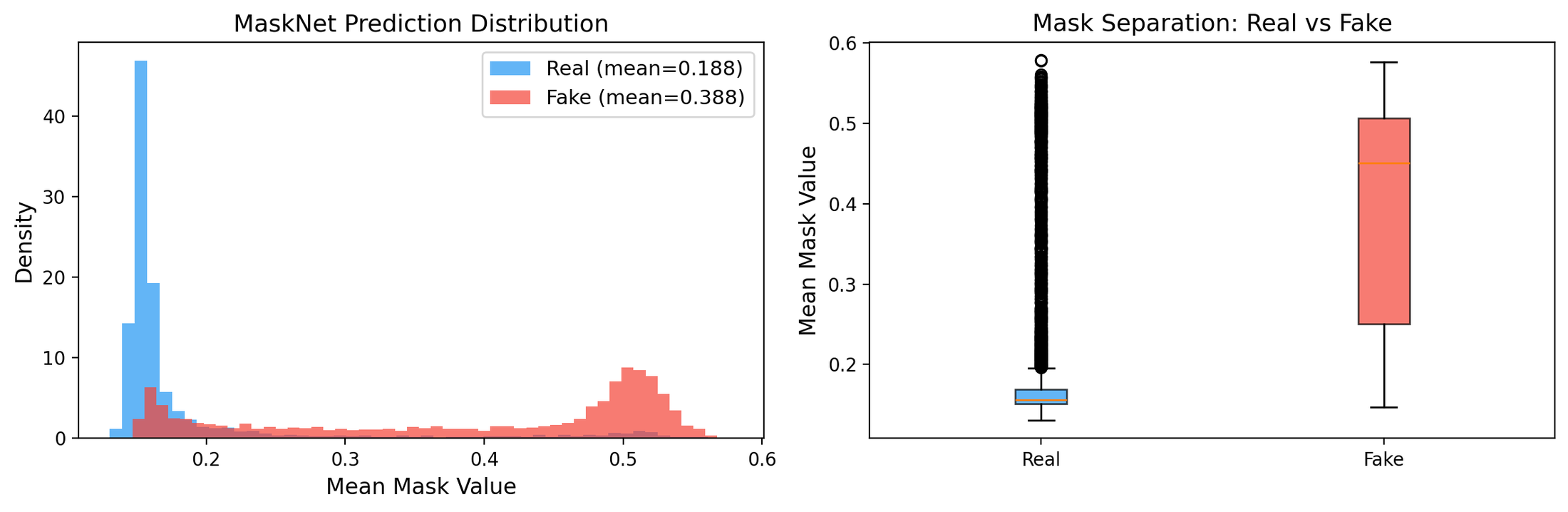}\\{\small Composite-2 (+Mouth)}
  \end{minipage}\hfill
  \begin{minipage}[t]{0.50\linewidth}\centering
    \includegraphics[width=\linewidth,height=0.30\textwidth,keepaspectratio]{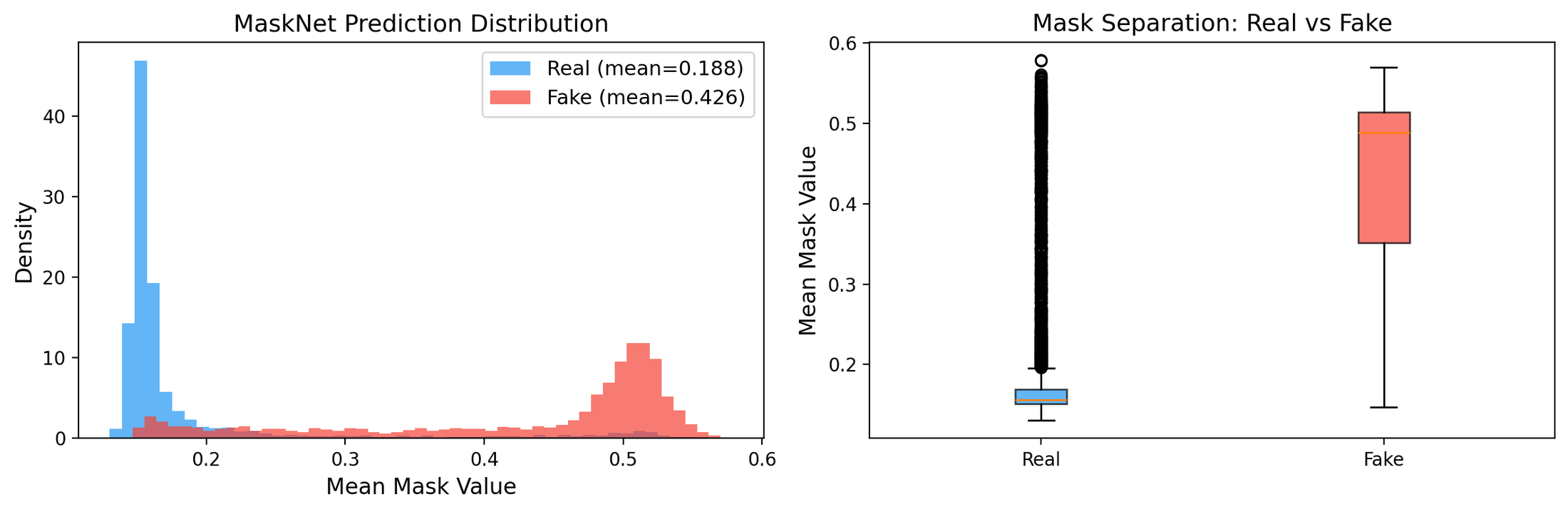}\\{\small Composite-3 (+Eyebrow)}
  \end{minipage}\\[6pt]
  \begin{minipage}[t]{0.50\linewidth}\centering
    \includegraphics[width=\linewidth,height=0.30\textwidth,keepaspectratio]{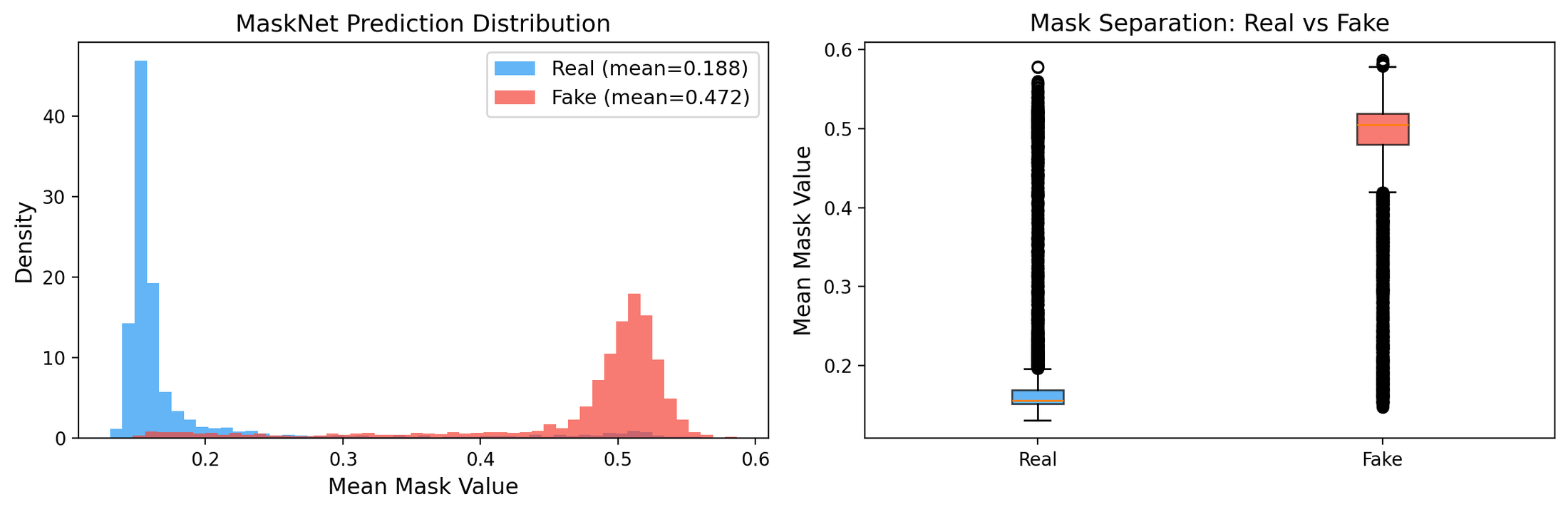}\\{\small Full Face}
  \end{minipage}
  \vspace{4mm}
  \caption{\textbf{MaskNet activation: BlendFace across all 7 regions.}
  Each panel is the mean per-patch forgery score $\hat{\mathbf{m}}$ heatmap.
  Without any spatial supervision, MaskNet progressively expands its focus as the manipulated region grows.}
  \label{fig:mask_region_prog}
\end{sidewaysfigure}

\begin{sidewaysfigure}

    \centering
% ══════════════════════════════════════════════════════════════════════════
  % TOP HALF — Eye region across all 8 swap methods
  % ══════════════════════════════════════════════════════════════════════════

  % Row 1
  \begin{minipage}[t]{0.32\linewidth}\centering
    \includegraphics[width=\linewidth,height=0.30\textwidth,keepaspectratio]{images/blendface_eye_mask.png}\\[4pt]
    {\small BlendFace}
  \end{minipage}\hfill
  \begin{minipage}[t]{0.32\linewidth}\centering
    \includegraphics[width=\linewidth,height=0.30\textwidth,keepaspectratio]{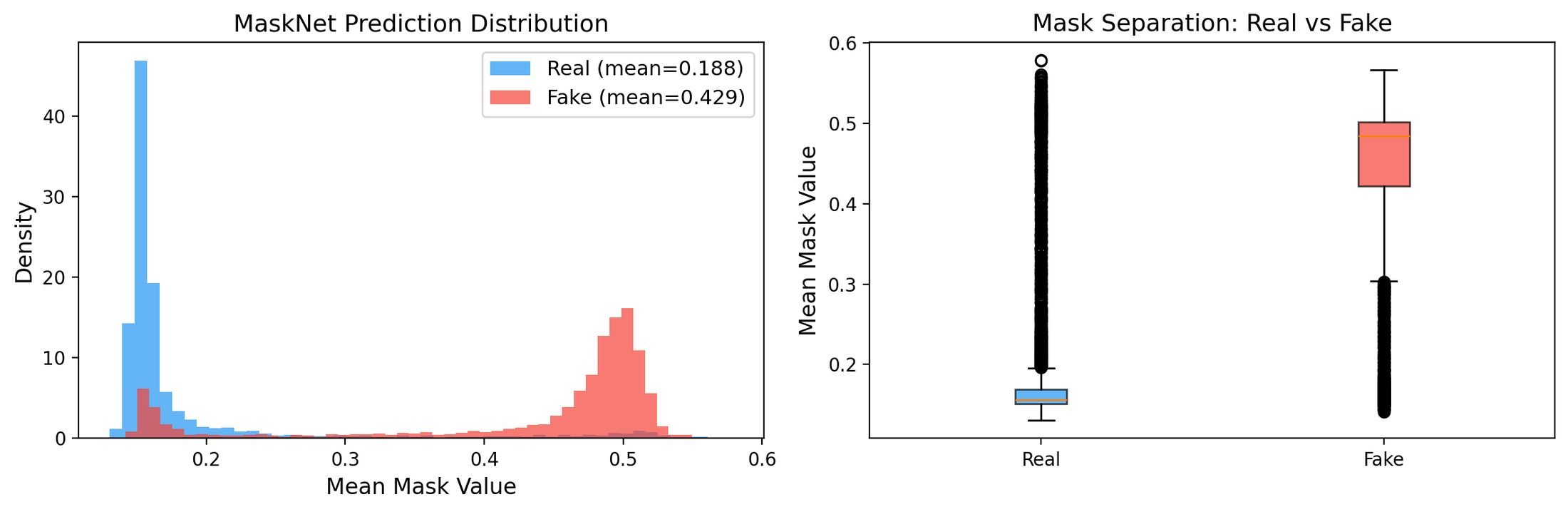}\\[4pt]
    {\small E4S}
  \end{minipage}\hfill
  \begin{minipage}[t]{0.32\linewidth}\centering
    \includegraphics[width=\linewidth,height=0.30\textwidth,keepaspectratio]{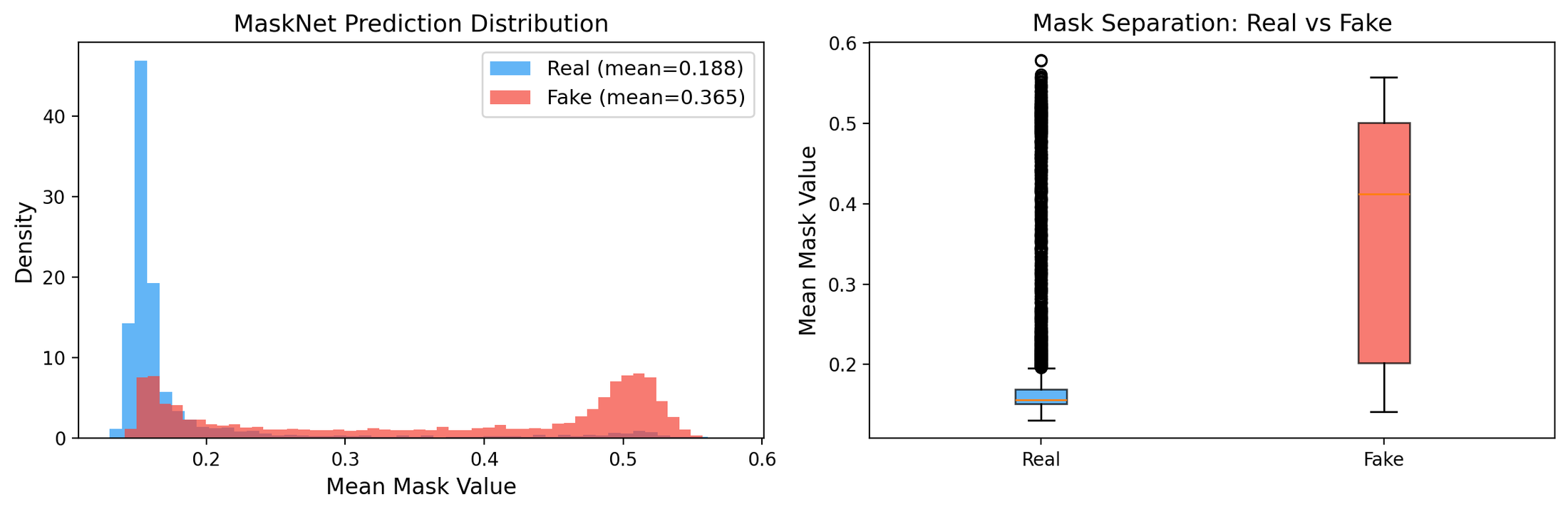}\\[4pt]
    {\small FaceDancer}
  \end{minipage}

  \vspace{14pt}

  % Row 2
  \begin{minipage}[t]{0.32\linewidth}\centering
    \includegraphics[width=\linewidth,height=0.30\textwidth,keepaspectratio]{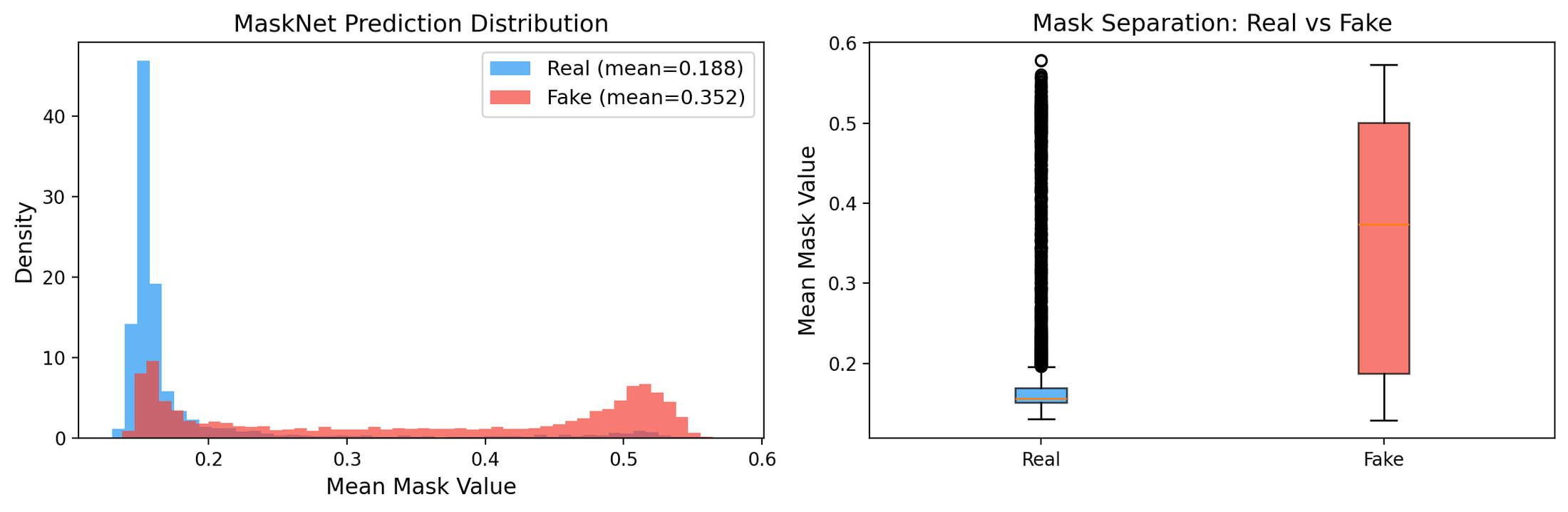}\\[4pt]
    {\small FSGAN}
  \end{minipage}\hfill
  \begin{minipage}[t]{0.32\linewidth}\centering
    \includegraphics[width=\linewidth,height=0.30\textwidth,keepaspectratio]{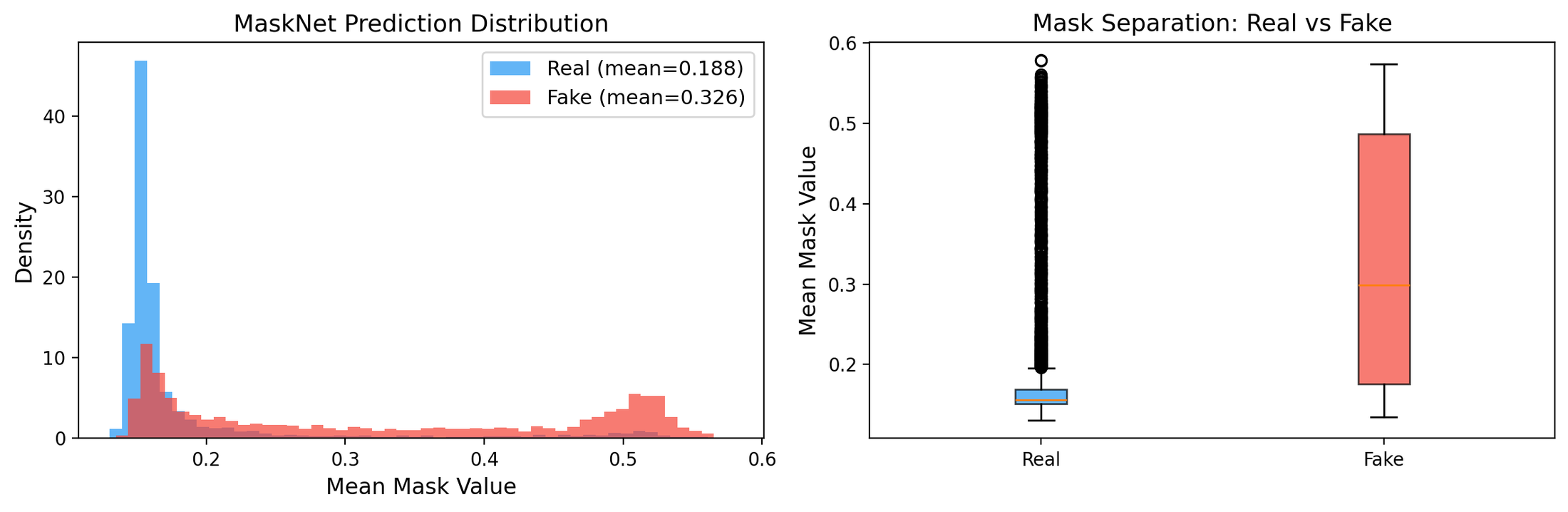}\\[4pt]
    {\small GHOST}
  \end{minipage}\hfill
  \begin{minipage}[t]{0.32\linewidth}\centering
    \includegraphics[width=\linewidth,height=0.30\textwidth,keepaspectratio]{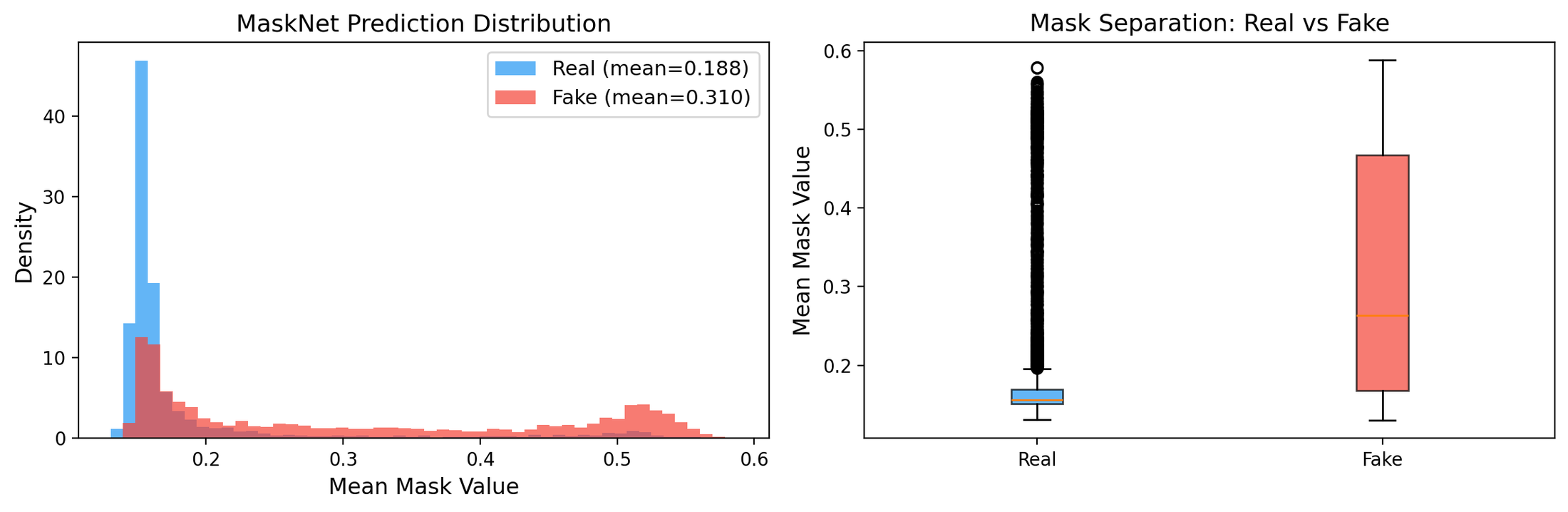}\\[4pt]
    {\small InSwapper}
  \end{minipage}

  \vspace{14pt}

  % Row 3 — centred pair
  \makebox[\linewidth][c]{%
    \begin{minipage}[t]{0.32\linewidth}\centering
      \includegraphics[width=\linewidth,height=0.30\textwidth,keepaspectratio]{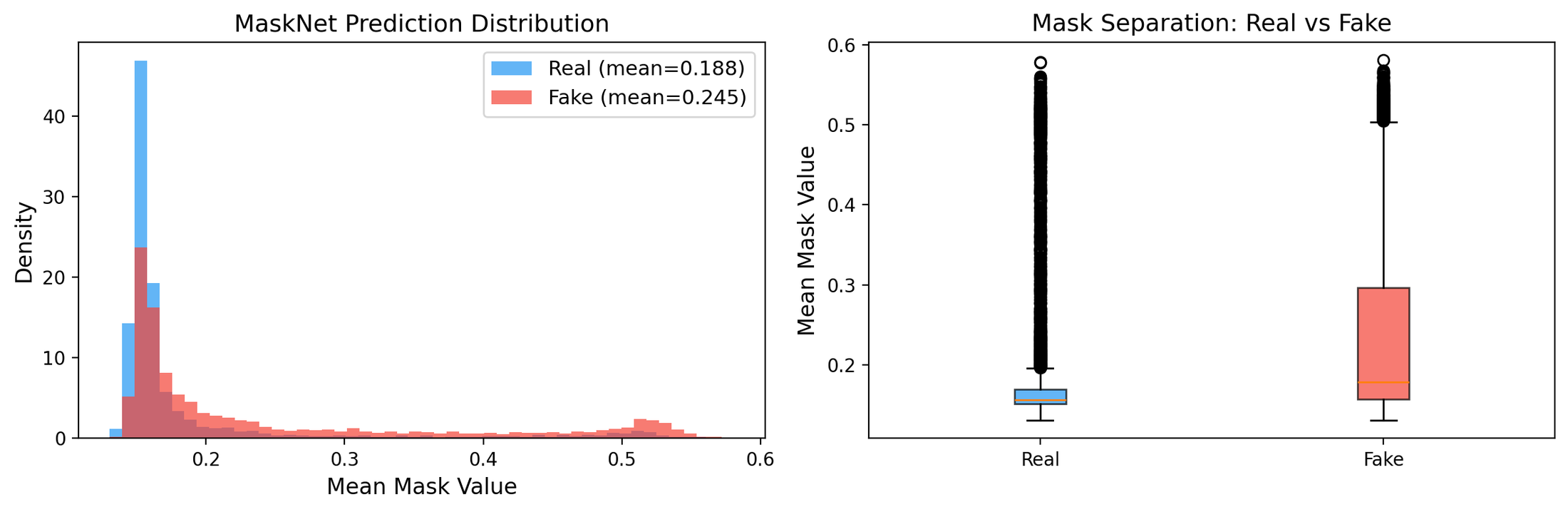}\\[4pt]
      {\small SimSwap}
    \end{minipage}\hspace{0.04\linewidth}%
    \begin{minipage}[t]{0.32\linewidth}\centering
      \includegraphics[width=\linewidth,height=0.30\textwidth,keepaspectratio]{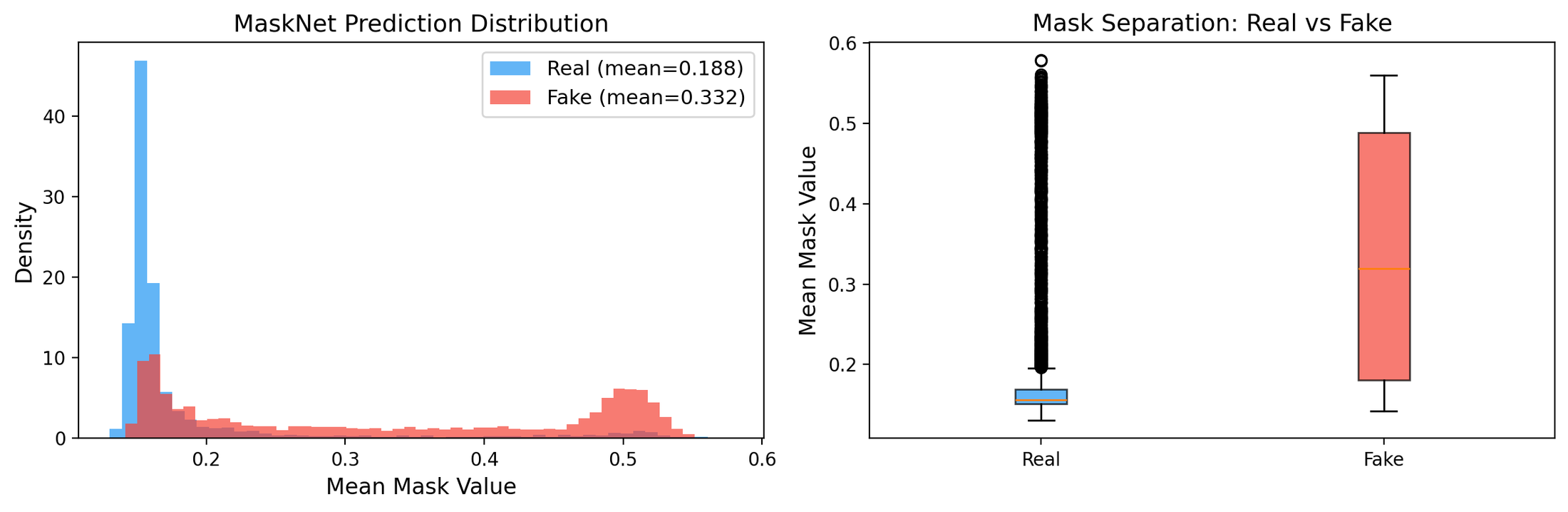}\\[4pt]
      {\small UniFace}
    \end{minipage}%
  }

  \vspace{4pt}
  \captionof{figure}{\textbf{MaskNet activation: Eye region across all 8 swap
  methods.} All methods show consistent eye-patch focus. Activation intensity
  correlates with detectability: BlendFace (AUC~0.85) produces diffuse masks,
  while E4S and FaceDancer (AUC~$\geq$~0.96) yield sharp, localised activations.}
  \label{fig:mask_method_eye}

  \vspace{8pt}
  
  \vspace{8pt}
\end{sidewaysfigure}

\vspace{-1mm}
\noindent\textbf{MaskNet Spatial Analysis.}
Figures~\ref{fig:mask_region_prog} and~\ref{fig:mask_method_eye} show the spatial distribution of per-patch forgery scores $\hat{\mathbf{m}}$ predicted by MaskNet, visualised as $16{\times}16$ heatmaps (one cell per CLIP patch position), averaged over all test images for a given (method, region) combination.

\emph{Region progression} (Figure~\ref{fig:mask_region_prog}): For BlendFace fakes, high mask scores progressively expand as the manipulated region grows from a tight cluster at the eye patches for Eye-only fakes, through overlapping nose/mouth areas for Composite regions, to broad coverage for Full Face fakes.
This spatial alignment emerges \emph{purely from the classification gradient} flowing through the log-space soft mask in Eq.~(3), with no spatial supervision at any point during training.
This directly validates the differentiable gate design: MaskNet and MCA co-adapt to localise the forgery without pixel-level annotation.

\emph{Method variation at Eye region} (Figure~\ref{fig:mask_method_eye}): Across all eight swap methods, MaskNet consistently concentrates activation around the eye patches for Eye-region fakes.
Mask intensity varies by method: BlendFace produces the weakest activation (seam-minimising; hardest to detect, AUC~0.85), while E4S and FaceDancer produce the strongest (style-transfer artefacts; AUC~$\geq$~0.96).
This variation directly predicts per-method detectability, confirming that MaskNet's learned spatial response serves as an implicit indicator of forgery strength.

\begin{figure}[H]
% ══════════════════════════════════════════════════════════════════════════
  % BOTTOM HALF — Cross-dataset benchmarks
  % ══════════════════════════════════════════════════════════════════════════

  % Row 1
  \begin{minipage}[t]{0.48\linewidth}\centering
    \includegraphics[width=\linewidth,height=0.22\textheight,keepaspectratio]{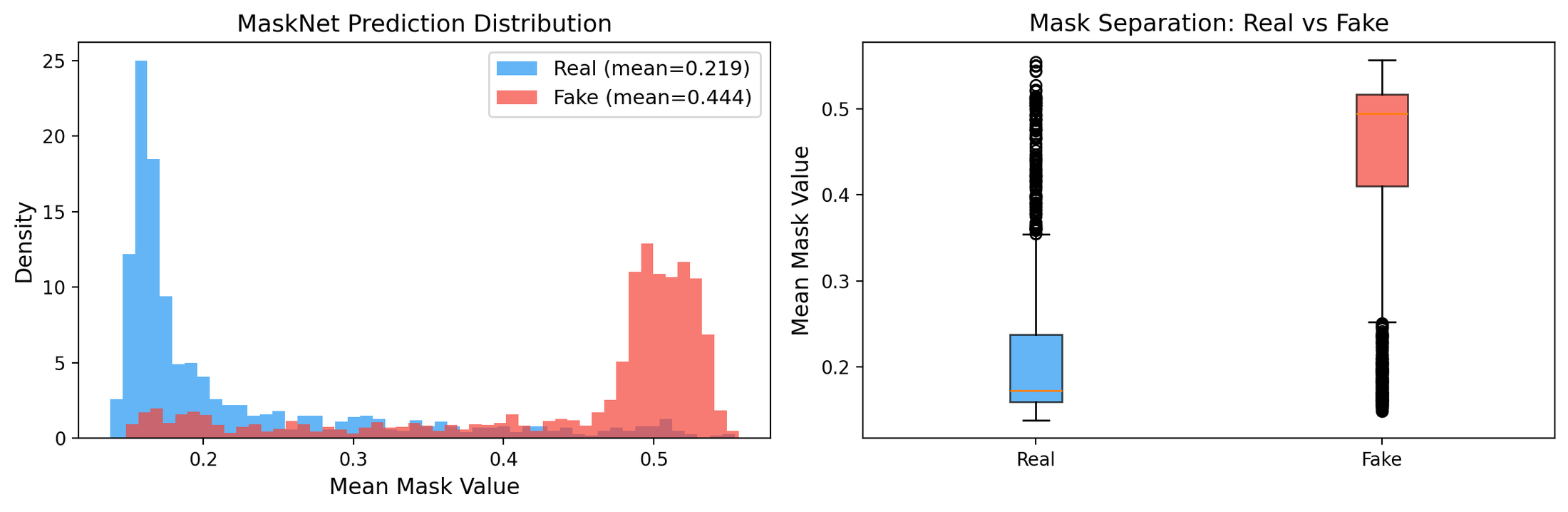}\\[2pt]
    {\small CelebDF-v1}
  \end{minipage}\hfill
  \begin{minipage}[t]{0.48\linewidth}\centering
    \includegraphics[width=\linewidth,height=0.22\textheight,keepaspectratio]{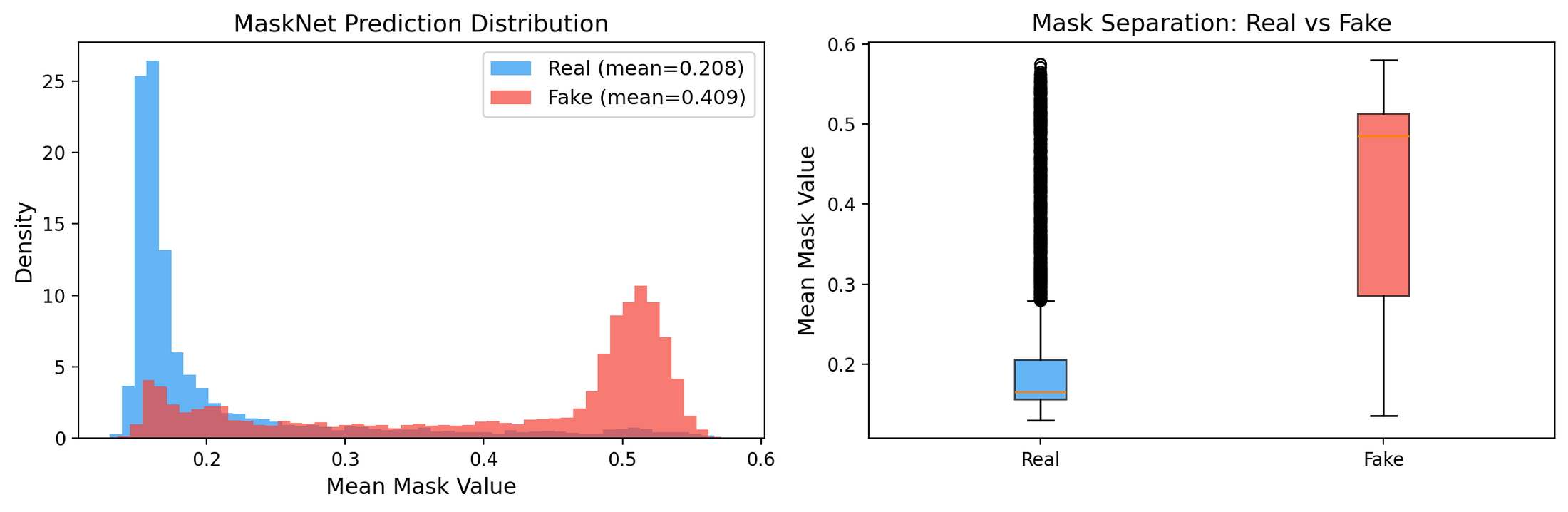}\\[2pt]
    {\small CelebDF-v2}
  \end{minipage}

  \vspace{6pt}

  % Row 2
  \begin{minipage}[t]{0.48\linewidth}\centering
    \includegraphics[width=\linewidth,height=0.22\textheight,keepaspectratio]{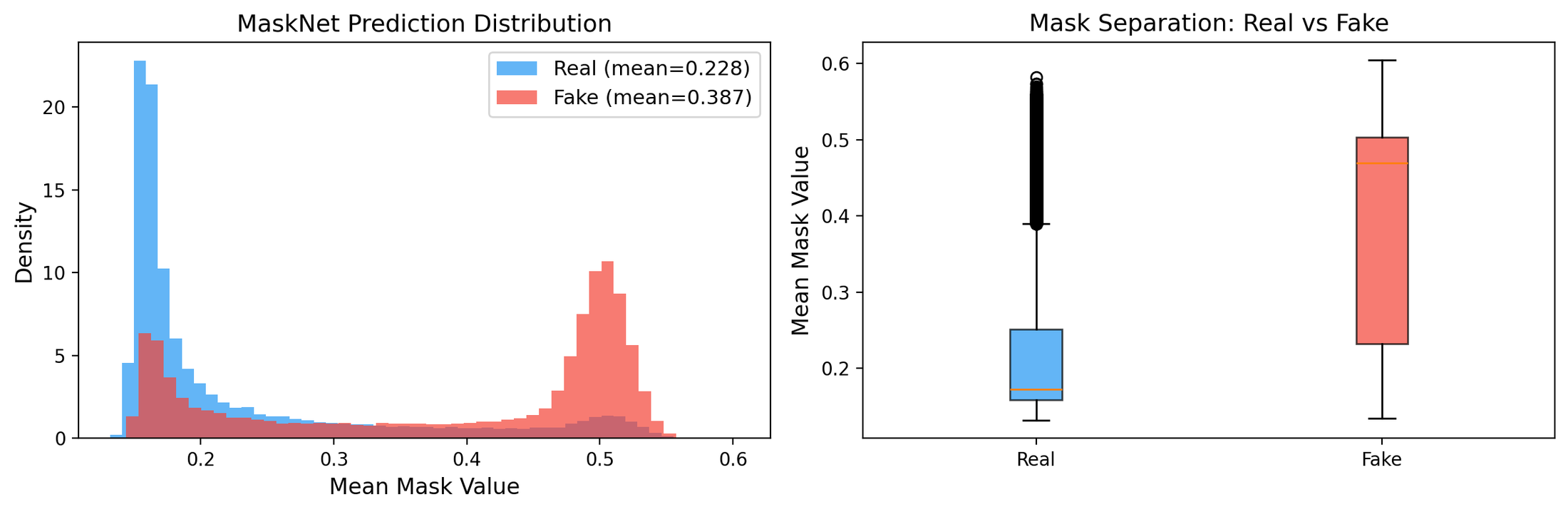}\\[2pt]
    {\small DFDC}
  \end{minipage}\hfill
  \begin{minipage}[t]{0.48\linewidth}\centering
    \includegraphics[width=\linewidth,height=0.22\textheight,keepaspectratio]{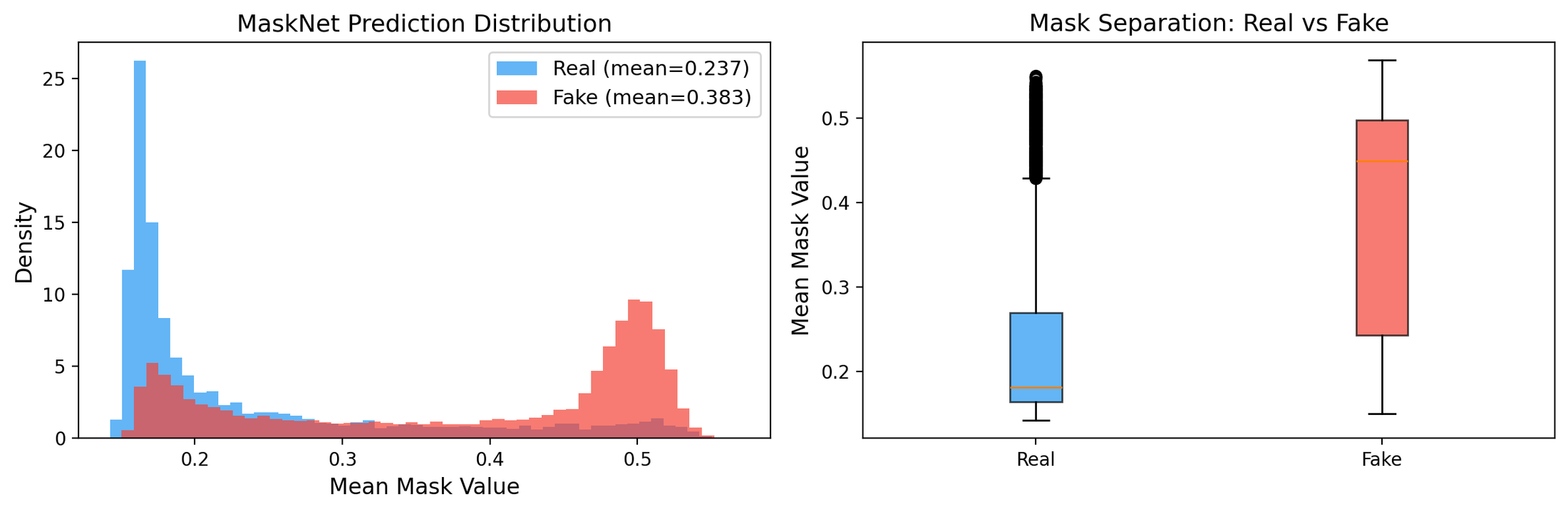}\\[2pt]
    {\small DFDCP}
  \end{minipage}

  \vspace{6pt}

  % Row 3 — centred
  \makebox[\linewidth][c]{%
    \begin{minipage}[t]{0.60\linewidth}\centering
      \includegraphics[width=\linewidth,height=0.22\textheight,keepaspectratio]{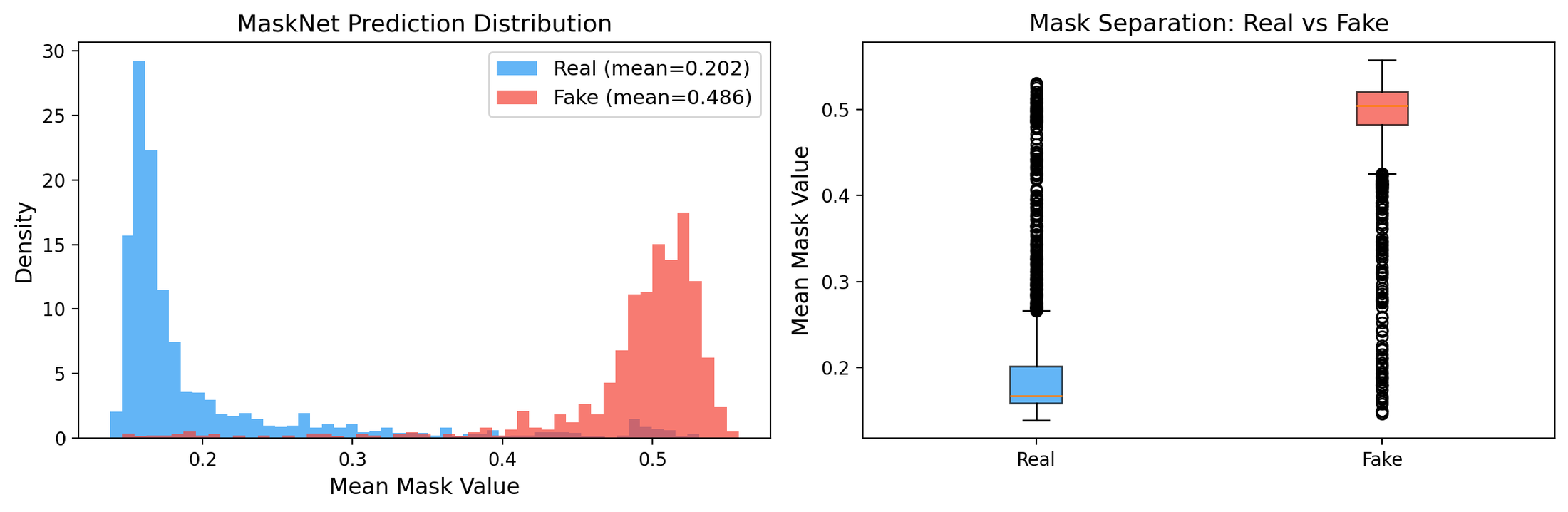}\\[2pt]
      {\small UADFV}
    \end{minipage}%
  }

  \vspace{4pt}
  \captionof{figure}{\textbf{MaskNet activation on unseen cross-dataset
  benchmarks.} Despite training exclusively on FF++, MaskNet produces
  structured, face-centric masks across CelebDF, DFDC, DFDCP, and UADFV,
  consistently activating around eyes, nose, and mouth.}
  \label{fig:mask_crossdataset}
\end{figure}

\emph{Cross-dataset generalization} (Figure~\ref{fig:mask_crossdataset}): On unseen cross-dataset benchmarks (CelebDF-v1/v2, DFDC, DFDCP, UADFV), MaskNet has never seen these images during training yet produces informative spatial masks.
Scores cluster around the central face region eyes, nose bridge, and mouth, which are the dominant loci of identity and the areas most altered by full face swapping methods.
Crucially, the masks do not collapse to uniform noise (which would happen if MaskNet had no forgery-relevant signal), nor do they saturate the entire image (which would indicate over-confident global detection).
Instead they exhibit \emph{structured, face-centric activation} that generalises the spatial prior learned on FF++ training data to entirely different datasets and generation methods, providing qualitative evidence that MaskNet learns a dataset-agnostic notion of ``where faces get swapped'' rather than overfitting to FF++ artefact statistics.

% ── Figure 1: BlendFace across all 7 regions (2 per row) ─────────────────────

% ── Both MaskNet figures — single full page ───────────────────────────────────
\clearpage

\clearpage
% ── E. Per-Method Region-Level Tables ─────────────────────────────────────────
\subsection*{I.\; Per-Method Region-Level Analysis (Protocol~\#3)}
\label{ap:H}
Tables~\ref{tab:supp_blendface}--\ref{tab:supp_uniface} report the complete
per-method, per-region detection performance under Protocol~\#3 across all
13 evaluated detectors and seven manipulation regions
(Eye, Nose, Mouth, Composite-1, Composite-2, Composite-3, Full Face).

Several consistent trends emerge across all eight methods.
First, detection accuracy increases monotonically with region size:
the Eye and Nose columns consistently show the lowest AUC across all detectors
and all methods, while the Full Face column shows the highest.
This confirms that detection difficulty is primarily governed by the spatial
extent of the manipulation rather than the generation method itself.

\begin{figure}[htbp]
    \centering
\includegraphics[width=\linewidth,height=0.9\textheight,keepaspectratio]{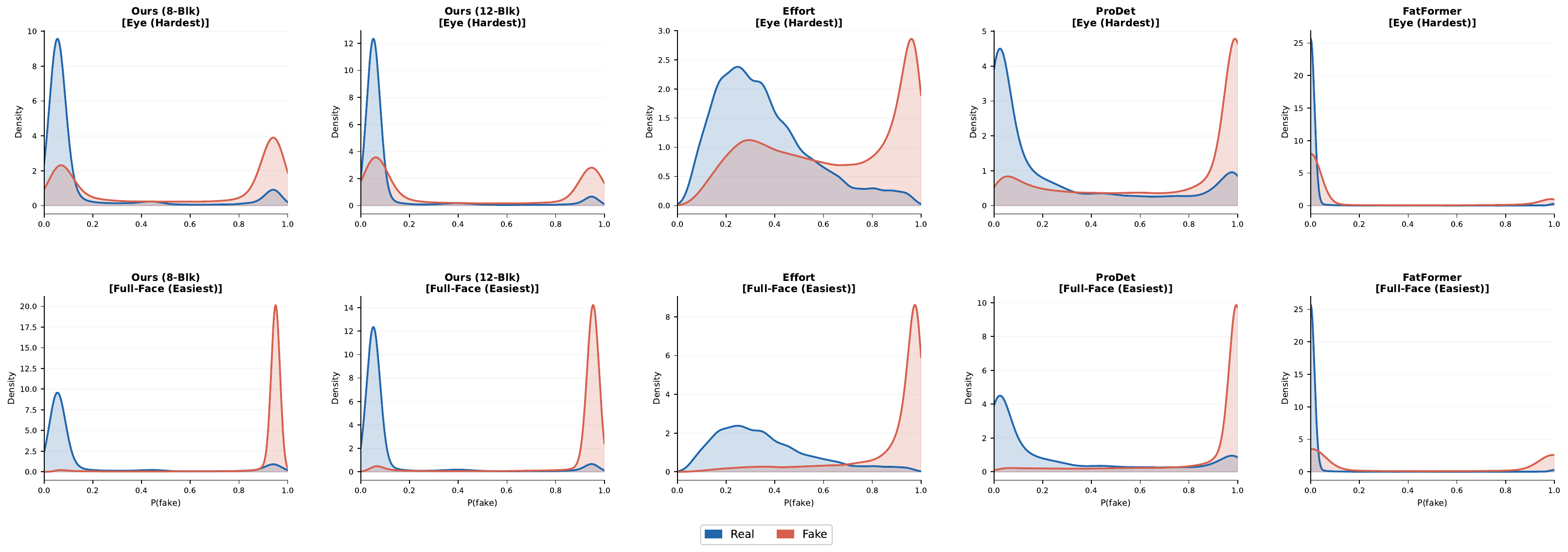}
    \caption{\textbf{Eye vs Full Face} score distributions. Eye shows near-complete real/fake overlap, Full Face shows clear bimodal separation.}
    \label{fig:supp_score_eye_full}
\end{figure}

\begin{sidewaysfigure}
    \centering
\includegraphics[width=\linewidth,height=0.92\textwidth,keepaspectratio]{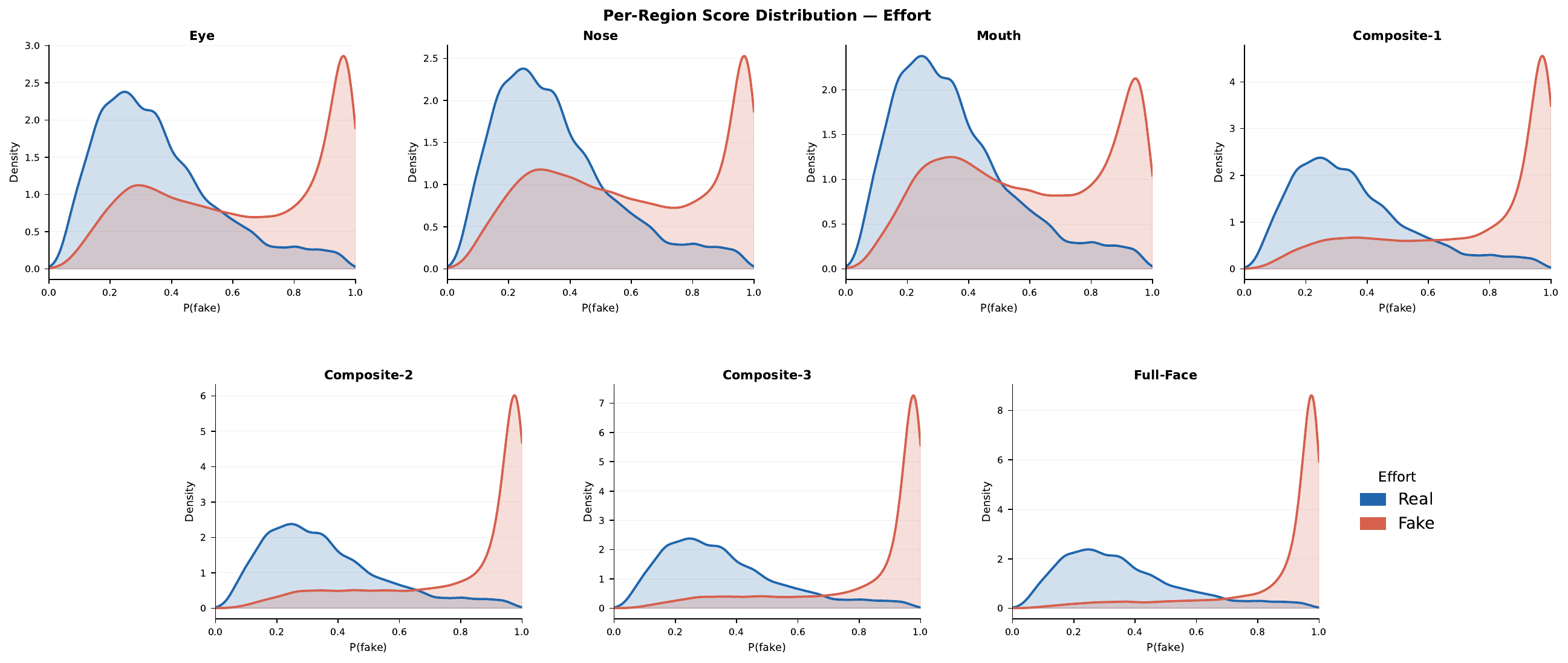}
    \caption{\textbf{Effort}: Per-region real/fake score distributions.}
    \label{fig:supp_score_effort}
\end{sidewaysfigure}

\begin{sidewaysfigure}
    \centering
\includegraphics[width=\linewidth,height=0.92\textwidth,keepaspectratio]{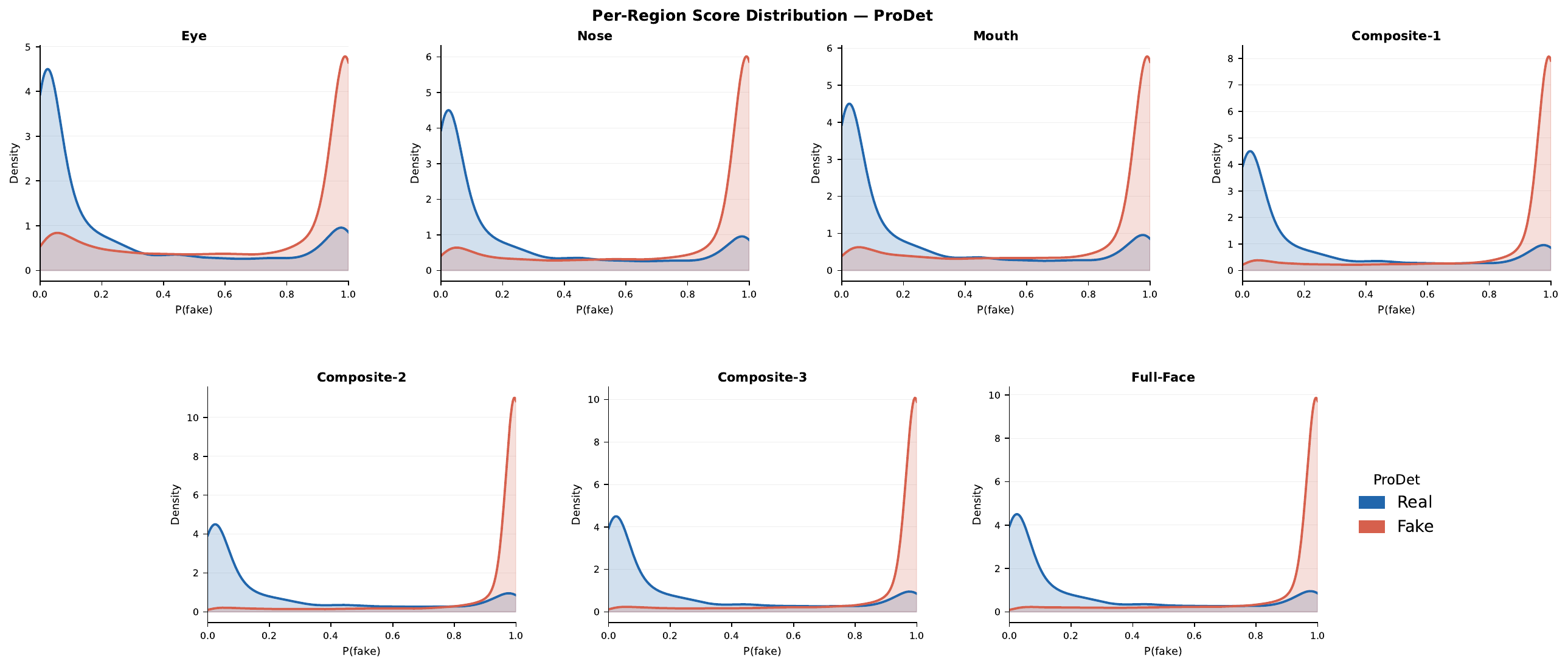}
    \caption{\textbf{ProDet}: Per-region real/fake score distributions.}
    \label{fig:supp_score_prodet}
\end{sidewaysfigure}

\begin{sidewaysfigure}
    \centering
\includegraphics[width=\linewidth,height=0.92\textwidth,keepaspectratio]{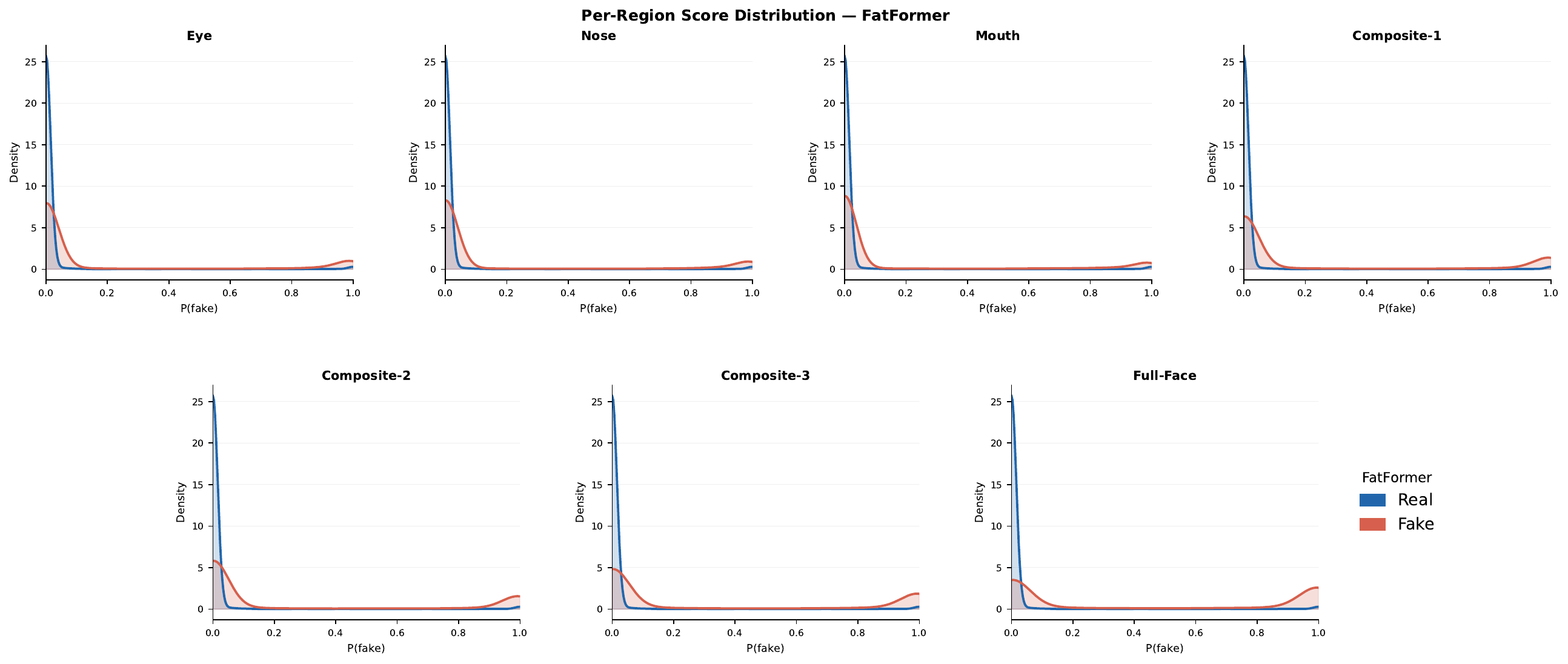}
\caption{\textbf{FatFormer}: Per-region real/fake score distributions.}
    \label{fig:supp_score_fatformer}
\end{sidewaysfigure}

\begin{table}[htbp]
\centering
\caption{\textbf{Protocol~\#3 (\textbf{BlendFace}): Region-level performance (ACC\,/\,AUC).} \textbf{Bold} = best, \underline{underline} = second-best (per metric, per column). C-1: eye$+$nose, C-2: eye$+$nose$+$mouth, C-3: eye$+$nose$+$mouth$+$eyebrow. }
\label{tab:supp_blendface}
\setlength{\tabcolsep}{3pt}
\renewcommand{\arraystretch}{1.3}
\resizebox{\linewidth}{!}{%
\begin{tabular}{l c cc cc cc cc cc cc cc}
\toprule
\textbf{Detector} & \textbf{Venue} & \multicolumn{2}{c}{\small\textbf{Eye}} & \multicolumn{2}{c}{\small\textbf{Nose}} & \multicolumn{2}{c}{\small\textbf{Mouth}} & \multicolumn{2}{c}{\small\textbf{C-1}} & \multicolumn{2}{c}{\small\textbf{C-2}} & \multicolumn{2}{c}{\small\textbf{C-3}} & \multicolumn{2}{c}{\small\textbf{Full}} \\
\cmidrule(lr){3-4} \cmidrule(lr){5-6} \cmidrule(lr){7-8} \cmidrule(lr){9-10} \cmidrule(lr){11-12} \cmidrule(lr){13-14} \cmidrule(lr){15-16}
 & & \small ACC & \small AUC & \small ACC & \small AUC & \small ACC & \small AUC & \small ACC & \small AUC & \small ACC & \small AUC & \small ACC & \small AUC & \small ACC & \small AUC \\
\midrule
Xception~\cite{ff++}  & \small\scriptsize ICCV'19   & 0.62 & 0.68 & 0.64 & 0.70 & \underline{0.66} & 0.73 & 0.66 & 0.72 & 0.70 & 0.78 & 0.71 & 0.79 & 0.72 & 0.83 \\
F3Net~\cite{F3Net}     & \small\scriptsize ECCV'20   & 0.56 & 0.66 & 0.55 & 0.64 & 0.59 & 0.69 & 0.56 & 0.67 & 0.60 & 0.72 & 0.62 & 0.75 & 0.67 & 0.81 \\
RECCE~\cite{RECCE}     & \small\scriptsize CVPR'22   & 0.59 & 0.68 & 0.59 & 0.68 & 0.58 & 0.69 & 0.61 & 0.71 & 0.61 & 0.73 & 0.64 & 0.76 & 0.67 & 0.78 \\
SBI~\cite{SBI}       & \small\scriptsize CVPR'22   & 0.62 & 0.73 & 0.62 & 0.73 & 0.60 & 0.72 & 0.70 & 0.82 & 0.76 & 0.87 & 0.77 & 0.89 & 0.85 & 0.94 \\
LGrad~\cite{LGrad}     & \small\scriptsize CVPR'23   & 0.50 & 0.54 & 0.50 & 0.53 & 0.50 & 0.53 & 0.50 & 0.54 & 0.50 & 0.54 & 0.50 & 0.54 & 0.50 & 0.53 \\
UCF~\cite{yan2023ucf}       & \small\scriptsize ICCV'23   & 0.59 & 0.68 & 0.59 & 0.68 & 0.61 & 0.72 & 0.61 & 0.72 & 0.65 & 0.77 & 0.69 & 0.81 & 0.73 & 0.84 \\
FatFormer~\cite{fatformer} & \small\scriptsize CVPR'24   & 0.51 & 0.57 & 0.51 & 0.59 & 0.51 & 0.57 & 0.52 & 0.61 & 0.53 & 0.62 & 0.54 & 0.64 & 0.57 & 0.70 \\
FreqNet~\cite{FreqNet}   & \small\scriptsize AAAI'24   & 0.39 & 0.45 & 0.39 & 0.45 & 0.38 & 0.45 & 0.39 & 0.45 & 0.39 & 0.45 & 0.38 & 0.45 & 0.37 & 0.44 \\
ProDet~\cite{prodet}    & \small\scriptsize NeurIPS'24 & \textbf{0.75} & \textbf{0.82} & \textbf{0.78} & \textbf{0.86} & \textbf{0.77} & \textbf{0.84} & \textbf{0.82} & \textbf{0.91} & \textbf{0.86} & \textbf{0.94} & \textbf{0.85} & \textbf{0.94} & \underline{0.87} & \underline{0.96} \\
DFD-FCG~\cite{dfd_fcg}   & \small\scriptsize CVPR'25   & 0.58 & 0.59 & 0.57 & 0.58 & 0.59 & 0.63 & 0.61 & 0.65 & 0.66 & 0.74 & 0.67 & 0.77 & 0.69 & 0.81 \\
Effort~\cite{effort}    & \small\scriptsize ICML'25   & 0.58 & 0.63 & 0.57 & 0.60 & 0.57 & 0.61 & 0.63 & 0.67 & 0.65 & 0.71 & 0.68 & 0.74 & 0.71 & 0.77 \\
ForAda~\cite{ForAda}    & \small\scriptsize CVPR'25   & \underline{0.65} & 0.72 & \underline{0.66} & 0.72 & 0.63 & 0.71 & 0.70 & 0.76 & 0.71 & 0.77 & 0.73 & 0.80 & 0.73 & 0.81 \\
\midrule
\rowcolor{rldfblue}
\textbf{Ours (12-Blk)} & — & 0.61 & \underline{0.77} & 0.62 & \underline{0.78} & 0.59 & \underline{0.77} & \underline{0.72} & \underline{0.85} & \underline{0.77} & \underline{0.89} & \underline{0.82} & \underline{0.93} & \textbf{0.90} & \textbf{0.96} \\
\bottomrule
\end{tabular}}
\end{table}
\begin{sidewaysfigure}
    \centering
\includegraphics[width=\linewidth,height=0.92\textwidth,keepaspectratio]{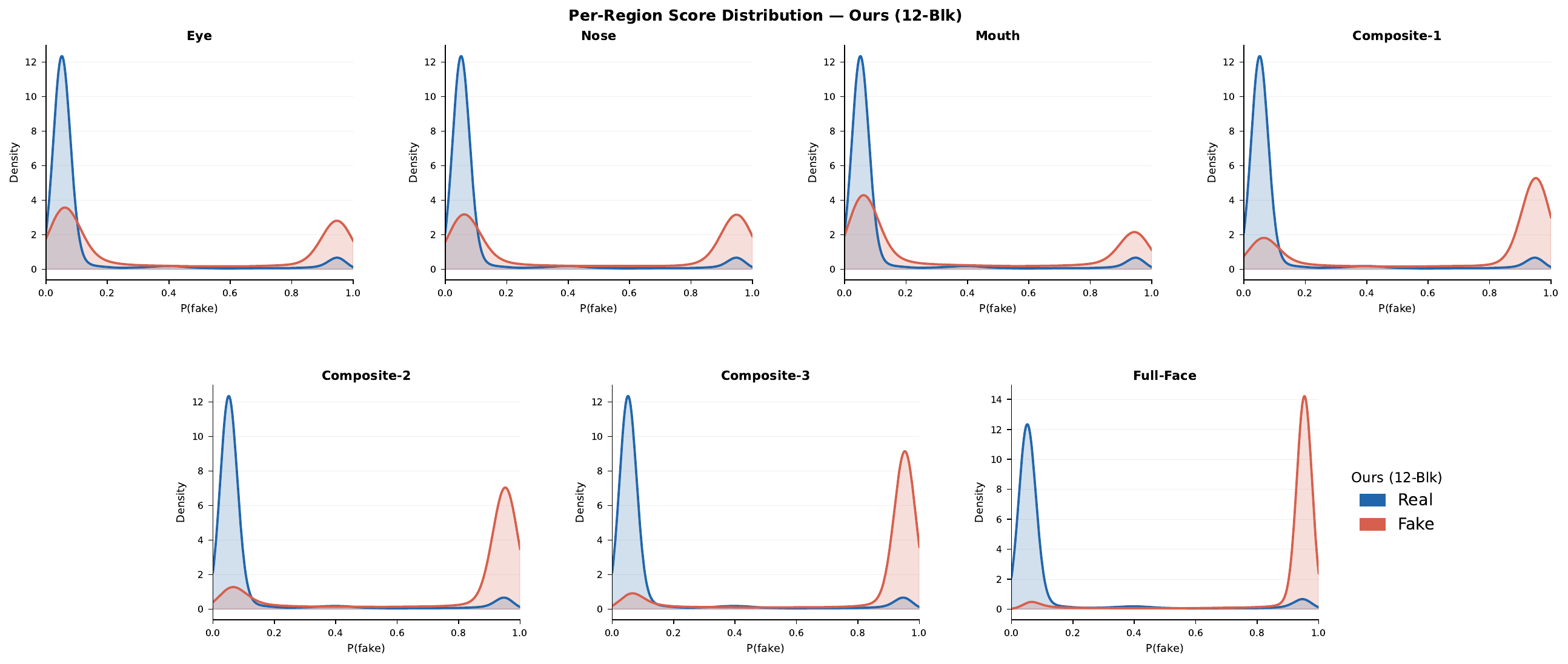}
    \caption{\textbf{Ours (12-Blk)}: Per-region real/fake score distributions.}
    \label{fig:supp_score_ours12}
\end{sidewaysfigure}
Second, detector behaviour is strongly method-dependent.
\textbf{E4S} (Table~\ref{tab:supp_e4s}) is the most detectable across all
detectors AUC values exceed 0.90 for most methods even at the Eye region
owing to its visible style-transfer boundaries that are detectable even in
small regions.
\textbf{BlendFace} (Table~\ref{tab:supp_blendface}) is the hardest: its
seam-minimising latent blending produces the lowest AUC values consistently,
with several detectors falling below 0.70 at the Eye region.
\textbf{SimSwap} (Table~\ref{tab:supp_simswap}) similarly challenges all
detectors, particularly at single-region granularities, due to its
weak feature matching that preserves target attributes and minimises
swap-induced discontinuities.
Third, frequency-based detectors (LGrad, FreqNet) fail systematically across
all methods and all regions (AUC near 0.50), confirming that the C23
compression artefacts exploited during training do not transfer to
region-level manipulations.
Conversely, our proposed 8-Block and 12-Block baselines maintain competitive
AUC across most method-region pairs, with performance particularly strong
on FSGAN (Table~\ref{tab:supp_fsgan}), FaceDancer (Table~\ref{tab:supp_facedancer}),
and UniFace (Table~\ref{tab:supp_uniface}), where localised blending seams
provide strong patch-level signals for MaskNet.
\begin{table}[htbp]
\centering
\caption{\textbf{Protocol~\#3 (\textbf{E4S}): Region-level performance (ACC\,/\,AUC).} \textbf{Bold} = best, \underline{underline} = second-best (per metric, per column). C-1: eye$+$nose, C-2: eye$+$nose$+$mouth, C-3: eye$+$nose$+$mouth$+$eyebrow. }
\label{tab:supp_e4s}
\setlength{\tabcolsep}{3pt}
\renewcommand{\arraystretch}{1.3}
\resizebox{\linewidth}{!}{%
\begin{tabular}{l c cc cc cc cc cc cc cc}
\toprule
\textbf{Detector} & \textbf{Venue} & \multicolumn{2}{c}{\small\textbf{Eye}} & \multicolumn{2}{c}{\small\textbf{Nose}} & \multicolumn{2}{c}{\small\textbf{Mouth}} & \multicolumn{2}{c}{\small\textbf{C-1}} & \multicolumn{2}{c}{\small\textbf{C-2}} & \multicolumn{2}{c}{\small\textbf{C-3}} & \multicolumn{2}{c}{\small\textbf{Full}} \\
\cmidrule(lr){3-4} \cmidrule(lr){5-6} \cmidrule(lr){7-8} \cmidrule(lr){9-10} \cmidrule(lr){11-12} \cmidrule(lr){13-14} \cmidrule(lr){15-16}
 & & \small ACC & \small AUC & \small ACC & \small AUC & \small ACC & \small AUC & \small ACC & \small AUC & \small ACC & \small AUC & \small ACC & \small AUC & \small ACC & \small AUC \\
\midrule
Xception~\cite{ff++}  & \small\scriptsize ICCV'19    & 0.70 & 0.75 & 0.70 & 0.77 & 0.74 & 0.82 & 0.71 & 0.77 & 0.67 & 0.74 & 0.70 & 0.78 & 0.71 & 0.83 \\
F3Net~\cite{F3Net}     & \small\scriptsize ECCV'20    & 0.68 & 0.78 & 0.70 & 0.79 & 0.72 & 0.82 & 0.67 & 0.76 & 0.68 & 0.76 & 0.69 & 0.79 & 0.55 & 0.66 \\
RECCE~\cite{RECCE}     & \small\scriptsize CVPR'22    & 0.74 & 0.81 & 0.80 & 0.88 & 0.66 & 0.77 & 0.81 & 0.89 & 0.73 & 0.83 & 0.76 & 0.85 & 0.64 & 0.76 \\
SBI~\cite{SBI}       & \small\scriptsize CVPR'22    & 0.88 & 0.92 & 0.92 & 0.98 & 0.89 & 0.96 & 0.95 & 1.00 & 0.95 & \underline{1.00} & 0.95 & 1.00 & 0.95 & 1.00 \\
LGrad~\cite{LGrad}     & \small\scriptsize CVPR'23    & 0.50 & 0.56 & 0.50 & 0.56 & 0.50 & 0.57 & 0.50 & 0.57 & 0.50 & 0.58 & 0.50 & 0.57 & 0.50 & 0.58 \\
UCF~\cite{yan2023ucf}       & \small\scriptsize ICCV'23    & 0.74 & 0.83 & 0.84 & 0.91 & 0.83 & 0.92 & 0.82 & 0.91 & 0.81 & 0.91 & 0.83 & 0.92 & 0.65 & 0.79 \\
FatFormer~\cite{fatformer} & \small\scriptsize CVPR'24    & \textbf{0.89} & \underline{0.93} & \underline{0.94} & 0.99 & \underline{0.91} & 0.99 & \textbf{0.98} & 0.99 & \textbf{0.99} & 0.99 & \textbf{0.99} & 0.99 & \textbf{0.99} & 1.00 \\
FreqNet~\cite{FreqNet}   & \small\scriptsize AAAI'24    & 0.37 & 0.44 & 0.39 & 0.47 & 0.38 & 0.45 & 0.38 & 0.46 & 0.38 & 0.46 & 0.39 & 0.47 & 0.35 & 0.42 \\
ProDet~\cite{prodet}    & \small\scriptsize NeurIPS'24 & 0.83 & 0.93 & 0.88 & 0.98 & 0.86 & 0.95 & 0.88 & 0.99 & 0.88 & 0.99 & 0.88 & 0.99 & 0.88 & 0.99 \\
DFD-FCG~\cite{dfd_fcg}   & \small\scriptsize CVPR'25    & 0.70 & 0.91 & 0.74 & 0.99 & 0.74 & \textbf{0.99} & 0.74 & 1.00 & 0.74 & \textbf{1.00} & 0.74 & \textbf{1.00} & 0.74 & \textbf{1.00} \\
Effort~\cite{effort}    & \small\scriptsize ICML'25    & 0.84 & 0.93 & 0.90 & \underline{1.00} & 0.90 & 0.99 & 0.90 & \underline{1.00} & 0.90 & 1.00 & 0.90 & \underline{1.00} & 0.90 & \underline{1.00} \\
ForAda~\cite{ForAda}    & \small\scriptsize CVPR'25    & 0.88 & \textbf{0.94} & 0.94 & \textbf{1.00} & \textbf{0.93} & \underline{0.99} & 0.94 & \textbf{1.00} & 0.94 & 1.00 & 0.94 & 1.00 & 0.93 & 0.99 \\
\midrule
\rowcolor{rldfblue}
\textbf{Ours (12-Blk)} & — & \underline{0.88} & 0.93 & \textbf{0.96} & 0.99 & 0.84 & 0.93 & \underline{0.96} & 0.99 & \underline{0.96} & 0.99 & \underline{0.96} & 0.99 & \underline{0.96} & 0.99 \\
\bottomrule
\end{tabular}}
\end{table}
\noindent\textbf{Score Distributions:}
Figure~\ref{fig:supp_score_eye_full} directly compares real/fake score distributions for Eye-region fakes vs Full Face fakes.
The Eye region shows near-complete overlap between real and fake distributions, explaining the low detection accuracy, full face shows clear bimodal separation.
Detection difficulty scales monotonically with the manipulated area, validating RLDF's progressive region granularity as a diagnostic tool.
Figure~\ref{fig:supp_score_eye_full} confirms this across all detectors: mean $P(\text{fake})$ on fake samples rises monotonically from Eye to Full Face for every method evaluated.
Figures~\ref{fig:supp_score_ours12}--\ref{fig:supp_score_fatformer} show per-region real/fake KDE score distributions for four detectors.

\begin{table}[htbp]
\centering
\caption{\textbf{Protocol~\#3 (\textbf{FaceDancer}): Region-level performance (ACC\,/\,AUC).} \textbf{Bold} = best; \underline{underline} = second-best (per metric, per column). C-1: eye$+$nose, C-2: eye$+$nose$+$mouth, C-3: eye$+$nose$+$mouth$+$eyebrow. }
\label{tab:supp_facedancer}
\setlength{\tabcolsep}{3pt}
\renewcommand{\arraystretch}{1.3}
\resizebox{\linewidth}{!}{%
\begin{tabular}{l c cc cc cc cc cc cc cc}
\toprule
\textbf{Detector} & \textbf{Venue} & \multicolumn{2}{c}{\small\textbf{Eye}} & \multicolumn{2}{c}{\small\textbf{Nose}} & \multicolumn{2}{c}{\small\textbf{Mouth}} & \multicolumn{2}{c}{\small\textbf{C-1}} & \multicolumn{2}{c}{\small\textbf{C-2}} & \multicolumn{2}{c}{\small\textbf{C-3}} & \multicolumn{2}{c}{\small\textbf{Full}} \\
\cmidrule(lr){3-4} \cmidrule(lr){5-6} \cmidrule(lr){7-8} \cmidrule(lr){9-10} \cmidrule(lr){11-12} \cmidrule(lr){13-14} \cmidrule(lr){15-16}
 & & \small ACC & \small AUC & \small ACC & \small AUC & \small ACC & \small AUC & \small ACC & \small AUC & \small ACC & \small AUC & \small ACC & \small AUC & \small ACC & \small AUC \\
\midrule
Xception~\cite{ff++}  & \small\scriptsize ICCV'19    & 0.63 & 0.69 & 0.67 & 0.74 & 0.67 & 0.75 & 0.68 & 0.76 & 0.74 & 0.81 & 0.73 & 0.80 & 0.73 & 0.85 \\
F3Net~\cite{F3Net}     & \small\scriptsize ECCV'20    & 0.58 & 0.67 & 0.61 & 0.72 & 0.58 & 0.68 & 0.62 & 0.73 & 0.65 & 0.76 & 0.65 & 0.76 & 0.74 & 0.84 \\
RECCE~\cite{RECCE}     & \small\scriptsize CVPR'22    & 0.61 & 0.68 & 0.65 & 0.73 & 0.64 & 0.75 & 0.66 & 0.74 & 0.71 & 0.81 & 0.70 & 0.80 & 0.79 & 0.87 \\
SBI~\cite{SBI}       & \small\scriptsize CVPR'22    & 0.58 & 0.68 & 0.56 & 0.67 & 0.56 & 0.65 & 0.61 & 0.74 & 0.63 & 0.77 & 0.64 & 0.78 & 0.77 & 0.89 \\
LGrad~\cite{LGrad}     & \small\scriptsize CVPR'23    & 0.50 & 0.53 & 0.50 & 0.54 & 0.50 & 0.54 & 0.50 & 0.53 & 0.50 & 0.53 & 0.50 & 0.52 & 0.50 & 0.52 \\
UCF~\cite{yan2023ucf}       & \small\scriptsize ICCV'23    & 0.62 & 0.72 & 0.66 & 0.77 & 0.64 & 0.75 & 0.68 & 0.79 & 0.71 & 0.83 & 0.70 & 0.82 & 0.79 & 0.89 \\
FatFormer & \small\scriptsize CVPR'24    & 0.52 & 0.67 & 0.51 & 0.60 & 0.51 & 0.61 & 0.54 & 0.71 & 0.56 & 0.74 & 0.65 & 0.84 & 0.70 & 0.87 \\
FreqNet~\cite{FreqNet}   & \small\scriptsize AAAI'24    & 0.38 & 0.46 & 0.39 & 0.47 & 0.39 & 0.46 & 0.39 & 0.47 & 0.40 & 0.47 & 0.40 & 0.48 & 0.41 & 0.48 \\
ProDet~\cite{prodet}    & \small\scriptsize NeurIPS'24 & 0.70 & 0.77 & 0.70 & 0.77 & \underline{0.69} & \underline{0.76} & 0.76 & 0.83 & 0.78 & 0.86 & 0.78 & 0.85 & 0.81 & 0.90 \\
DFD-FCG~\cite{dfd_fcg}   & \small\scriptsize CVPR'25    & 0.65 & 0.73 & 0.62 & 0.67 & 0.61 & 0.65 & 0.69 & 0.82 & 0.71 & 0.87 & 0.71 & 0.90 & 0.72 & 0.93 \\
Effort~\cite{effort}    & \small\scriptsize ICML'25    & 0.72 & 0.77 & 0.66 & 0.71 & 0.66 & 0.71 & 0.79 & 0.87 & 0.83 & \underline{0.91} & 0.86 & \underline{0.95} & 0.87 & \underline{0.96} \\
ForAda~\cite{ForAda}    & \small\scriptsize CVPR'25    & \underline{0.73} & \underline{0.78} & \textbf{0.74} & \underline{0.80} & \textbf{0.69} & 0.76 & \underline{0.82} & \underline{0.89} & \underline{0.84} & 0.90 & \underline{0.87} & 0.93 & \underline{0.88} & 0.95 \\
\midrule
\rowcolor{rldfblue}
\textbf{Ours (12-Blk)} & — & \textbf{0.75} & \textbf{0.87} & \underline{0.70} & \textbf{0.84} & 0.61 & \textbf{0.79} & \textbf{0.86} & \textbf{0.94} & \textbf{0.89} & \textbf{0.96} & \textbf{0.92} & \textbf{0.97} & \textbf{0.94} & \textbf{0.98} \\
\bottomrule
\end{tabular}}
\end{table}
% ── Supplementary: Per-Method AUC/ACC
\begin{table*}[tb]
\centering
\caption{\textbf{Protocol~\#3 (\textbf{FSGAN}): Region-level performance (ACC\,/\,AUC).} \textbf{Bold} = best, \underline{underline} = second-best (per metric, per column). C-1: eye$+$nose, C-2: eye$+$nose$+$mouth, C-3: eye$+$nose$+$mouth$+$eyebrow. }
\label{tab:supp_fsgan}
\setlength{\tabcolsep}{3pt}
\renewcommand{\arraystretch}{1.3}
\resizebox{\linewidth}{!}{%
\begin{tabular}{l c cc cc cc cc cc cc cc}
\toprule
\textbf{Detector} & \textbf{Venue} & \multicolumn{2}{c}{\small\textbf{Eye}} & \multicolumn{2}{c}{\small\textbf{Nose}} & \multicolumn{2}{c}{\small\textbf{Mouth}} & \multicolumn{2}{c}{\small\textbf{C-1}} & \multicolumn{2}{c}{\small\textbf{C-2}} & \multicolumn{2}{c}{\small\textbf{C-3}} & \multicolumn{2}{c}{\small\textbf{Full}} \\
\cmidrule(lr){3-4} \cmidrule(lr){5-6} \cmidrule(lr){7-8} \cmidrule(lr){9-10} \cmidrule(lr){11-12} \cmidrule(lr){13-14} \cmidrule(lr){15-16}
 & & \small ACC & \small AUC & \small ACC & \small AUC & \small ACC & \small AUC & \small ACC & \small AUC & \small ACC & \small AUC & \small ACC & \small AUC & \small ACC & \small AUC \\
\midrule
Xception~\cite{ff++} & \small\scriptsize ICCV'19 & 0.67 & 0.75 & 0.76 & 0.84 & 0.82 & 0.91 & 0.79 & 0.88 & 0.86 & 0.96 & 0.85 & 0.94 & 0.85 & 0.95 \\
F3Net~\cite{F3Net} & \small\scriptsize ECCV'20 & 0.64 & 0.80 & 0.76 & 0.89 & 0.80 & 0.91 & 0.81 & 0.92 & 0.90 & 0.97 & 0.87 & 0.95 & 0.92 & 0.98 \\
RECCE~\cite{RECCE} & \small\scriptsize CVPR'22 & 0.68 & 0.79 & 0.78 & 0.88 & 0.86 & 0.93 & 0.82 & 0.91 & 0.90 & 0.97 & 0.88 & 0.95 & 0.91 & 0.98 \\
SBI~\cite{SBI} & \small\scriptsize CVPR'22 & 0.64 & 0.77 & 0.69 & 0.83 & 0.66 & 0.79 & 0.80 & 0.91 & 0.86 & 0.95 & 0.86 & 0.95 & 0.87 & 0.96 \\
LGrad~\cite{LGrad} & \small\scriptsize CVPR'23 & 0.49 & 0.48 & 0.49 & 0.48 & 0.49 & 0.48 & 0.49 & 0.48 & 0.49 & 0.49 & 0.49 & 0.49 & 0.49 & 0.48 \\
UCF~\cite{yan2023ucf} & \small\scriptsize ICCV'23 & 0.77 & \underline{0.87} & \textbf{0.85} & \textbf{0.93} & \textbf{0.87} & \textbf{0.95} & 0.89 & 0.96 & 0.92 & 0.98 & 0.91 & 0.98 & 0.93 & \underline{0.99} \\
FatFormer~\cite{fatformer} & \small\scriptsize CVPR'24 & 0.54 & 0.65 & 0.54 & 0.64 & 0.51 & 0.56 & 0.59 & 0.75 & 0.60 & 0.75 & 0.63 & 0.79 & 0.59 & 0.75 \\
FreqNet~\cite{FreqNet} & \small\scriptsize AAAI'24 & 0.32 & 0.35 & 0.32 & 0.35 & 0.33 & 0.35 & 0.32 & 0.35 & 0.32 & 0.36 & 0.32 & 0.36 & 0.31 & 0.33 \\
ProDet~\cite{prodet} & \small\scriptsize NeurIPS'24 & \textbf{0.83} & \textbf{0.92} & 0.84 & \underline{0.92} & \underline{0.86} & \underline{0.95} & 0.86 & 0.95 & 0.88 & 0.98 & 0.88 & 0.98 & 0.88 & 0.98 \\
DFD-FCG~\cite{dfd_fcg} & \small\scriptsize CVPR'25 & 0.65 & 0.77 & 0.67 & 0.80 & 0.67 & 0.79 & 0.72 & 0.92 & 0.73 & 0.96 & 0.73 & 0.97 & 0.73 & 0.97 \\
Effort~\cite{effort} & \small\scriptsize ICML'25 & \underline{0.77} & 0.85 & 0.81 & 0.89 & 0.75 & 0.83 & 0.87 & \underline{0.97} & 0.89 & \underline{0.99} & 0.89 & \underline{0.99} & 0.89 & \textbf{0.99} \\
ForAda~\cite{ForAda} & \small\scriptsize CVPR'25 & 0.76 & 0.84 & \underline{0.85} & 0.92 & 0.83 & 0.91 & \textbf{0.91} & \textbf{0.97} & \underline{0.93} & \textbf{0.99} & \underline{0.93} & \textbf{0.99} & \underline{0.93} & 0.99 \\
\midrule
\rowcolor{rldfblue}
\textbf{Ours (12-Blk)} & — & 0.69 & 0.81 & 0.79 & 0.88 & 0.64 & 0.80 & \underline{0.90} & 0.96 & \textbf{0.93} & 0.97 & \textbf{0.94} & 0.98 & \textbf{0.96} & 0.99 \\
\bottomrule
\end{tabular}}
\end{table*}

\begin{table*}[tb]
\centering
\caption{\textbf{Protocol~\#3 (\textbf{GHOST}): Region-level performance (ACC\,/\,AUC).} \textbf{Bold} = best, \underline{underline} = second-best (per metric, per column). C-1: eye$+$nose, C-2: eye$+$nose$+$mouth, C-3: eye$+$nose$+$mouth$+$eyebrow. }
\label{tab:supp_ghost}
\setlength{\tabcolsep}{3pt}
\renewcommand{\arraystretch}{1.3}
\resizebox{\linewidth}{!}{%
\begin{tabular}{l c cc cc cc cc cc cc cc}
\toprule
\textbf{Detector} & \textbf{Venue} & \multicolumn{2}{c}{\small\textbf{Eye}} & \multicolumn{2}{c}{\small\textbf{Nose}} & \multicolumn{2}{c}{\small\textbf{Mouth}} & \multicolumn{2}{c}{\small\textbf{C-1}} & \multicolumn{2}{c}{\small\textbf{C-2}} & \multicolumn{2}{c}{\small\textbf{C-3}} & \multicolumn{2}{c}{\small\textbf{Full}} \\
\cmidrule(lr){3-4} \cmidrule(lr){5-6} \cmidrule(lr){7-8} \cmidrule(lr){9-10} \cmidrule(lr){11-12} \cmidrule(lr){13-14} \cmidrule(lr){15-16}
 & & \small ACC & \small AUC & \small ACC & \small AUC & \small ACC & \small AUC & \small ACC & \small AUC & \small ACC & \small AUC & \small ACC & \small AUC & \small ACC & \small AUC \\
\midrule
Xception~\cite{ff++} & \small\scriptsize ICCV'19 & 0.61 & 0.68 & 0.65 & 0.72 & 0.74 & 0.82 & 0.64 & 0.73 & 0.74 & 0.82 & 0.76 & 0.84 & 0.77 & 0.88 \\
F3Net~\cite{F3Net} & \small\scriptsize ECCV'20 & 0.61 & 0.76 & 0.61 & 0.76 & 0.67 & 0.82 & 0.61 & 0.77 & 0.67 & 0.81 & 0.73 & 0.87 & 0.61 & 0.79 \\
RECCE~\cite{RECCE} & \small\scriptsize CVPR'22 & 0.64 & 0.73 & 0.67 & 0.77 & 0.72 & 0.83 & 0.66 & 0.78 & 0.73 & 0.83 & 0.76 & 0.86 & 0.57 & 0.65 \\
SBI~\cite{SBI} & \small\scriptsize CVPR'22 & 0.61 & 0.70 & 0.62 & 0.70 & 0.60 & 0.67 & 0.72 & 0.83 & 0.78 & 0.88 & 0.77 & 0.88 & 0.77 & 0.90 \\
LGrad~\cite{LGrad} & \small\scriptsize CVPR'23 & 0.49 & 0.57 & 0.50 & 0.57 & 0.50 & 0.57 & 0.50 & 0.57 & 0.50 & 0.57 & 0.50 & 0.57 & 0.50 & 0.57 \\
UCF~\cite{yan2023ucf} & \small\scriptsize ICCV'23 & 0.71 & \textbf{0.84} & \underline{0.74} & \textbf{0.86} & \underline{0.75} & \textbf{0.87} & 0.75 & 0.87 & 0.78 & 0.89 & 0.83 & 0.93 & 0.69 & 0.83 \\
FatFormer~\cite{fatformer} & \small\scriptsize CVPR'24 & 0.53 & 0.70 & 0.51 & 0.68 & 0.51 & 0.68 & 0.56 & 0.75 & 0.58 & 0.78 & 0.62 & 0.81 & 0.69 & 0.87 \\
FreqNet~\cite{FreqNet} & \small\scriptsize AAAI'24 & 0.38 & 0.45 & 0.38 & 0.45 & 0.38 & 0.46 & 0.38 & 0.46 & 0.39 & 0.46 & 0.39 & 0.46 & 0.37 & 0.44 \\
ProDet~\cite{prodet} & \small\scriptsize NeurIPS'24 & \textbf{0.73} & \underline{0.82} & \textbf{0.77} & \underline{0.85} & \textbf{0.76} & \underline{0.84} & \textbf{0.81} & \underline{0.89} & \underline{0.84} & 0.92 & 0.84 & 0.92 & 0.77 & 0.84 \\
DFD-FCG~\cite{dfd_fcg} & \small\scriptsize CVPR'25 & 0.59 & 0.64 & 0.56 & 0.59 & 0.63 & 0.70 & 0.62 & 0.71 & 0.68 & 0.81 & 0.70 & 0.84 & 0.71 & 0.87 \\
Effort~\cite{effort} & \small\scriptsize ICML'25 & \underline{0.73} & 0.80 & 0.68 & 0.73 & 0.69 & 0.76 & \underline{0.80} & 0.88 & 0.84 & \underline{0.92} & \underline{0.85} & \underline{0.94} & \underline{0.88} & \underline{0.96} \\
ForAda~\cite{ForAda} & \small\scriptsize CVPR'25 & 0.66 & 0.72 & 0.67 & 0.74 & 0.68 & 0.75 & 0.75 & 0.83 & 0.81 & 0.89 & 0.84 & 0.91 & 0.82 & 0.89 \\
\midrule
\rowcolor{rldfblue}
\textbf{Ours (12-Blk)} & — & 0.65 & 0.79 & 0.65 & 0.77 & 0.62 & 0.77 & 0.79 & \textbf{0.89} & \textbf{0.86} & \textbf{0.94} & \textbf{0.91} & \textbf{0.96} & \textbf{0.93} & \textbf{0.97} \\
\bottomrule
\end{tabular}}
\end{table*}

\begin{table*}[tb]
\centering
\caption{\textbf{Protocol~\#3 (\textbf{InSwapper}): Region-level performance (ACC\,/\,AUC).} \textbf{Bold} = best, \underline{underline} = second-best (per metric, per column). C-1: eye$+$nose, C-2: eye$+$nose$+$mouth, C-3: eye$+$nose$+$mouth$+$eyebrow. }
\label{tab:supp_inswapper}
\setlength{\tabcolsep}{3pt}
\renewcommand{\arraystretch}{1.3}
\resizebox{\linewidth}{!}{%
\begin{tabular}{l c cc cc cc cc cc cc cc}
\toprule
\textbf{Detector} & \textbf{Venue} & \multicolumn{2}{c}{\small\textbf{Eye}} & \multicolumn{2}{c}{\small\textbf{Nose}} & \multicolumn{2}{c}{\small\textbf{Mouth}} & \multicolumn{2}{c}{\small\textbf{C-1}} & \multicolumn{2}{c}{\small\textbf{C-2}} & \multicolumn{2}{c}{\small\textbf{C-3}} & \multicolumn{2}{c}{\small\textbf{Full}} \\
\cmidrule(lr){3-4} \cmidrule(lr){5-6} \cmidrule(lr){7-8} \cmidrule(lr){9-10} \cmidrule(lr){11-12} \cmidrule(lr){13-14} \cmidrule(lr){15-16}
 & & \small ACC & \small AUC & \small ACC & \small AUC & \small ACC & \small AUC & \small ACC & \small AUC & \small ACC & \small AUC & \small ACC & \small AUC & \small ACC & \small AUC \\
\midrule
Xception~\cite{ff++} & \small\scriptsize ICCV'19 & 0.64 & 0.73 & 0.79 & 0.87 & 0.81 & 0.88 & 0.81 & 0.89 & 0.86 & 0.94 & 0.83 & 0.91 & 0.84 & 0.93 \\
F3Net~\cite{F3Net} & \small\scriptsize ECCV'20 & 0.70 & \underline{0.86} & 0.84 & \underline{0.94} & 0.80 & 0.91 & \underline{0.89} & 0.96 & 0.92 & 0.98 & 0.88 & 0.96 & 0.86 & 0.95 \\
RECCE~\cite{RECCE} & \small\scriptsize CVPR'22 & 0.67 & 0.79 & \underline{0.85} & 0.93 & \underline{0.86} & \underline{0.95} & 0.84 & 0.93 & 0.90 & 0.98 & 0.88 & 0.96 & 0.87 & 0.95 \\
SBI~\cite{SBI} & \small\scriptsize CVPR'22 & 0.68 & 0.79 & 0.73 & 0.85 & 0.61 & 0.73 & 0.88 & \underline{0.96} & \textbf{0.93} & \underline{0.98} & \textbf{0.92} & \textbf{0.98} & \underline{0.91} & 0.97 \\
LGrad~\cite{LGrad} & \small\scriptsize CVPR'23 & 0.50 & 0.59 & 0.50 & 0.60 & 0.50 & 0.60 & 0.50 & 0.59 & 0.49 & 0.59 & 0.50 & 0.58 & 0.50 & 0.59 \\
UCF~\cite{yan2023ucf} & \small\scriptsize ICCV'23 & \textbf{0.80} & \textbf{0.91} & \textbf{0.89} & \textbf{0.96} & \textbf{0.88} & \textbf{0.95} & \textbf{0.91} & \textbf{0.97} & \underline{0.92} & \textbf{0.98} & \underline{0.91} & \underline{0.98} & 0.91 & 0.97 \\
FatFormer~\cite{fatformer} & \small\scriptsize CVPR'24 & 0.52 & 0.71 & 0.50 & 0.69 & 0.51 & 0.69 & 0.53 & 0.75 & 0.54 & 0.75 & 0.56 & 0.77 & 0.66 & 0.87 \\
FreqNet~\cite{FreqNet} & \small\scriptsize AAAI'24 & 0.38 & 0.46 & 0.38 & 0.46 & 0.38 & 0.46 & 0.39 & 0.46 & 0.39 & 0.47 & 0.39 & 0.47 & 0.40 & 0.47 \\
ProDet~\cite{prodet} & \small\scriptsize NeurIPS'24 & 0.69 & 0.78 & 0.79 & 0.86 & 0.78 & 0.86 & 0.83 & 0.90 & 0.85 & 0.92 & 0.83 & 0.89 & 0.80 & 0.87 \\
DFD-FCG~\cite{dfd_fcg} & \small\scriptsize CVPR'25 & 0.63 & 0.70 & 0.61 & 0.66 & 0.68 & 0.79 & 0.67 & 0.78 & 0.72 & 0.90 & 0.73 & 0.91 & 0.73 & 0.94 \\
Effort~\cite{effort} & \small\scriptsize ICML'25 & \underline{0.73} & 0.80 & 0.70 & 0.76 & 0.74 & 0.81 & 0.81 & 0.89 & 0.87 & 0.95 & 0.88 & 0.96 & 0.89 & \underline{0.98} \\
ForAda~\cite{ForAda} & \small\scriptsize CVPR'25 & 0.66 & 0.74 & 0.71 & 0.78 & 0.75 & 0.83 & 0.82 & 0.88 & 0.89 & 0.95 & 0.89 & 0.95 & 0.89 & 0.94 \\
\midrule
\rowcolor{rldfblue}
\textbf{Ours (12-Blk)} & — & 0.62 & 0.76 & 0.64 & 0.77 & 0.65 & 0.80 & 0.77 & 0.88 & 0.88 & 0.95 & 0.90 & 0.96 & \textbf{0.94} & \textbf{0.98} \\
\bottomrule
\end{tabular}}
\end{table*}

\begin{table*}[tb]
\centering
\caption{\textbf{Protocol~\#3 (\textbf{SimSwap}): Region-level performance (ACC\,/\,AUC).} \textbf{Bold} = best, \underline{underline} = second-best (per metric, per column). C-1: eye$+$nose, C-2: eye$+$nose$+$mouth, C-3: eye$+$nose$+$mouth$+$eyebrow. }
\label{tab:supp_simswap}
\setlength{\tabcolsep}{3pt}
\renewcommand{\arraystretch}{1.3}
\resizebox{\linewidth}{!}{%
\begin{tabular}{l c cc cc cc cc cc cc cc}
\toprule
\textbf{Detector} & \textbf{Venue} & \multicolumn{2}{c}{\small\textbf{Eye}} & \multicolumn{2}{c}{\small\textbf{Nose}} & \multicolumn{2}{c}{\small\textbf{Mouth}} & \multicolumn{2}{c}{\small\textbf{C-1}} & \multicolumn{2}{c}{\small\textbf{C-2}} & \multicolumn{2}{c}{\small\textbf{C-3}} & \multicolumn{2}{c}{\small\textbf{Full}} \\
\cmidrule(lr){3-4} \cmidrule(lr){5-6} \cmidrule(lr){7-8} \cmidrule(lr){9-10} \cmidrule(lr){11-12} \cmidrule(lr){13-14} \cmidrule(lr){15-16}
 & & \small ACC & \small AUC & \small ACC & \small AUC & \small ACC & \small AUC & \small ACC & \small AUC & \small ACC & \small AUC & \small ACC & \small AUC & \small ACC & \small AUC \\
\midrule
Xception~\cite{ff++} & \small\scriptsize ICCV'19 & 0.62 & 0.69 & 0.63 & 0.70 & 0.70 & 0.78 & 0.63 & 0.71 & 0.69 & 0.77 & 0.70 & 0.79 & 0.70 & 0.82 \\
F3Net~\cite{F3Net} & \small\scriptsize ECCV'20 & 0.60 & 0.75 & 0.60 & 0.75 & 0.64 & 0.79 & 0.61 & 0.74 & 0.63 & 0.78 & 0.64 & 0.80 & 0.59 & 0.73 \\
RECCE~\cite{RECCE} & \small\scriptsize CVPR'22 & 0.63 & 0.73 & 0.63 & 0.73 & 0.69 & 0.80 & 0.65 & 0.75 & 0.68 & 0.80 & 0.69 & 0.81 & 0.61 & 0.70 \\
SBI~\cite{SBI} & \small\scriptsize CVPR'22 & 0.58 & 0.67 & 0.58 & 0.68 & 0.58 & 0.67 & 0.67 & 0.78 & \underline{0.74} & \underline{0.86} & 0.74 & 0.86 & 0.82 & \underline{0.92} \\
LGrad~\cite{LGrad} & \small\scriptsize CVPR'23 & 0.50 & 0.58 & 0.50 & 0.58 & 0.50 & 0.58 & 0.50 & 0.58 & 0.50 & 0.58 & 0.50 & 0.58 & 0.50 & 0.58 \\
UCF~\cite{yan2023ucf} & \small\scriptsize ICCV'23 & \underline{0.70} & \textbf{0.83} & \underline{0.71} & \textbf{0.84} & \underline{0.72} & \textbf{0.85} & \underline{0.73} & \textbf{0.85} & 0.74 & 0.86 & 0.77 & \textbf{0.89} & 0.70 & 0.83 \\
FatFormer~\cite{fatformer} & \small\scriptsize CVPR'24 & 0.51 & 0.65 & 0.51 & 0.67 & 0.50 & 0.64 & 0.53 & 0.70 & 0.54 & 0.72 & 0.57 & 0.75 & 0.75 & 0.90 \\
FreqNet~\cite{FreqNet} & \small\scriptsize AAAI'24 & 0.38 & 0.45 & 0.38 & 0.46 & 0.38 & 0.46 & 0.39 & 0.46 & 0.39 & 0.47 & 0.39 & 0.47 & 0.39 & 0.46 \\
ProDet~\cite{prodet} & \small\scriptsize NeurIPS'24 & \textbf{0.71} & \underline{0.79} & \textbf{0.75} & \underline{0.82} & \textbf{0.74} & \underline{0.83} & \textbf{0.77} & \underline{0.84} & \textbf{0.81} & \textbf{0.89} & \textbf{0.81} & 0.88 & 0.80 & 0.88 \\
DFD-FCG~\cite{dfd_fcg} & \small\scriptsize CVPR'25 & 0.54 & 0.55 & 0.51 & 0.51 & 0.58 & 0.61 & 0.56 & 0.58 & 0.62 & 0.68 & 0.63 & 0.71 & 0.68 & 0.80 \\
Effort~\cite{effort} & \small\scriptsize ICML'25 & 0.62 & 0.67 & 0.61 & 0.66 & 0.60 & 0.65 & 0.68 & 0.73 & 0.71 & 0.78 & 0.74 & 0.81 & \underline{0.82} & 0.89 \\
ForAda~\cite{ForAda} & \small\scriptsize CVPR'25 & 0.60 & 0.65 & 0.64 & 0.69 & 0.63 & 0.68 & 0.67 & 0.73 & 0.71 & 0.78 & 0.73 & 0.80 & 0.70 & 0.76 \\
\midrule
\rowcolor{rldfblue}
\textbf{Ours (12-Blk)} & — & 0.54 & 0.65 & 0.59 & 0.72 & 0.55 & 0.66 & 0.65 & 0.79 & 0.70 & 0.83 & \underline{0.77} & \underline{0.88} & \textbf{0.84} & \textbf{0.93} \\
\bottomrule
\end{tabular}}
\end{table*}

\begin{table*}[tb]
\centering
\caption{\textbf{Protocol~\#3 (\textbf{UniFace}): Region-level performance (ACC\,/\,AUC).} \textbf{Bold} = best, \underline{underline} = second-best (per metric, per column). C-1: eye$+$nose, C-2: eye$+$nose$+$mouth, C-3: eye$+$nose$+$mouth$+$eyebrow. }
\label{tab:supp_uniface}
\setlength{\tabcolsep}{3pt}
\renewcommand{\arraystretch}{1.3}
\resizebox{\linewidth}{!}{%
\begin{tabular}{l c cc cc cc cc cc cc cc}
\toprule
\textbf{Detector} & \textbf{Venue} & \multicolumn{2}{c}{\small\textbf{Eye}} & \multicolumn{2}{c}{\small\textbf{Nose}} & \multicolumn{2}{c}{\small\textbf{Mouth}} & \multicolumn{2}{c}{\small\textbf{C-1}} & \multicolumn{2}{c}{\small\textbf{C-2}} & \multicolumn{2}{c}{\small\textbf{C-3}} & \multicolumn{2}{c}{\small\textbf{Full}} \\
\cmidrule(lr){3-4} \cmidrule(lr){5-6} \cmidrule(lr){7-8} \cmidrule(lr){9-10} \cmidrule(lr){11-12} \cmidrule(lr){13-14} \cmidrule(lr){15-16}
 & & \small ACC & \small AUC & \small ACC & \small AUC & \small ACC & \small AUC & \small ACC & \small AUC & \small ACC & \small AUC & \small ACC & \small AUC & \small ACC & \small AUC \\
\midrule
Xception~\cite{ff++} & \small\scriptsize ICCV'19 & 0.61 & 0.67 & 0.67 & 0.73 & 0.67 & 0.74 & 0.68 & 0.74 & 0.71 & 0.79 & 0.70 & 0.78 & 0.71 & 0.81 \\
F3Net~\cite{F3Net} & \small\scriptsize ECCV'20 & 0.55 & 0.63 & 0.57 & 0.68 & 0.57 & 0.69 & 0.58 & 0.68 & 0.60 & 0.72 & 0.60 & 0.71 & 0.67 & 0.80 \\
RECCE~\cite{RECCE} & \small\scriptsize CVPR'22 & 0.59 & 0.67 & 0.63 & 0.72 & 0.61 & 0.72 & 0.66 & 0.75 & 0.68 & 0.78 & 0.67 & 0.77 & 0.78 & 0.86 \\
SBI~\cite{SBI} & \small\scriptsize CVPR'22 & 0.67 & 0.79 & 0.60 & 0.72 & 0.59 & 0.71 & 0.72 & 0.84 & 0.77 & 0.88 & 0.78 & 0.89 & 0.83 & 0.93 \\
LGrad~\cite{LGrad} & \small\scriptsize CVPR'23 & 0.50 & 0.53 & 0.50 & 0.53 & 0.50 & 0.54 & 0.50 & 0.53 & 0.50 & 0.53 & 0.50 & 0.53 & 0.50 & 0.52 \\
UCF~\cite{yan2023ucf} & \small\scriptsize ICCV'23 & 0.59 & 0.67 & 0.60 & 0.70 & 0.62 & 0.73 & 0.63 & 0.73 & 0.65 & 0.76 & 0.64 & 0.75 & 0.72 & 0.84 \\
FatFormer~\cite{fatformer} & \small\scriptsize CVPR'24 & 0.52 & 0.63 & 0.51 & 0.59 & 0.51 & 0.60 & 0.53 & 0.69 & 0.56 & 0.74 & 0.60 & 0.79 & 0.73 & 0.89 \\
FreqNet~\cite{FreqNet} & \small\scriptsize AAAI'24 & 0.38 & 0.45 & 0.39 & 0.46 & 0.39 & 0.45 & 0.39 & 0.46 & 0.39 & 0.46 & 0.39 & 0.46 & 0.38 & 0.46 \\
ProDet~\cite{prodet} & \small\scriptsize NeurIPS'24 & \textbf{0.74} & \underline{0.82} & \textbf{0.76} & \textbf{0.84} & \textbf{0.76} & \textbf{0.83} & 0.81 & \underline{0.89} & 0.83 & \underline{0.92} & 0.83 & 0.91 & 0.85 & 0.94 \\
DFD-FCG~\cite{dfd_fcg} & \small\scriptsize CVPR'25 & 0.64 & 0.71 & 0.59 & 0.63 & 0.61 & 0.65 & 0.69 & 0.81 & 0.71 & 0.87 & 0.72 & 0.89 & 0.72 & 0.93 \\
Effort~\cite{effort} & \small\scriptsize ICML'25 & \underline{0.71} & 0.77 & 0.64 & 0.69 & 0.65 & 0.70 & 0.79 & 0.86 & 0.83 & 0.91 & 0.85 & \underline{0.94} & 0.87 & \underline{0.96} \\
ForAda~\cite{ForAda} & \small\scriptsize CVPR'25 & 0.71 & 0.78 & \underline{0.73} & 0.80 & \underline{0.69} & 0.77 & \underline{0.81} & 0.87 & \underline{0.85} & 0.91 & \underline{0.86} & 0.93 & \underline{0.87} & 0.94 \\
\midrule
\rowcolor{rldfblue}
\textbf{Ours (12-Blk)} & — & 0.70 & \textbf{0.84} & 0.68 & \underline{0.82} & 0.60 & \underline{0.78} & \textbf{0.83} & \textbf{0.92} & \textbf{0.88} & \textbf{0.95} & \textbf{0.91} & \textbf{0.97} & \textbf{0.95} & \textbf{0.98} \\
\bottomrule
\end{tabular}}
\end{table*}

% ── G. Qualitative Failure Cases ──────────────────────────────────────────────
\clearpage
\raggedbottom
\subsection*{J.\; Qualitative Analysis and Failure Cases}
\label{ap:I}
\begin{figure}[htbp]
    \centering
\includegraphics[width=0.95\linewidth,height=0.80\textheight,keepaspectratio]{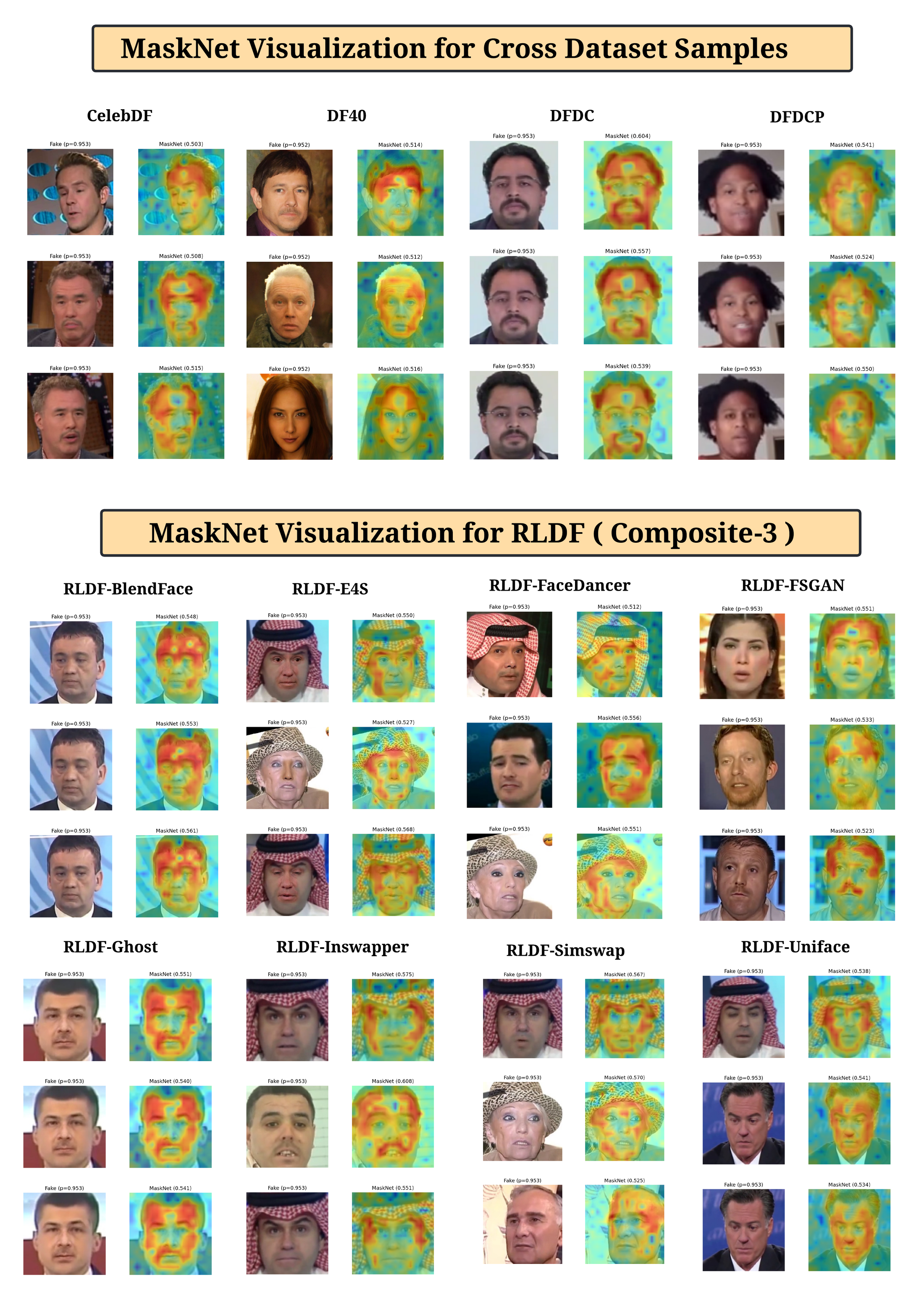}
\caption{MaskNet visualization: predicted forgery responses highlighting manipulated regions.}
    \label{fig:masknetviz}
\end{figure}

\begin{figure}[htbp]
    \centering
\includegraphics[width=\linewidth,height=0.92\textheight,keepaspectratio]{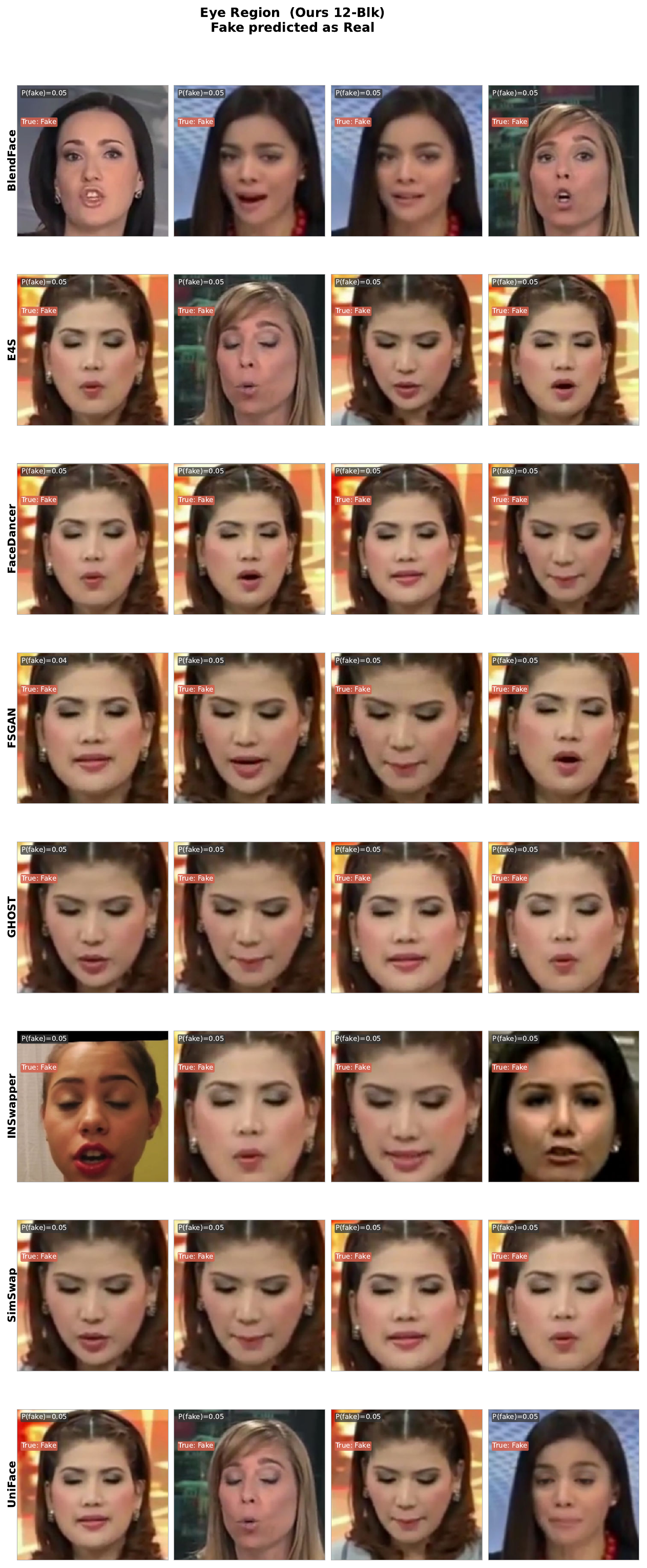}
    \caption{Failure cases: \textbf{Eye} region. Fake images predicted as Real by Ours~(12-Blk).}
    \label{fig:supp_fail_eye}
\end{figure}

\begin{figure}[htbp]
    \centering
\includegraphics[width=\linewidth,height=0.92\textheight,keepaspectratio]{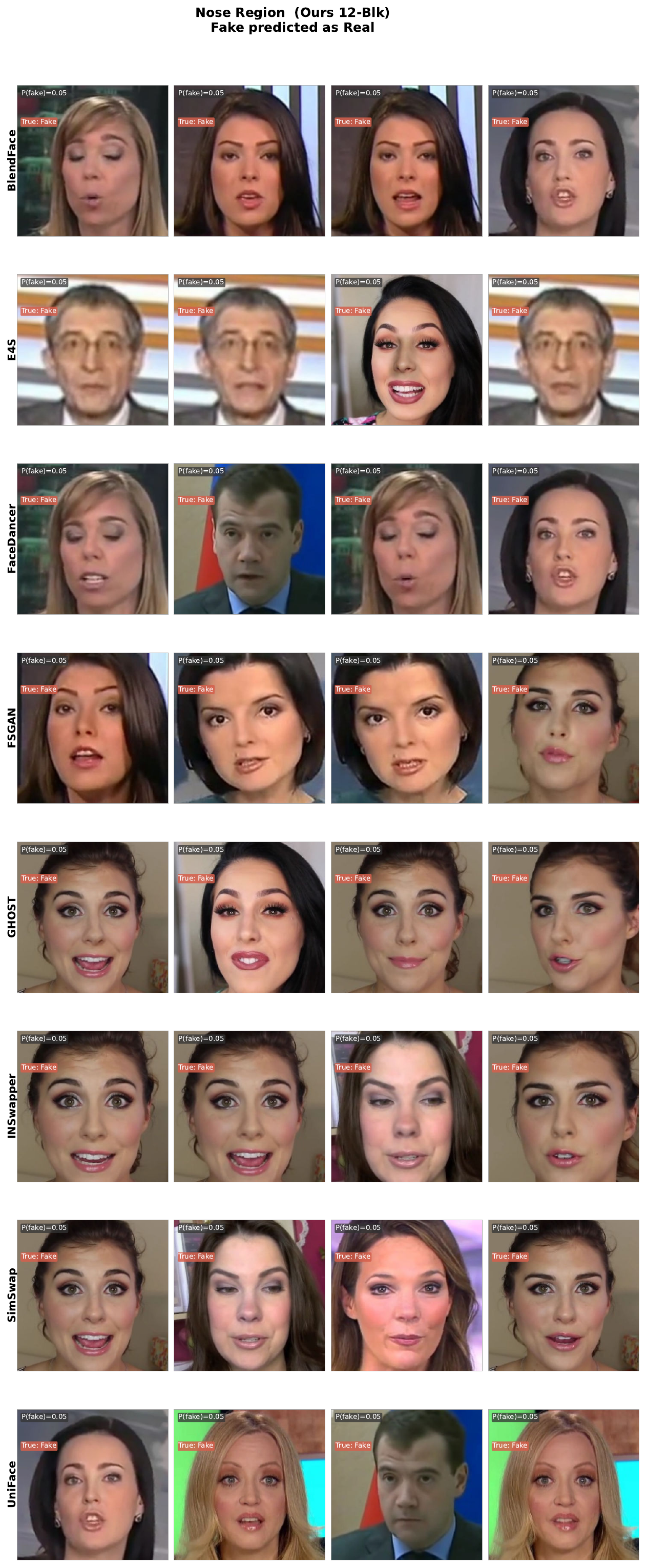}
    \caption{Failure cases: \textbf{Nose} region.}
    \label{fig:supp_fail_nose}
\end{figure}

\begin{figure}[htbp]
    \centering
\includegraphics[width=\linewidth,height=0.92\textheight,keepaspectratio]{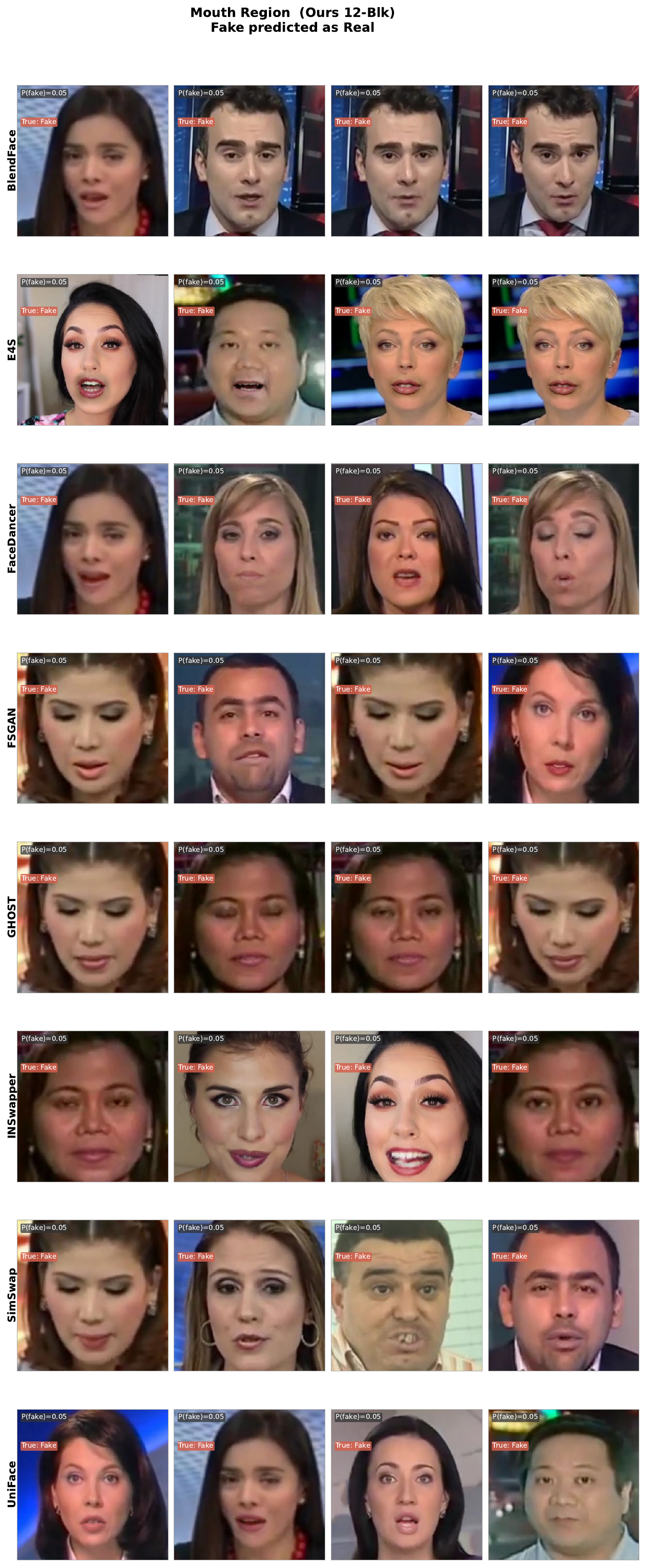}
    \caption{Failure cases: \textbf{Mouth} region.}
    \label{fig:supp_fail_mouth}
\end{figure}

\begin{figure}[htbp]
    \centering
\includegraphics[width=\linewidth,height=0.92\textheight,keepaspectratio]{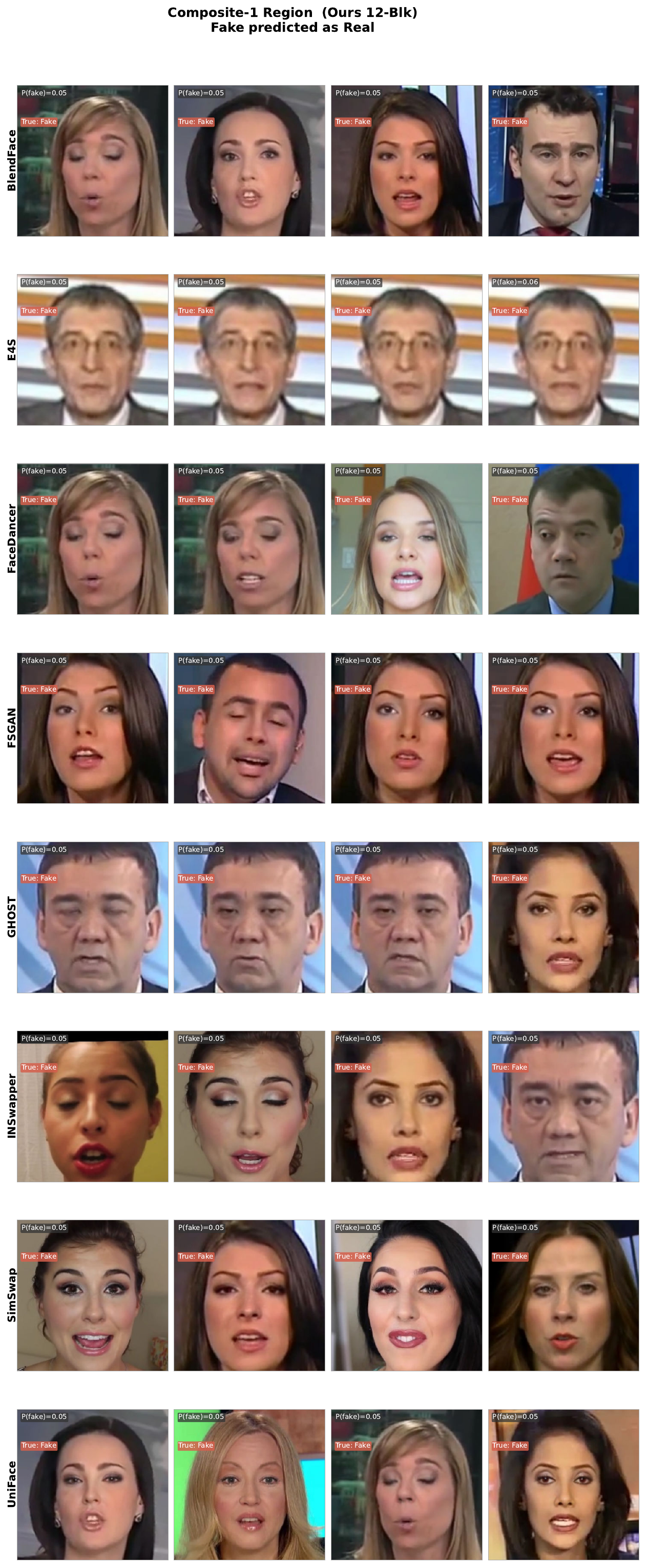}
    \caption{Failure cases: \textbf{Composite-1} (Eye+Nose).}
    \label{fig:supp_fail_c1}
\end{figure}

\begin{figure}[htbp]
    \centering
\includegraphics[width=\linewidth,height=0.92\textheight,keepaspectratio]{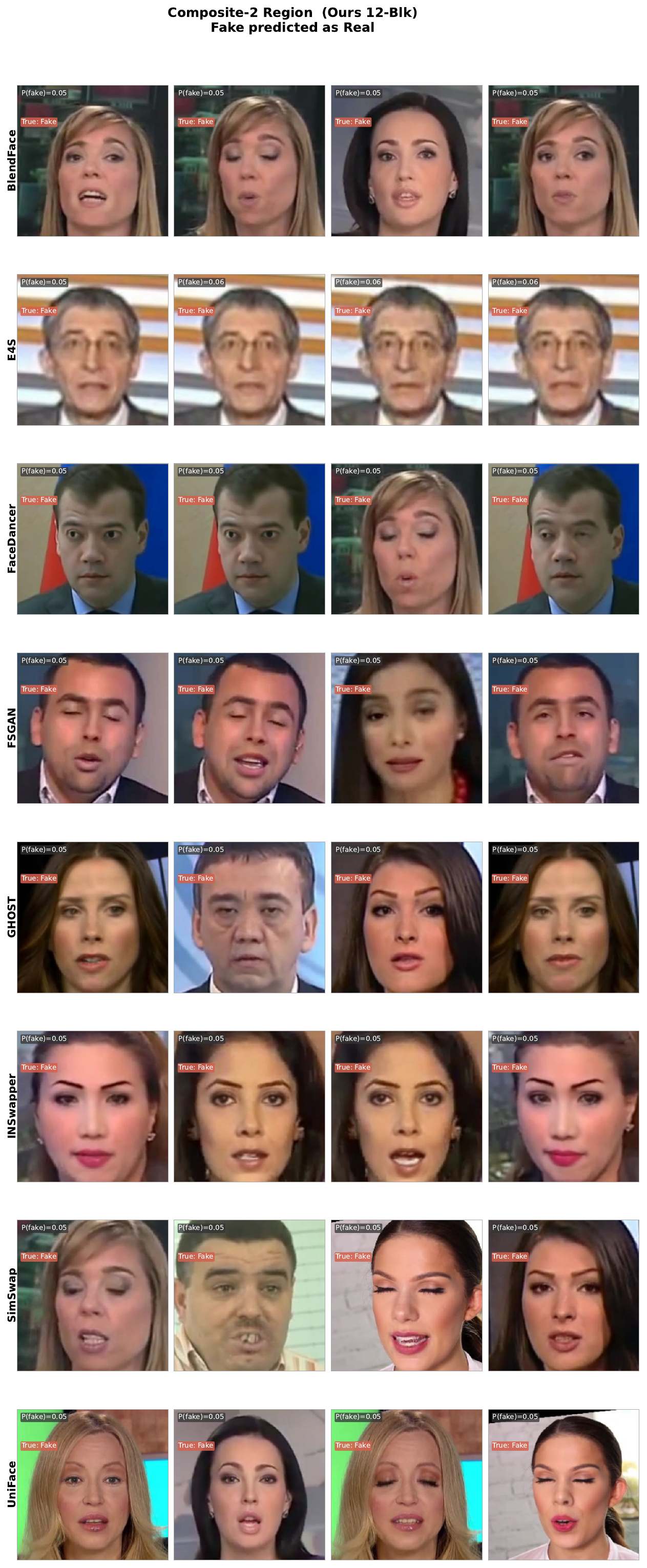}
    \caption{Failure cases: \textbf{Composite-2} (Eye+Nose+Mouth).}
    \label{fig:supp_fail_c2}
\end{figure}

\begin{figure}[htbp]
    \centering
\includegraphics[width=\linewidth,height=0.92\textheight,keepaspectratio]{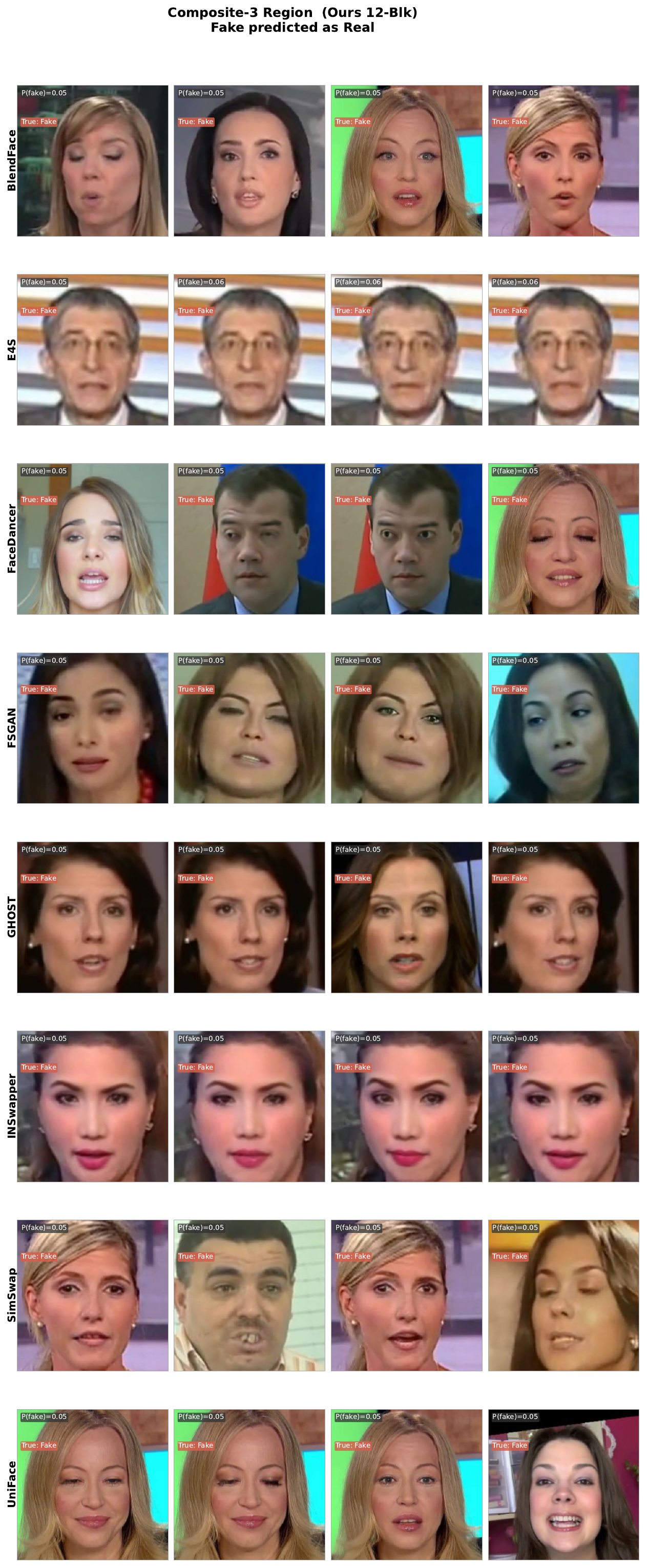}
    \caption{Failure cases: \textbf{Composite-3} (Eye+Nose+Mouth+Eyebrow).}
    \label{fig:supp_fail_c3}
\end{figure}

\begin{figure}[htbp]
    \centering
\includegraphics[width=\linewidth,height=0.92\textheight,keepaspectratio]{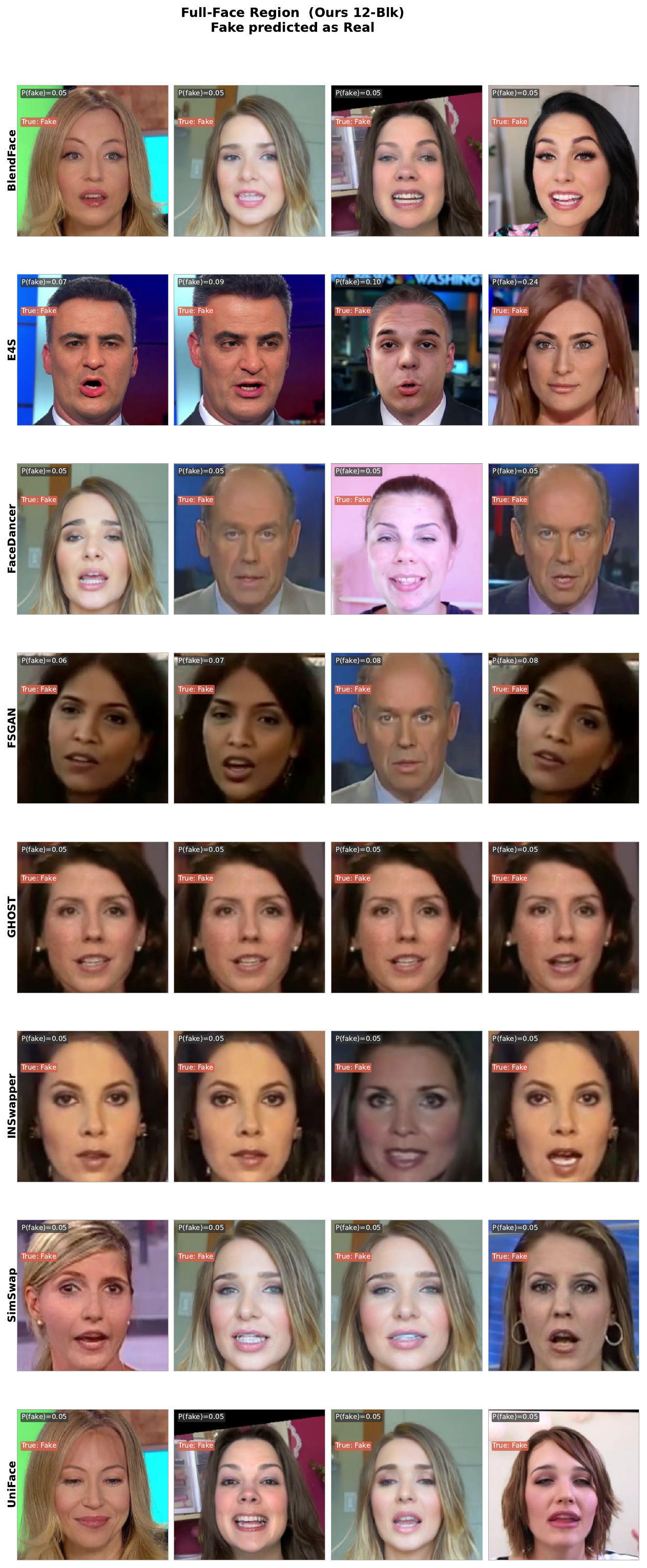}
    \caption{Failure cases:  \textbf{Full Face} region.}
    \label{fig:supp_fail_fullface}
\end{figure}

\vspace{-1mm}
Figures~\ref{fig:masknetviz} and \ref{fig:supp_fail_eye}--\ref{fig:supp_fail_fullface} show MaskNet predictions and failure cases of Ours~(12-Blk). MaskNet (Fig.~\ref{fig:masknetviz}) highlights manipulated regions, while failures correspond to false negatives, mainly in Eye and Nose, where the manipulated area is $<$10\% and artifacts are subtle. Many Eye failures arise from BlendFace and SimSwap, indicating difficulty due to both \emph{region size} and \emph{subtle artifacts}.


\end{document}